\RequirePackage{fix-cm}
\documentclass[twocolumn]{svjour3}          
\smartqed  
\usepackage{graphicx}
\PassOptionsToPackage{hyphens}{url}\usepackage{hyperref}       

\usepackage{xurl}  

\usepackage{algorithm} 
\usepackage{algorithmicx} 
\usepackage{algpseudocode}
\usepackage{adjustbox}
\usepackage{graphicx}
\usepackage{caption}
\usepackage{subcaption}
\usepackage{amsmath}
\usepackage{wrapfig}
\usepackage{multirow}
\usepackage{xcolor}
\usepackage{amsfonts}
\usepackage{enumitem}

\usepackage[numbers]{natbib}
 
\DeclareMathOperator{\D}{\mathcal{D}}
\DeclareMathOperator{\R}{\mathcal{R}}
\DeclareMathOperator{\Gnew}{G_{\mbox{\textrm{new}}}}
\DeclareMathOperator{\G}{G}
\DeclareMathOperator{\Anew}{A_{\mbox{\textrm{new}}}}

\begin{document}

\title{Lucid Dreaming for Experience Replay: \\ Refreshing Past States with the Current Policy 
}

\titlerunning{Lucid Dreaming for Experience Replay}        

\author{Yunshu Du   \and
        Garrett Warnell \and 
        Assefaw Gebremedhin \and
        \\ Peter Stone \and 
        Matthew E. Taylor 
}


\institute{Yunshu Du (corresponding author) \at
              Washington State University \\
              \email{yunshu.du@wsu.edu}           
          \and
          Garrett Warnell \at
              Army Research Laboratory \\
              \email{garrett.a.warnell.civ@mail.mil}
          \and 
          Assefaw Gebremedhin \at
            Washington State University \\
            \email{assefaw.gebremedhin@wsu.edu}
          \and 
          Peter Stone \at 
            The University of Texas at Austin \\
            Sony AI \\
            \email{pstone@cs.utexas.edu}
         \and 
         Matthew E. Taylor \at 
            University of Alberta \\
            Alberta Machine Intelligence Institute \\
            Washington State University \\
            \email{matthew.e.taylor@ualberta.ca}
}

\date{Received: date / Accepted: date}

\maketitle

\begin{abstract}
Experience replay (ER) improves the data efficiency of off-policy reinforcement learning (RL) algorithms by allowing an agent to store and reuse its past experiences in a replay buffer. While many techniques have been proposed to enhance ER by biasing how experiences are sampled from the buffer, thus far they have not considered strategies for \emph{refreshing} experiences inside the buffer. In this work, we introduce \emph{\textbf{L}uc\textbf{i}d \textbf{D}reaming for \textbf{E}xperience \textbf{R}eplay (LiDER)}, a conceptually new framework that allows replay experiences to be refreshed by leveraging the agent's current policy. 
LiDER consists of three steps: First, LiDER moves an agent back to a past state. Second, from that state, LiDER then lets the agent execute a sequence of actions by following its current policy---as if the agent were ``dreaming'' about the past 
and can try out different behaviors to encounter new experiences in the dream. Third, LiDER stores and reuses the new experience if it turned out better than what the agent previously experienced, i.e., to \emph{refresh} its memories. LiDER is designed to be easily incorporated into off-policy, multi-worker RL algorithms that use ER; we present in this work a case study of applying LiDER to an actor-critic based algorithm. Results show LiDER consistently improves performance over the baseline in six Atari 2600 games. Our open-source implementation of LiDER and the data used to generate all plots in this work are available at \url{github.com/duyunshu/lucid-dreaming-for-exp-replay}.

\end{abstract}

\section{Introduction}
\label{sec:intro}
One of the critical components contributing to the recent success of integrating reinforcement learning (RL) with deep learning is the experience replay (ER) mechanism \cite{dqn}. While deep RL algorithms are often data-hungry, ER enhances data efficiency by allowing the agent to store and reuse its past experiences in a replay buffer \cite{ERlin1992}. Several techniques have been proposed to enhance ER to further reduce data complexity and one of the commonly used techniques is to influence the order of replayed experiences. Instead of replaying experiences uniformly at random \cite{ddpg,dqn}, studies have found that sampling experiences with different priorities can speed up the learning \cite{RER,ReFER,PER,ER-RND,ERO}. 

Biased experience sampling affects \emph{how} the experiences are replayed. However, it does not consider \emph{what} experience to replay. An experience comprises a state, the action taken at that state, and the return\footnote{A one-step reward $r$ is usually stored instead of the cumulative return (e.g., \citet{dqn}). In this work, we follow \citet{sil} and store the Monte-Carlo return $\G$; we fully describe the buffer structure in Section \ref{sec:lider}.} obtained by following the agent's current policy from that state. Existing ER methods usually operate on a fixed set of experiences. That is, once an experience is stored, it remains static inside the buffer until it ages out. An experience from several steps ago may no longer be useful for the current policy to replay because it was generated in the past with a much worse policy. If the agent were given a chance to try again at the same place, its current policy might be able to take different actions that lead to higher returns than what it obtained in the past. \emph{What} the agent should replay is therefore the newer and updated experience, instead of the older one. Given this intuition, we propose in this work \emph{\textbf{L}uc\textbf{i}d \textbf{D}reaming for \textbf{E}xperience \textbf{R}eplay (LiDER)}, a conceptually new framework that \emph{refreshes} past experiences by leveraging the agent's current policy, allowing the agent to learn from valuable data generated by its newer self.  

LiDER refreshes replay experiences in three steps: First, LiDER moves the agent back to a state it has visited before. Second, LiDER lets the agent follow its current policy to generate a new trajectory from that state. Third, if the new trajectory led to a better outcome than what the agent previously experienced from that state, LiDER stores the new experience into a separate replay buffer and reuses it during training. We refer to this process as ``lucid dreaming for experience replay,'' because it is as if the agent were ``dreaming'' about the past 
and can control the dream to practice again in a past state to achieve better rewards---much like how research in sports science has found that a person's motor skills can be improved by consciously rehearsing the movements in a lucid dream (e.g., \citet{Stumbrys2016EffectivenessOM}).  

One limitation of LiDER is it requires environmental interactions to refresh past states. However, we carefully account for \emph{all} environment interactions, including steps taken to generate new trajectories, and show that LiDER reduces the overall sample complexity of learning compared to methods that do not refresh experiences. LiDER is applicable when a simulator exists for the task---either the task itself is a simulation like a video game or we can build a simulator of the real world---and the simulator is capable of teleporting the agent back to previously visited states and rolling forward in time from there.

The main contributions of this work are as follows: 
\begin{enumerate}
\item We propose LiDER, a conceptually new framework to \emph{refresh} replay experiences, allowing an agent to revisit and update past experiences using its current policy in off-policy, multi-worker RL algorithms. 
\item LiDER is implemented in an actor-critic based algorithm as a case study. 
\item We experimentally show LiDER outperforms the baseline method (where past experiences were not refreshed) in six Atari 2600 games, including two hard exploration games that are challenging for several RL benchmark algorithms. 
\item Analyses and ablation studies help illustrate the functioning of different components of LiDER. 
\item Two extensions demonstrate that LiDER is also capable of leveraging policies from external sources, i.e., a policy trained by a different RL algorithm and a behavior cloning policy pre-trained from non-expert human demonstrations. 
\item We open-source our implementation of LiDER and the data used to generate all plots in this work for reproducibility at \url{github.com/duyunshu/lucid-dreaming-for-exp-replay}.
\end{enumerate}

\section{Background}
\label{sec:background}
Our algorithm leverages several existing methods, which we briefly review in this section. 

\subsection{Reinforcement Learning}
We consider an RL problem to be modeled using a Markov decision process, represented by a 5-tuple \\ $\langle S, A, P, R, \gamma \rangle$. A \emph{state} $s_{t} \in S$ represents the environment at time $t$. An agent learns what \emph{action} $a_t \in \mathcal{A(}s\mathcal{)}$ to take in $s_{t}$ by interacting with the environment. The transition function $P(s_{t+1}|s_t,a_t)$ denotes the probability of reaching state $s_{t+1}$ after taking action $a_t$ at state $s_{t}$. A \emph{reward} $r_{t} \in \R \subset \mathbb{R}$ is given based on $a_t$ and $s_{t+1}$. The goal is to maximize the expected cumulative return $G_t = \sum_{k=0}^{\infty} \gamma^k r_{t+k}$ from time step $t$, where $\gamma \in [0,1]$ is a discount factor that determines the relative importance of future and immediate rewards \cite{sutton2018}. 

\subsection{Asynchronous Advantage Actor-Critic}
Policy-based methods such as the asynchronous advantage actor-critic (A3C) algorithm \cite{a3c} combine a deep neural network with the actor-critic framework. In this work, we leverage the A3C framework to learn both a \emph{policy function} $\pi(a_t|s_t;\theta)$ (parameterized as $\theta$) and a \emph{value function} $V(s_t;\theta_v)$ (parameterized as $\theta_v$). The policy function is the \emph{actor} that takes action. The value function is the \emph{critic} that evaluates the quality of the action against a baseline (e.g., state value). A3C directly minimizes the policy loss $L^{a3c}_{policy}$ as
\begin{equation*}
\begin{split}
L^{a3c}_{policy} = & \nabla_\theta \log(\pi(a_t|s_t;\theta))\bigl(Q^{(n)}(s_t, a_t;\theta,\theta_v)-V(s_t;\theta_v)\bigr) \\
& +\beta^{a3c}\mathcal{H}\nabla_\theta\bigl(\pi(s_t;\theta)\bigr),
\end{split}
\end{equation*}
where $Q^{(n)}(s_t, a_t;\theta,\theta_v)=\sum_{k=0}^{n-1} \gamma^{k}r_{t+k} + \gamma^{n}V(s_{t+n};\theta_v)$ is the $n$-step bootstrapped value that is bounded by a hyperparameter $t_{max}$ ($n\leq t_{max}$). $\mathcal{H}$ is an entropy regularizer for policy $\pi$ (weighted by $\beta^{a3c}$) which helps to prevent premature convergence to sub-optimal policies. The value loss $L^{a3c}_{value}$ is 
\begin{equation*}
L^{a3c}_{value} = \nabla_{\theta_v}\Bigl(\bigl(Q^{(n)}(s_t, a_t;\theta,\theta_v)-V(s_t;\theta_v)\bigr)^2\Bigr).
\end{equation*}
The full A3C loss $L^{a3c}$ given by \citet{a3c} is then 
\begin{equation} \label{eq:a3c_update}
L^{a3c}=L^{a3c}_{policy}+\alpha L^{a3c}_{value},
\end{equation}
where $\alpha$ is a weight for the value loss. A3C's architecture contains one global policy and $k$ parallel actor-critic workers. The workers run in parallel and each has its copy of the environment and parameters; each worker updates the global policy asynchronously using the data collected in its own environment. We use the feedforward version of A3C as it runs faster than, but with comparable performance to, the recurrent version \cite{a3c}. 

\subsection{Transformed Bellman Operator for A3C}
\label{sec:A3CTB}
The A3C algorithm uses reward clipping to help stabilize learning. However, \citet{dqfd} showed that clipping rewards to $[+1,-1]$ results in the agent being unable to distinguish between small and large rewards, thus hurting the performance in the long-term. \citet{apexdqn} introduced the \emph{transformed Bellman (TB) operator} to overcome this problem in the deep Q-network (DQN) algorithm \cite{dqn}. The authors consider reducing the scale of the action-value function while keeping the relative differences between rewards which enables DQN to use raw rewards instead of clipping. \citet{apexdqn} apply a transform function $h:\mathbb{R} \mapsto \mathbb{R}$ to reduce the scale of $Q^{(n)}(s_t, a_t;\theta,\theta_v)$ to 
\begin{equation*}
Q^{(n)}_{TB}(s_t, a_t;\theta,\theta_v) = \sum_{k=0}^{n-1} h\left( \gamma^{k}r_{t+k} + \gamma^{n}h^{-1} V\left(s_{t+n};\theta_v\right) \right), 
\end{equation*}
where 
\begin{equation*}
    h: z \mapsto \operatorname{sign}(z)\left(\sqrt{|z| + 1} - 1\right) + \varepsilon z
\end{equation*}
and
\begin{equation*}
h^{-1}: x \mapsto \operatorname{sign}(x)\left(\left(\frac{\sqrt{1+4 \varepsilon(|x|+1+\varepsilon)}-1}{2 \varepsilon}\right)^{2}-1\right),
\end{equation*}
and $\varepsilon$ is a constant that ensures $h^{-1}$ is Lipschitz continuous with a closed form inverse. \citet{apexdqn} also prove that the TB operator reduces the variance of the optimization goal while still enabling learning an optimal policy. Given this benefit, our previous work \cite{de2019jointly} applied the TB operator to A3C, denoted as A3CTB, and showed that A3CTB empirically outperforms A3C. 

\subsection{Self Imitation Learning for A3CTB} 
The \emph{self imitation learning} (SIL) algorithm \cite{sil} is motivated by the intuition that an agent can exploit its own past \emph{good} experiences and thus improve performance. Built upon the actor-critic framework \cite{a3c}, SIL adds a prioritized experience replay buffer $\D={(S, A, G)}$ to store the agent's past experiences, where $S$ is a state, $A$ is the action taken in $S$, and $\G$ is the \emph{Monte-Carlo} return from $S$ (i.e., the return is computed only after a terminal state is reached). In addition to the A3C loss in Equation \eqref{eq:a3c_update}, at each step $t$, SIL samples a minibatch from $\D$ for $M$ times and optimizes the following off-policy, actor-critic loss $L^{sil}_{policy}$ and $L^{sil}_{value}$: 
\begin{align}
    L^{sil}_{policy} &= - \log(\pi(a_t|s_t;\theta)) \bigr(G_t-V(s_t;\theta_v)\bigl)_{+}  \nonumber \\
    L^{sil}_{value} &= \frac{1}{2} ||\bigl(G_t-V(s_t;\theta_v)\bigr)_{+}||^2, \nonumber
\end{align}
where $G_t=\sum_{k=0}^{\infty}\gamma^{k}r_{t+k}=r_{t} + \gamma G_{t+1}$ is the discounted cumulative return, $V$ is the state value. The value of $\bigr(G_t-V(s_t;\theta_v)\bigl)$ is called the \emph{advantage}, and the max operator $(\cdot)_{+}=max(\cdot, 0)$ meaning that only experiences with positive advantage values (i.e., good experiences) can contribute to the policy update. The experience buffer is prioritized by $\bigl(G_t-V(s_t;\theta_v)\bigr)_{+}$ to increase the chance that a good experience is sampled. The SIL loss $L^{sil}$ is then 
\begin{equation}\label{eq:sil_update}
    L^{sil} = L^{sil}_{policy}+\beta^{sil}L^{sil}_{value},
\end{equation}
where $\beta^{sil}$ is a weight for the value loss. 

The SIL algorithm minimizes both the A3C loss $L^{a3c}$ (Equation \eqref{eq:a3c_update}) and the SIL loss $L^{sil}$ (Equation \eqref{eq:sil_update}). Minimizing $L^{a3c}$ lets the agent learn by interacting with the environment and minimizing $L^{sil}$ allows the agent to also learn by replaying its past good experiences. In our previous work \cite{de2019jointly}, we leveraged this framework to incorporate SIL into A3CTB, denoted as A3CTBSIL. Specifically, the return $G_t$ is transformed to the \emph{TB return} using operators $h$ and $h^{-1}$ discussed in Section \ref{sec:A3CTB} as: 
\begin{equation*}
    G_t = h(r_t + \gamma h^{-1}(G_{t+1})).
\end{equation*}
For simplicity, from this point on we will use the word ``return'' to refer to ``TB return.'' 
Our previous work has shown that A3CTBSIL outperformed both the A3C and A3CTB algorithms \cite{de2019jointly}. This article, therefore, uses an implementation of A3CTBSIL as the baseline.\footnote{The implementation of A3CTBSIL is open-sourced at \url{github.com/gabrieledcjr/DeepRL}. In \citet{de2019jointly}, we also considered using demonstrations to improve A3CTBSIL, which is not the baseline used in 
this work.}

\section{Lucid dreaming for experience replay} 
\label{sec:lider}

In this work, we introduce \emph{\textbf{L}uc\textbf{i}d \textbf{D}reaming for \textbf{E}xperi-ence \textbf{R}eplay (LiDER)}, a conceptually new framework that allows replay experiences to be refreshed by following the agent's current policy. LiDER consists of three steps: First, LiDER moves an agent back to a past state. Second, from that state, LiDER then lets the agent execute a sequence of actions by following its current policy---as if the agent were ``dreaming'' about the past 
and can try out different behaviors to encounter new experiences in the dream. Third, LiDER stores and reuses the new experience if it turned out better than what the agent previously experienced, i.e., to \emph{refresh} its memories. From a high-level perspective, we expect LiDER to help learning by allowing the agent to witness and learn from alternate and advantageous behaviors. 

LiDER is designed to be easily incorporated into off-policy, multi-worker RL algorithms that use ER. We implement LiDER in the A3C framework with SIL for two reasons. First, the A3C architecture \cite{a3c} allows us to conveniently add the ``refreshing'' component (which we will introduce in the next paragraph) in parallel with A3C and SIL workers, which saves wall-clock time for training. Second, the SIL framework \cite{sil} is an off-policy actor-critic algorithm that integrates an experience replay buffer with A3C in a straightforward way, enabling us to directly leverage the return G of an episode for a policy update---a key component of LiDER.\footnote{Note that while the A3C algorithm is on-policy, integrating A3C with SIL makes it an off-policy algorithm (as in \citet{sil}).}

\begin{figure}[t]
\centering
  \includegraphics[width=0.49\textwidth]{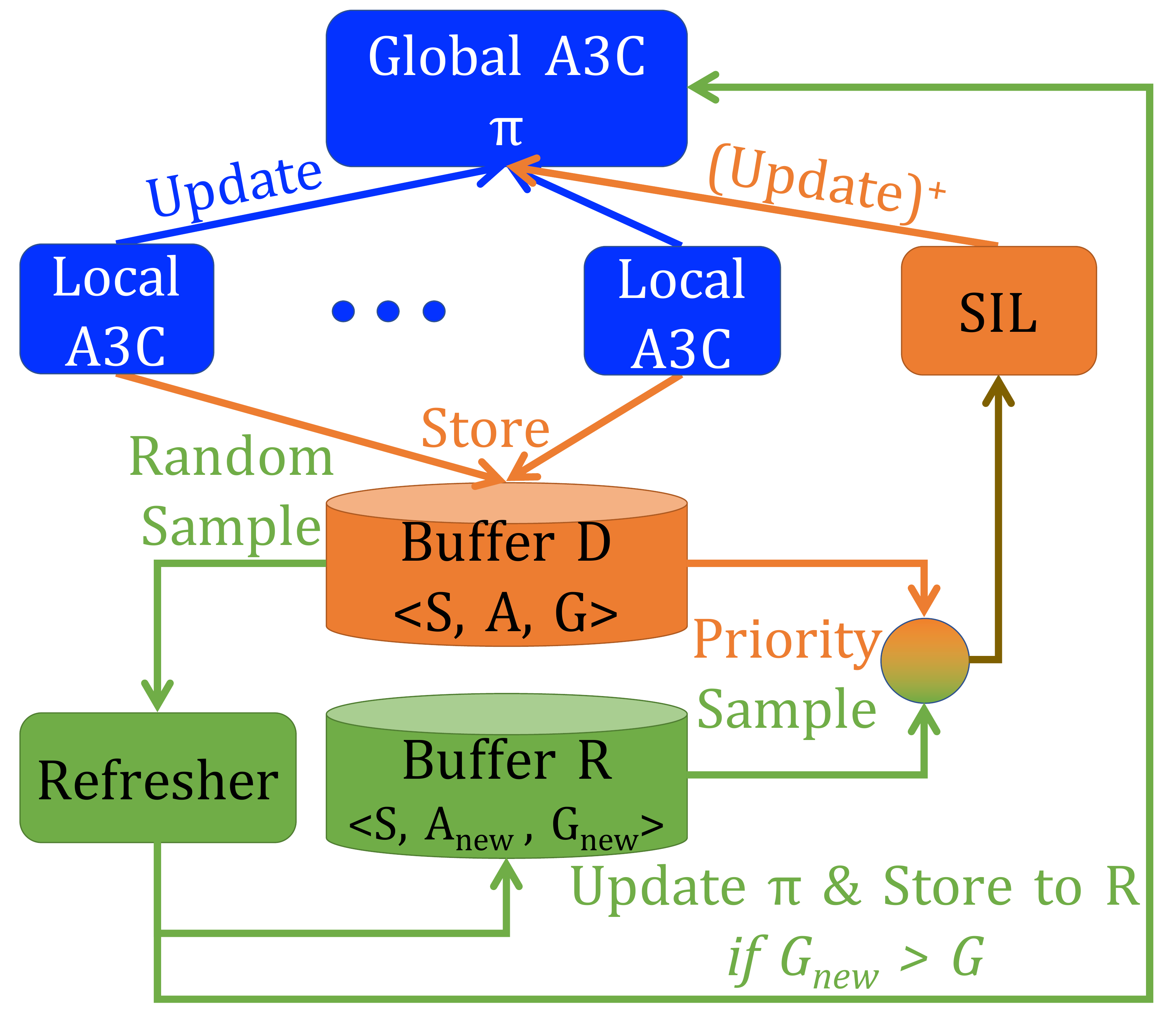}
  \caption{LiDER architecture. A3C components are in blue and
  SIL components are in orange. We introduce the novel concept of a refresher worker, in green, to generate new experiences from a randomly sampled past state from buffer $\D$ by leveraging the agent's \emph{current} policy. If the new experiences obtain a higher return than what is currently stored in the replay buffer $\D$, they are used to update global policy $\pi$ and are also stored into replay buffer $\R$ for reuse.} 
  \label{fig:arch}
\end{figure}

Figure \ref{fig:arch} shows the proposed implementation architecture for LiDER. A3C components are in blue: $k$ parallel workers interact with their own copies of the environment to update the global policy $\pi$ \cite{a3c}. SIL components are in orange: one SIL worker and a prioritized replay buffer $\D$ are added to A3C \cite{sil}. Buffer $\D$ stores all experiences from the A3C workers in the form of $\D=\{S, A, \G \}$ (as described in Section \ref{sec:background}). Buffer $\D$ is prioritized by the advantage value such that good states are more likely to be sampled. The SIL worker runs in parallel with the A3C workers but does not interact with the environment; it only samples from buffer $\D$ and updates $\pi$ using samples that have positive advantage values (Equation \eqref{eq:sil_update}). 

We introduce the novel concept of a ``refresher'' wor-ker in parallel with A3C and SIL to generate new data from past states (shown in green). The refresher has access to the environment and takes randomly sampled states from buffer $\D$ as input. For each state sampled, the refresher resets the environment to that state and uses the agent's current policy to perform a rollout until reaching a terminal state (e.g., the agent loses a life). If the Monte-Carlo return of the new trajectory, $\Gnew$, is higher than the previous return, $\G$ (sampled from buffer $\D$), the new trajectory is immediately used to update the global policy $\pi$. The update is done in the same way as the A3C workers (Equation \eqref{eq:a3c_update}, replacing $Q^{n}$ with $\Gnew$). The new trajectory is also stored in a prioritized buffer $\R=\{S, \Anew, \Gnew \}$ (prioritized by advantage, like in buffer $\D$) if $\Gnew > \G$. Finally, the SIL worker samples from both buffers as follows. A batch of samples is taken from each of the buffers $\D$ and $\R$ (i.e., two batches in total), prioritized by advantage. Samples from both batches are mixed together and put into a temporary buffer, shown in the green-orange circle in Figure \ref{fig:arch}; 
the temporary buffer treats all samples with an equal priority. One batch of samples is then taken (with replacement) from the mixture of the two batches (shown as the brown arrow) and SIL performs updates using the good samples from this batch. Note that, although samples in the temporary buffer were initialized with equal priorities, the sampling process is not uniformly at random since we use the implementation of prioritized sampling with stochastic prioritization as describe in Section 3.3 of \citet{PER}. Having this temporary buffer to mix together transitions from buffers $\D$ and $\R$ allows the agent to select past and/or refreshed experiences flexibly 
without needing a fixed sampling strategy.  
We summarize LiDER's refresher worker's procedure in Algorithm \ref{algo:refresher}. Full pseudo-code for the A3C and SIL workers is in Appendix \ref{sec:appe-a3csil}.

\begin{algorithm*}
    \caption{LiDER: Refresher Worker}
    \label{algo:refresher}
    \begin{algorithmic}[1]
    \State \textit{// Assume shared global policy $\pi$, replay buffer $\D$, replay buffer $\R$}
    \While{$T<T_{max}$} \Comment{$T_{max}$ = 50 million}
        \State Synchronize refresher's policy with the global policy: $\pi_{e}(\cdot|\theta_{e}) \leftarrow \pi$ \label{line:refresher-sync-policy}
        \State Synchronize global step $T$ from the most recent A3C worker
        \State Initialize $S \leftarrow \emptyset$, $\Anew \leftarrow \emptyset$, $R \leftarrow \emptyset$
        \State Randomly take a sample $\{s, a, G\}$ from buffer $\D$, reset the environment to $s$
        \While {not terminal}
            \State Execute an action $s$, $a$, $r$, $s^{\prime}$ $\sim$ $\pi_{e}(s|\theta_{e})$
            \State Store the experience $S \leftarrow S \cup s$, $\Anew \leftarrow \Anew \cup$ $a$, $R \leftarrow R \cup r$
            \State Go to next state $s \leftarrow s^{\prime}$
            \State $T \leftarrow T+1$ \label{line:global_t}
        \EndWhile
        \State $\Gnew=\sum_{k=0}^{\infty}\gamma^{k}r_{t+k}$, $\forall r \in R$         \Comment{Compute the new return}
        \If{$\Gnew > \G$} 
            \State Update $\pi$ using $\{S, \Anew, \Gnew \}$ \Comment{Equation \eqref{eq:a3c_update}, replace $Q^{(n)}$ with $\Gnew$}
            \State Store to buffer $\R \leftarrow \R \cup \{S, \Anew, \Gnew \}$
        \EndIf
    \EndWhile
    \end{algorithmic}
\end{algorithm*}

The main benefit of LiDER is that it allows an agent to leverage its \emph{current} policy to refresh past experiences. However, LiDER does require the refresher to use additional environmental steps (see Algorithm \ref{algo:refresher} line \ref{line:global_t}: we account for the refresher steps when measuring the global steps), which can be concerning if acting in the environment is expensive. Despite this shortcoming, we show in our experiments (Section~\ref{sec:exp}) that the learning speedup LiDER provides actually reduces the overall number of environment interactions required. It seems that the high \emph{quality} of the refreshed experiences compensates for the additional \emph{quantity} of experiences an agent needs to learn. That is, by leveraging the refresher worker, LiDER can achieve a certain level of performance within a shorter period of time compared to without the refresher---an important benefit as RL algorithms are often data-hungry.

\section{Experiments and analyses}
\label{sec:exp}
We empirically evaluate LiDER in six Atari 2600 games \cite{bellemare2013arcade}: Gopher, NameThisGame, Alien, Ms.~Pac-Man, 
Freeway, and Montezuma's Revenge. We selected these gam-es because they cover a range of properties and difficulties. Based on the Atari game taxonomy defined by \citet{bellemare2016unifying}, Gopher and NameThisGame are easy exploration games with dense reward functions; they are relatively easy to learn. Alien and Ms.~Pac-Man are hard exploration games with dense reward functions; they are considered to be hard games. Freeway and Montezuma's Revenge are also hard exploration games but with sparse reward functions; they are considered the hardest games and are challenging for several benchmark RL algorithms (e.g., \citet{bellemare2013arcade}, \citet{espeholt2018impala}, and \citet{a3c}). 

In the next subsection, we compare A3CTBSIL (the baseline method from \citet{de2019jointly}, which uses only the blue and the orange components in Figure \ref{fig:arch}) and LiDER (our proposed framework in which the agent's current policy is used as the refresher) to show LiDER outperforms A3CTBSIL in all games (See Appendix \ref{sec:appe-parameters} for implementation details). 
Section \ref{sec:analyses} then introduces analyses to understand why LiDER helps learning. In Section \ref{sec:ablation}, we conduct three ablation studies to validate that our design choices for LiDER were well-founded. Finally, in Section \ref{sec:lider-ta-bc}, we present two extensions and show that LiDER can leverage other policies, rather than its current policy, to refresh past states. 

\subsection{Leveraging the current policy to refresh past states}
\label{sec:lider-a3c}

\begin{figure*}
     \centering
     \begin{subfigure}[th]{0.35\textwidth}
         \centering
         \includegraphics[width=\textwidth]{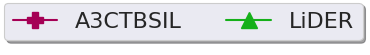}
     \end{subfigure}
     \hspace{0.5\textwidth} 
     \vspace{10pt}
     \begin{subfigure}[th]{0.48\textwidth}
         \centering
         \includegraphics[width=\textwidth]{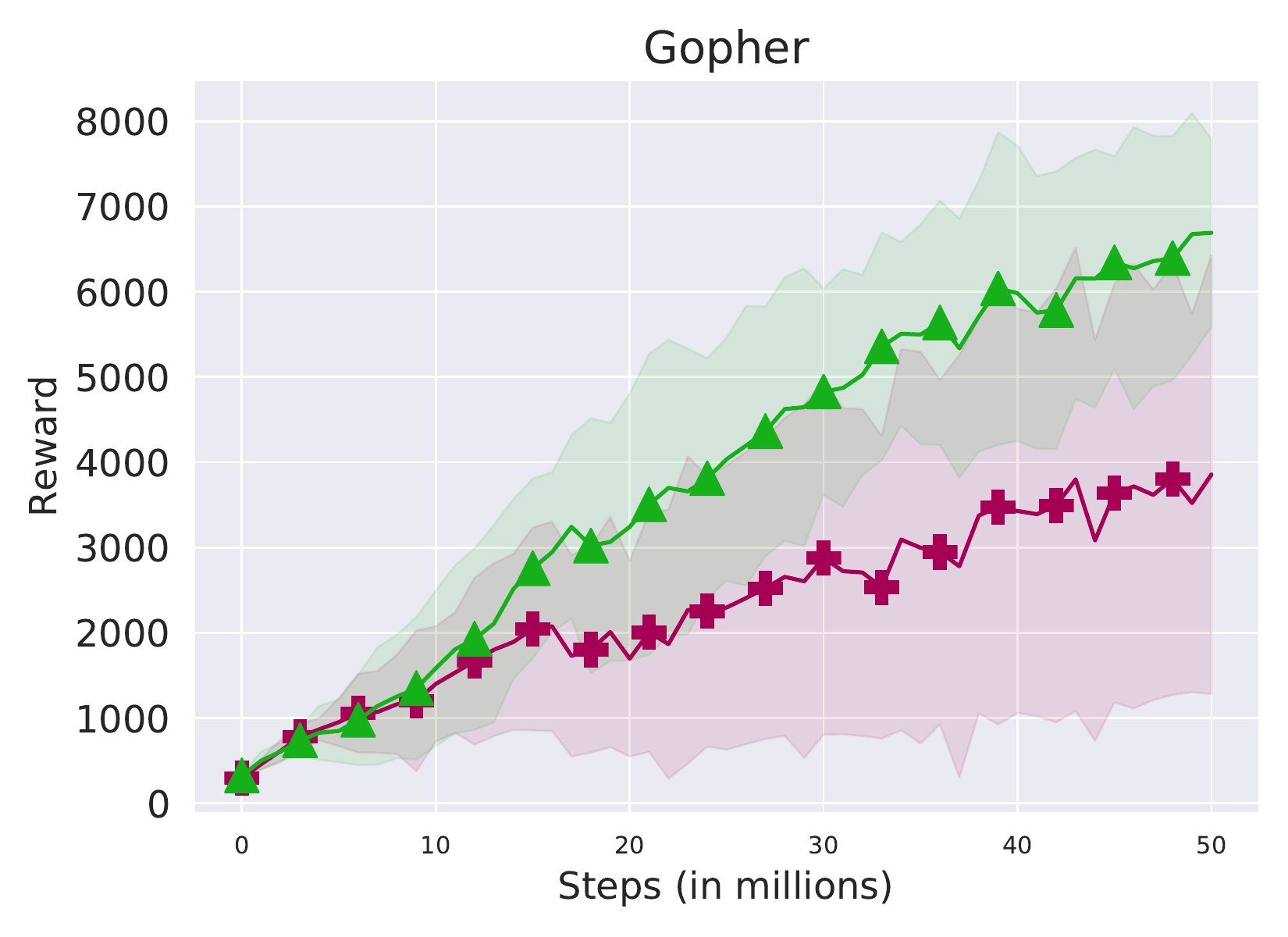}
         \caption{Gopher}
         \label{fig:lider-a3c-gopher}
     \end{subfigure}
     \begin{subfigure}[th]{0.48\textwidth}
         \centering
         \includegraphics[width=\textwidth]{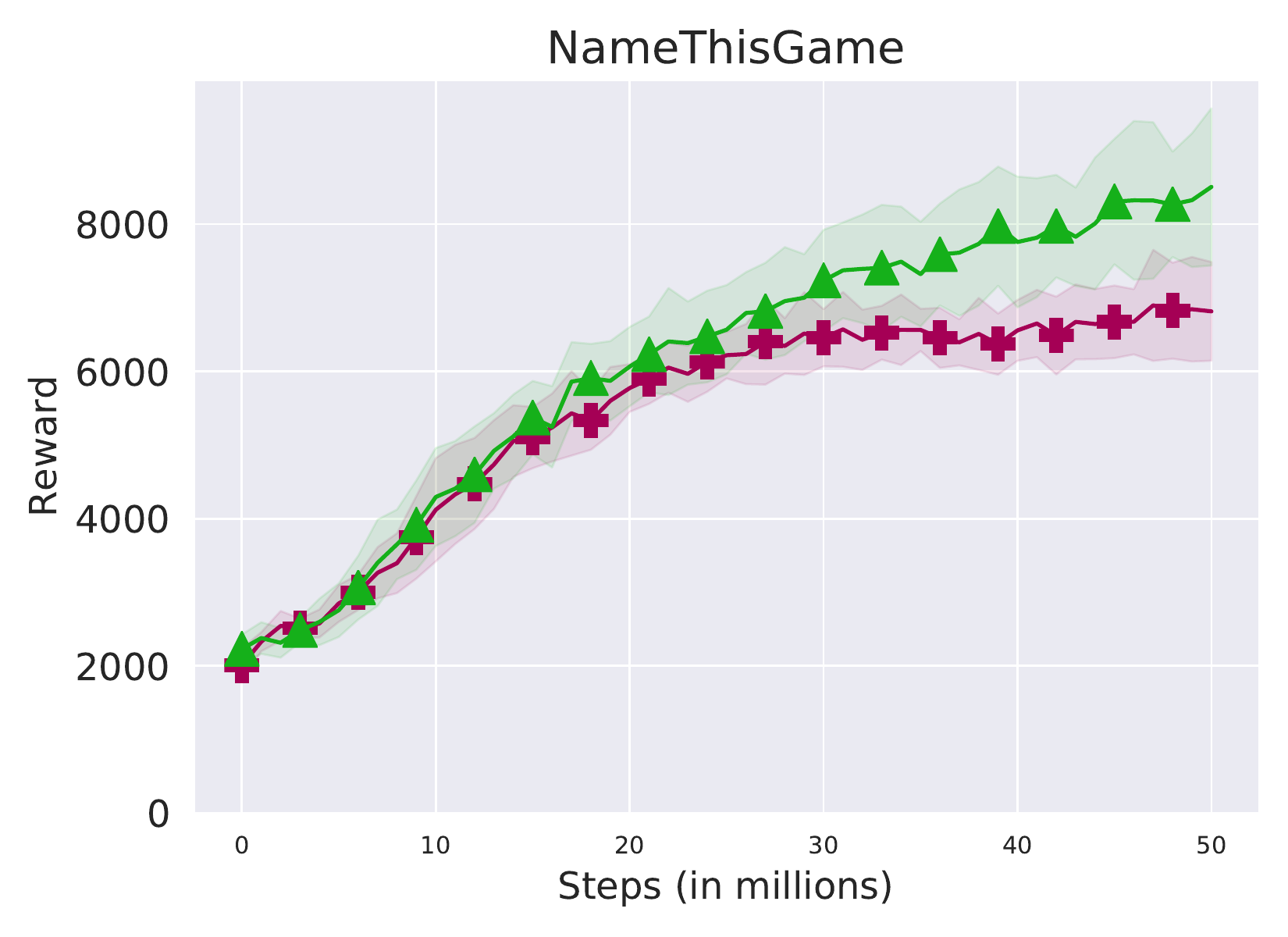}
         \caption{NameThisGame}
         \label{fig:lider-a3c-name}
     \end{subfigure}
     \begin{subfigure}[th]{0.48\textwidth}
         \centering
         \includegraphics[width=\textwidth]{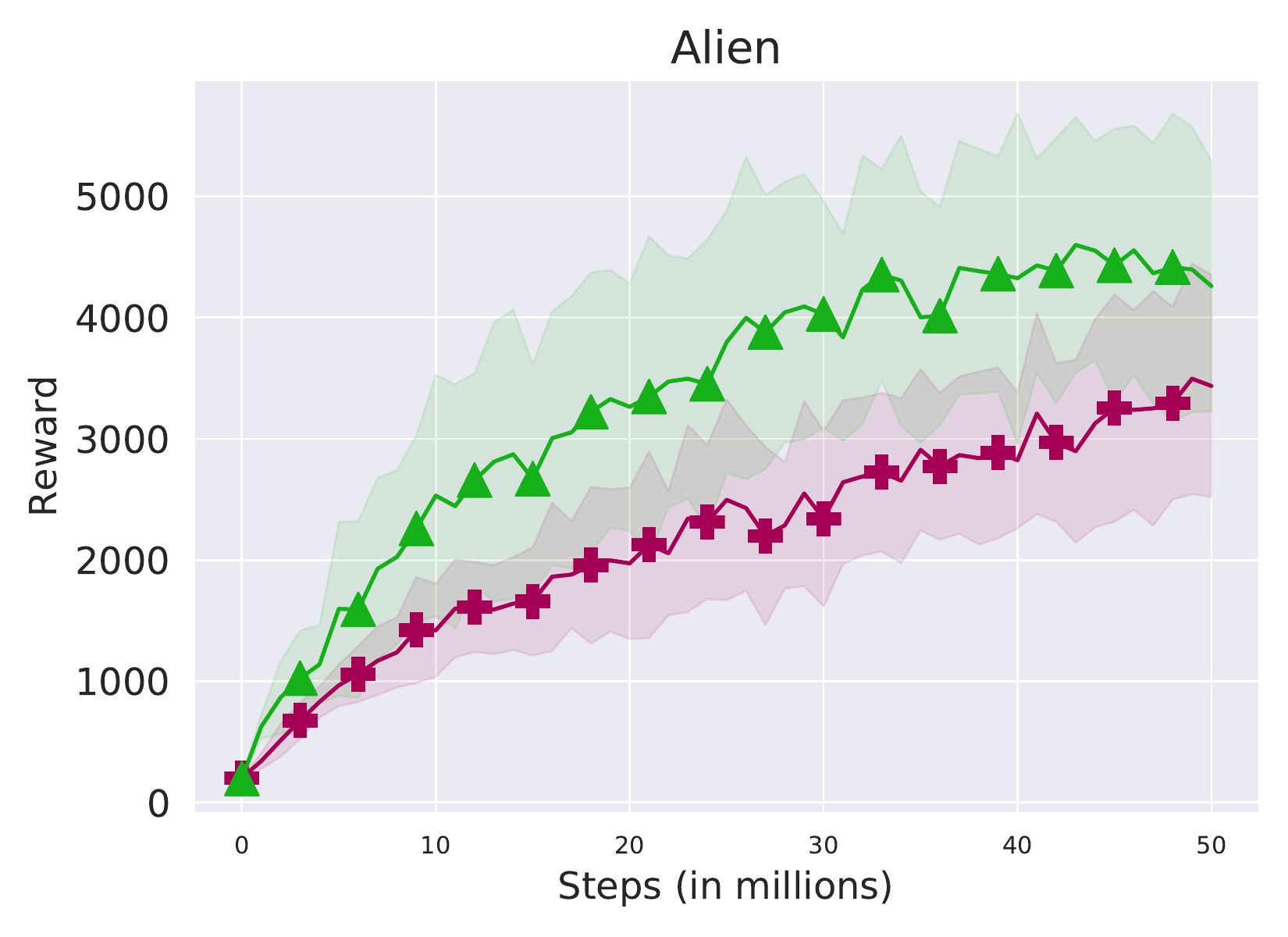}
         \caption{Alien}
         \label{fig:lider-a3c-alien}
     \end{subfigure}
     \begin{subfigure}[th]{0.48\textwidth}
         \centering
         \includegraphics[width=\textwidth]{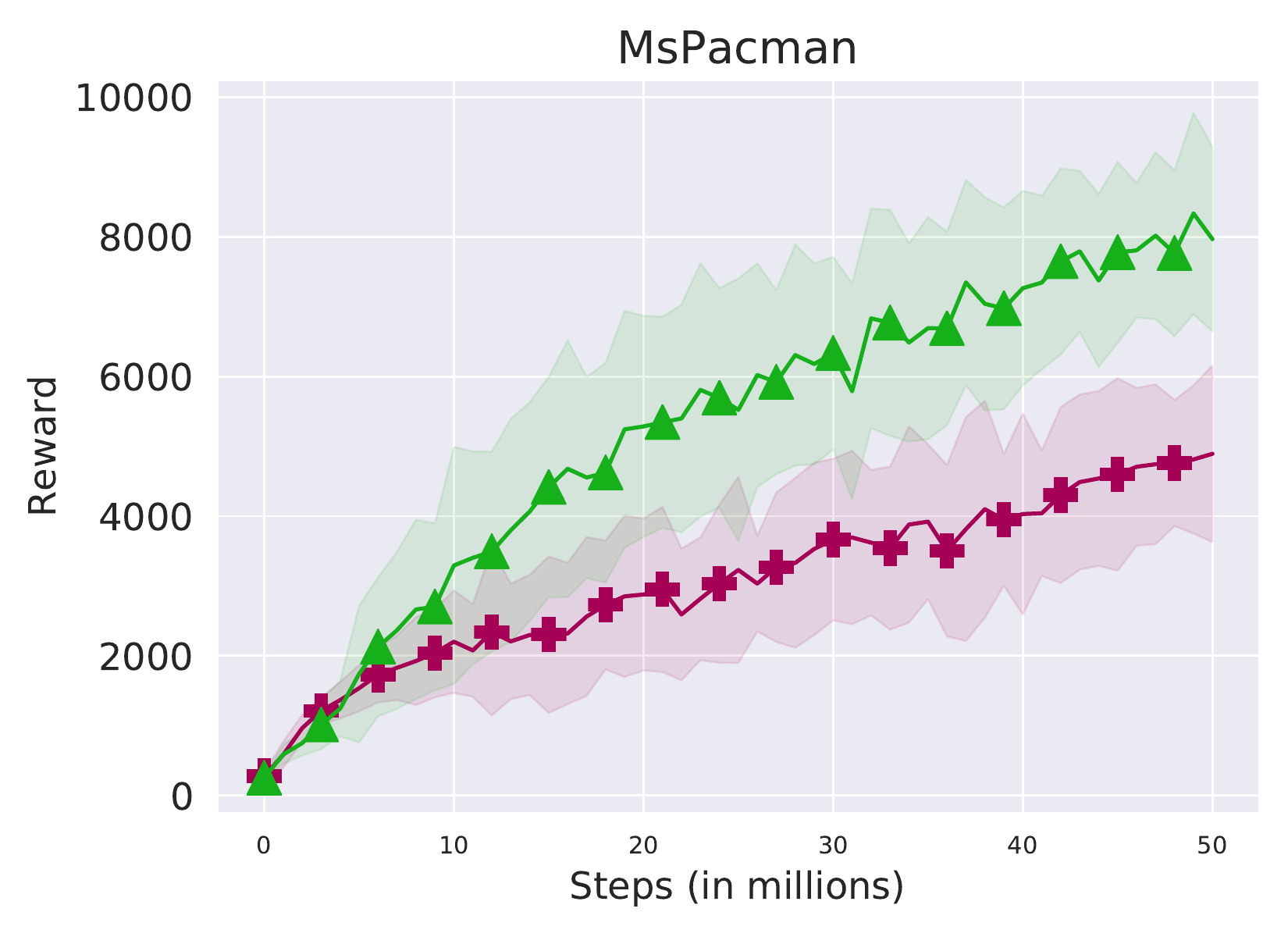}
         \caption{Ms.~Pac-Man}
         \label{fig:lider-a3c-mspacman}
     \end{subfigure}
     \begin{subfigure}[th]{0.48\textwidth}
         \centering
         \includegraphics[width=\textwidth]{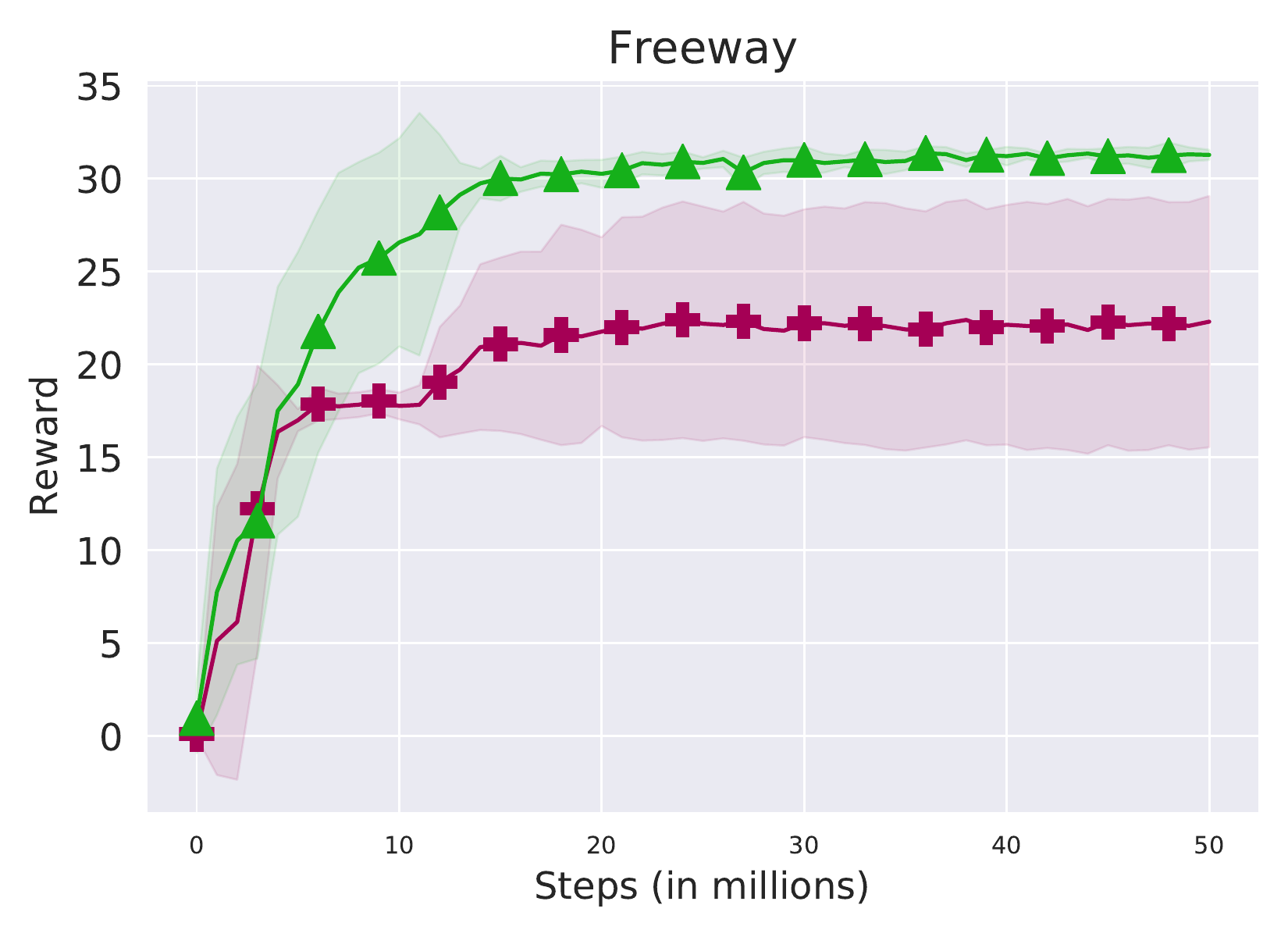}
         \caption{Freeway}
         \label{fig:lider-a3c-freeway}
     \end{subfigure}
     \begin{subfigure}[th]{0.48\textwidth}
         \centering
         \includegraphics[width=\textwidth]{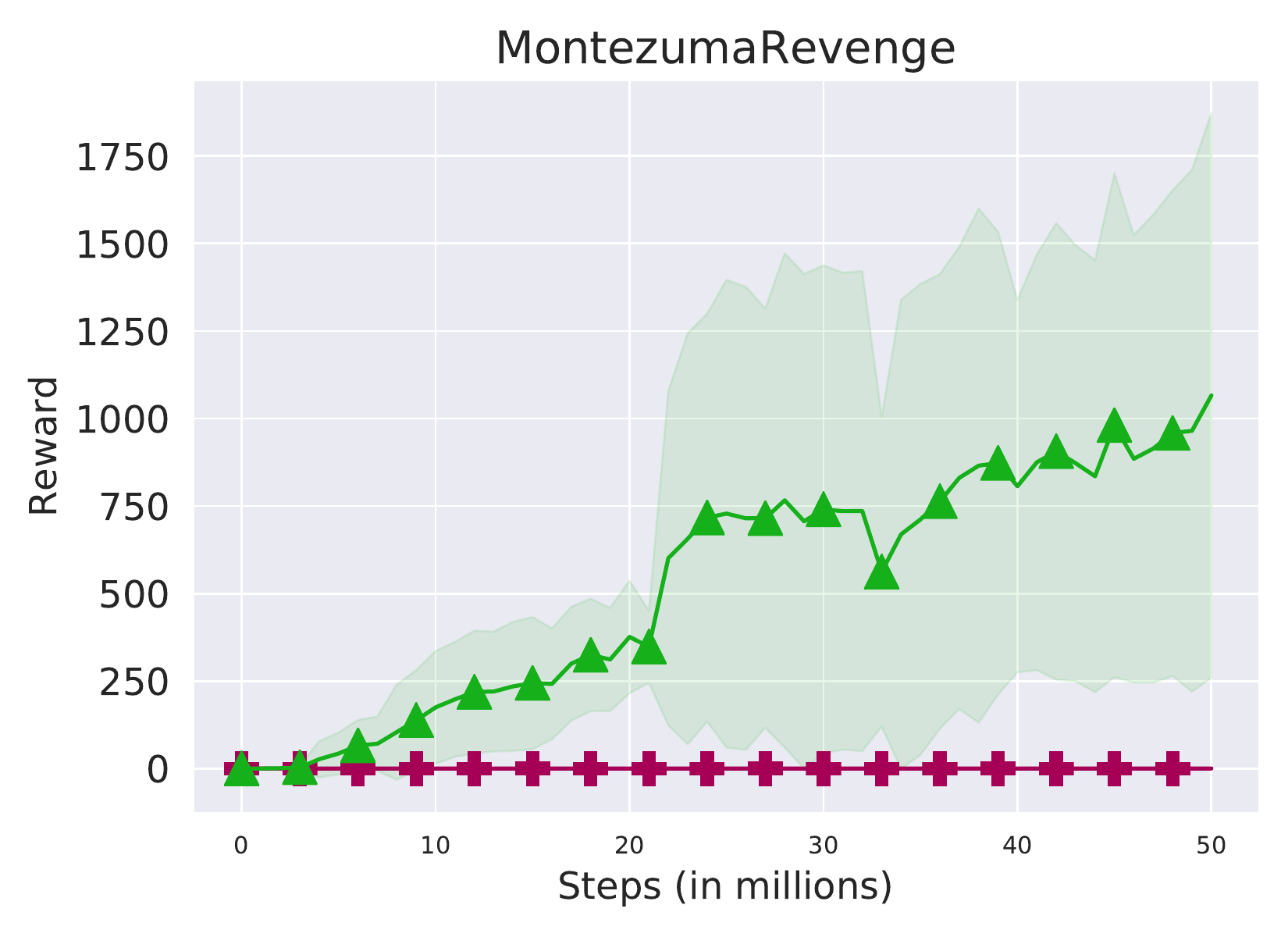}
         \caption{Montezuma's Revenge}
         \label{fig:lider-a3c-montezuma}
     \end{subfigure}
    \caption{LiDER performance is compared to A3CTBSIL on six Atari games. The x-axis is the total number of environmental steps: A3CTBSIL counts steps from 16 A3C workers, while LiDER counts steps from 15 A3C workers plus one refresher worker. The y-axis is the average testing score over eight trials; shaded regions show the standard deviation.} 
    \label{fig:lider-a3c}
\end{figure*}

First, we show that the agent's current policy can be effectively leveraged to refresh past experiences. Figure \ref{fig:lider-a3c} shows LiDER outperforms A3CTBSIL in all six games (averaged over eight trials); a one-tailed independent-samp-les t-test confirms statistical significance ($p \ll 0.001$, see Appendix \ref{sec:appe-ttest} for details of the t-tests).  
We train each trial for 50 million environmental steps. For every 1 million steps, we perform a test of 125,000 steps and report the average testing scores per episode (an episode ends when the agent loses all its lives).

We hypothesize that the performance improvement in the four dense reward games (Gopher, NameThisGame, Alien, and Ms.~Pac-Man) was because the likelihood for the refresher to encounter higher-return new trajectories is higher when rewards are dense. In addition, we observe in Ms.~Pac-Man that once the return and the action of a state have been refreshed, LiDER prefers to sample and reuse the newer rather than the older state-action-return transition from the same state, which could be another reason for the speed-up in learning---LiDER replays high-rewarding data mo-re frequently. We conduct a detailed analysis of LiDER's underlying behaviors in the next subsection that supports this hypothesis. 

LiDER also learns well in Freeway and Montezuma's Revenge, the two hard exploration, sparse reward games. In Freeway, the task is difficult because the agent only receives a non-zero reward after successfully crossing the highway. We hypothesize that LiDER is helpful in this case because the refresher can move the agent to an intermediate state (e.g., in the middle of the highway), which shortens the distance between the agent and the rewarding state, and thus allows the agent to learn faster. We can see LiDER's learning curve in Freeway from Figure \ref{fig:lider-a3c-freeway} that it consistently finds an optimal path after about 15 million steps of training (the standard deviation becomes negligible) but A3CTBSIL struggles to find a stable solution. The benefit of LiDER is evident particularly in Montezuma's Revenge. While A3CTBSIL fails to learn anything,\footnote{Note the performance in Montezuma's Revenge differs between A3CTBSIL \cite{de2019jointly} and the original SIL algorithm \cite{sil}---see the discussion in Appendix \ref{sec:appe-reducesil}.} LiDER is capable of reaching a reasonable score. Although the absolute performance of our method is not state-of-the-art, we have shown that LiDER is a light-weight addition to a baseline off-policy deep RL algorithm which helps improving performance even in the most difficult Atari games. 

\subsection{Analyses: why does LiDER help learning?}
\label{sec:analyses}
To understand why LiDER helps improve learning, in this section, we analyze the underlying behavior of LiD-ER from three perspectives. 
First, we look at the behaviors of the refresher worker since it is the novel component of LiDER. Inspecting whether the refresher worker can successfully generate higher return trajectories from past states, and 
how much better the refreshed data is compared to the older data, will give us insight into the quality of the data stored in buffer $\R$.  

Second, we examine the SIL worker in LiDER to reveal how SIL makes use of the refreshed data stored in buffer $\R$. Not only is it critical for the refresher to be able to generate better data, but the SIL worker must be able to effectively leverage these data to improve learning. 
The SIL worker should use data from buffer $\R$ more often for policy updates since the refreshed data is of a higher quality. 

Third, we compare the SIL worker between A3CTB-SIL and LiDER. In A3CTBSIL, the SIL worker samples only from one buffer; in LiDER, there are two buffers to sample from. It is thus interesting to investigate whether the SIL worker uses data from the two buffers differently. For example, if samples from buffer $\R$ have higher returns, we should see more samples with positive advantages in LiDER than in A3CTBSIL.  
The game of Ms.~Pac-Man is used as the running example for all analyses in this section.

\subsubsection{The refresher worker in LiDER}

\begin{figure}
     \centering
     \begin{subfigure}[th]{0.9\linewidth}
         \centering
         \includegraphics[width=0.95\linewidth]{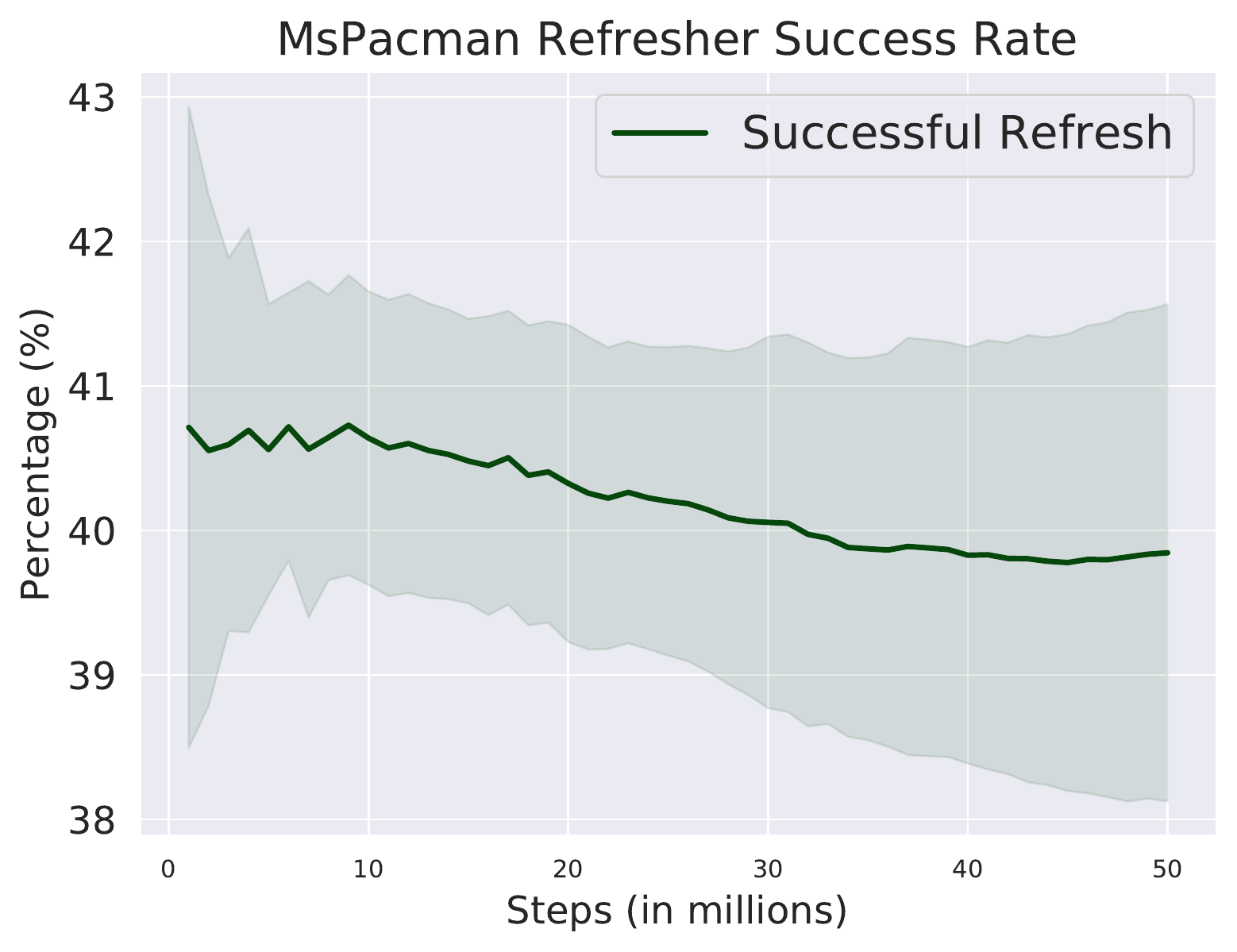}
         \caption{Refresher success rate}
         \label{fig:analysis-refresh-successrate}
     \end{subfigure}
     \begin{subfigure}[th]{0.9\linewidth}
         \centering
         \includegraphics[width=0.95\linewidth]{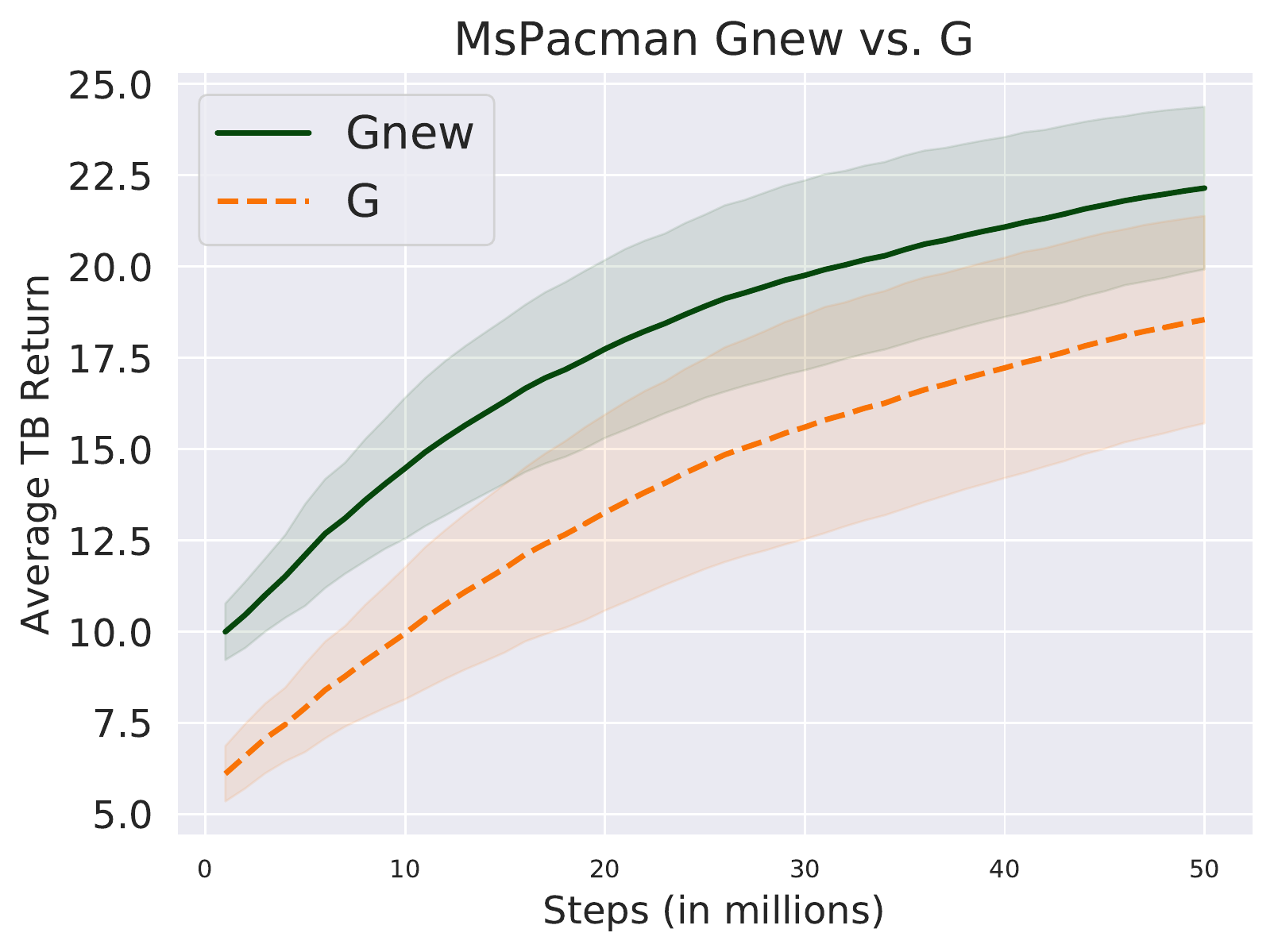}
         \caption{$\Gnew$ vs. $\G$}
         \label{fig:analysis-refresh-gnew}
     \end{subfigure}
    \caption{LiDER's refresher worker can consistently produce higher return trajectories. The x-axis is the total number of environmental steps. The y-axis value is averaged over eight trials; shaded regions show the standard deviation.} 
    \label{fig:analysis-refresh}
\end{figure}

First, we check two quantities of the refresher worker to get insight into the quality of the data it generated:
\begin{itemize}
    \item \emph{Success rate} (Figure \ref{fig:analysis-refresh-successrate}): how often can the refresher worker generate a better trajectory such that $\Gnew > \G$. 
    \item \emph{$\Gnew$ vs. $\G$} (Figure \ref{fig:analysis-refresh-gnew}): the average TB return $\Gnew$ compared to $\G$ for all successful refresh.
\end{itemize}

The \emph{success rate} is measured as the percentage of the number of successful rollouts over the total number of rollouts generated. Figure \ref{fig:analysis-refresh-successrate} shows that the success rate remains at approximately 40\%, indicating that the refresher is able to consistently produce higher return trajectories throughout the training. 

The improvement of the refreshed data over the older data can be measured by comparing $\Gnew$ to $\G$. $\G$ is the (TB) return of a state $S$ sampled from buffer $\D$; $\Gnew$ is the refreshed (TB) return of $S$. We record the value of $\Gnew$ and $\G$ for all successful rollouts then compute their average value. 
Figure \ref{fig:analysis-refresh-gnew} shows that $\Gnew$ is indeed higher than $\G$. 
Both measures from Figures \ref{fig:analysis-refresh-successrate} and \ref{fig:analysis-refresh-gnew} validate that the refresher worker is able to generate new trajectories with a higher return, and thus data in buffer $\R$ is expected to be of better quality than data in buffer $\D$. 

\subsubsection{The SIL worker in LiDER}
\label{sec:analyses-sil-lider}
\begin{figure}
     \centering
     \begin{subfigure}[th]{0.9\linewidth}
         \centering
         \includegraphics[width=0.95\linewidth]{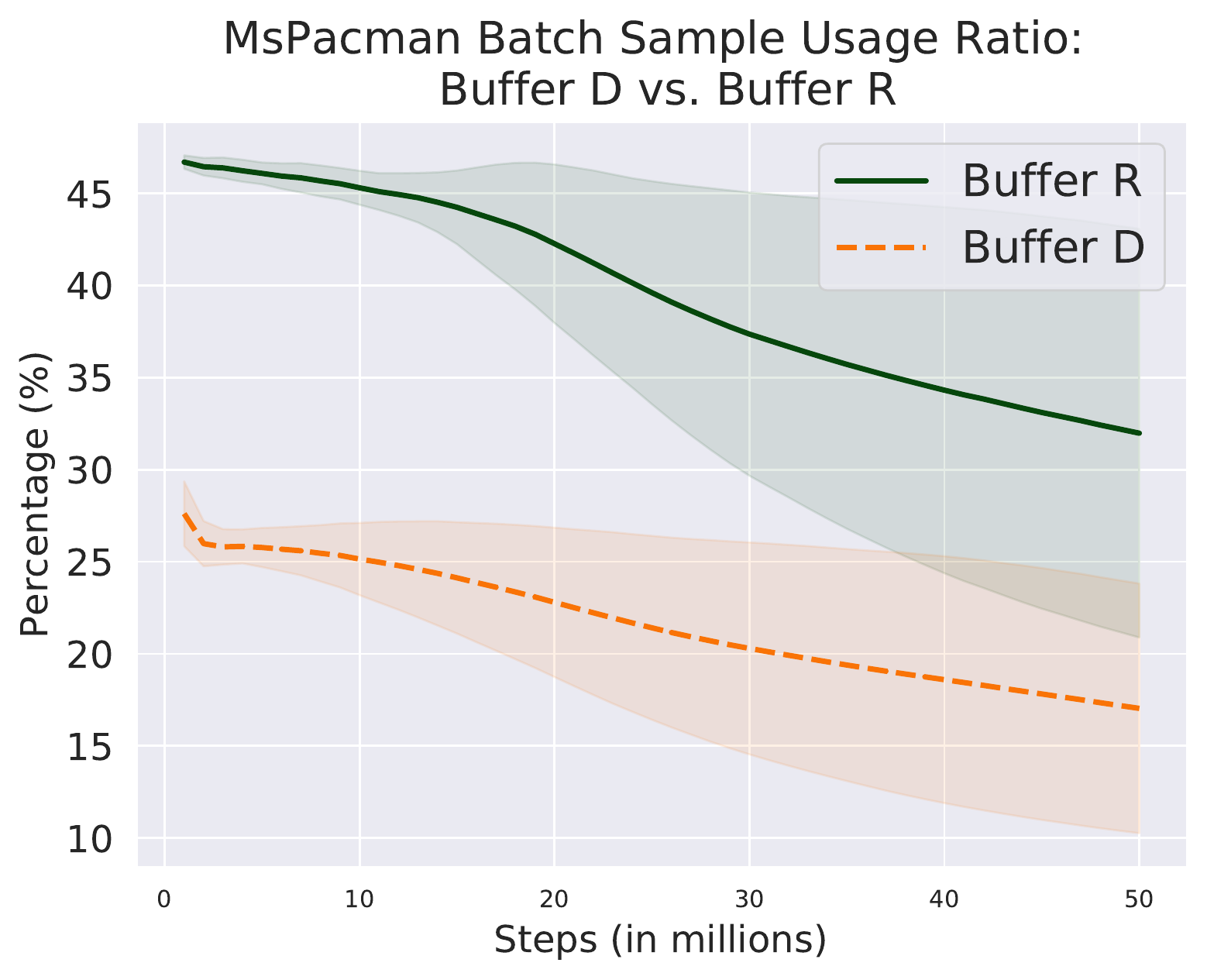}
         \caption{Batch sample usage ratio}
         \label{fig:analysis-batch-used-ratio}
     \end{subfigure}
     \begin{subfigure}[th]{0.9\linewidth}
         \centering
         \includegraphics[width=0.95\linewidth]{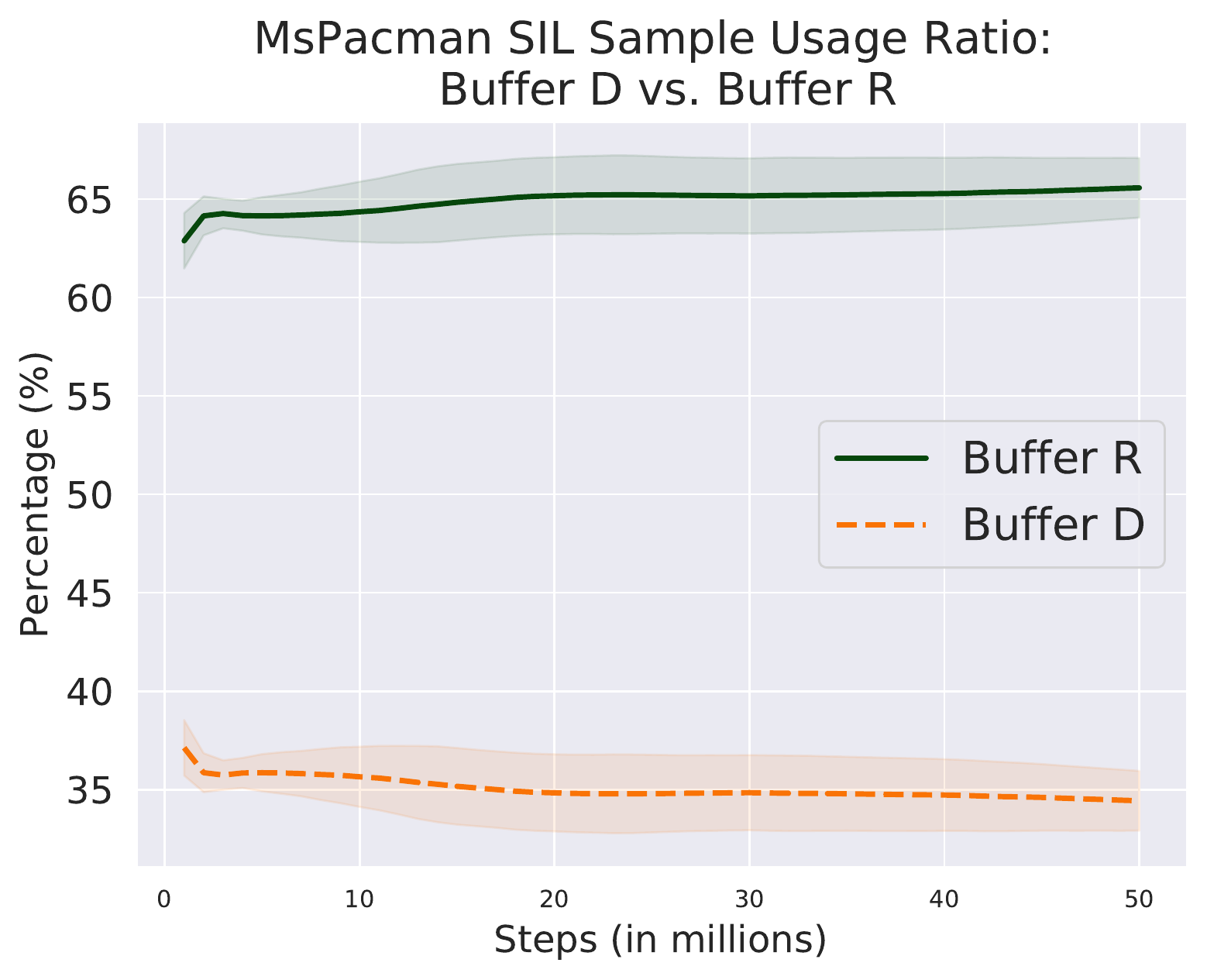}
         \caption{SIL sample usage ratio}
         \label{fig:analysis-sil-used-ratio}
     \end{subfigure}
     \begin{subfigure}[th]{0.9\linewidth}
         \centering
         \includegraphics[width=0.95\linewidth]{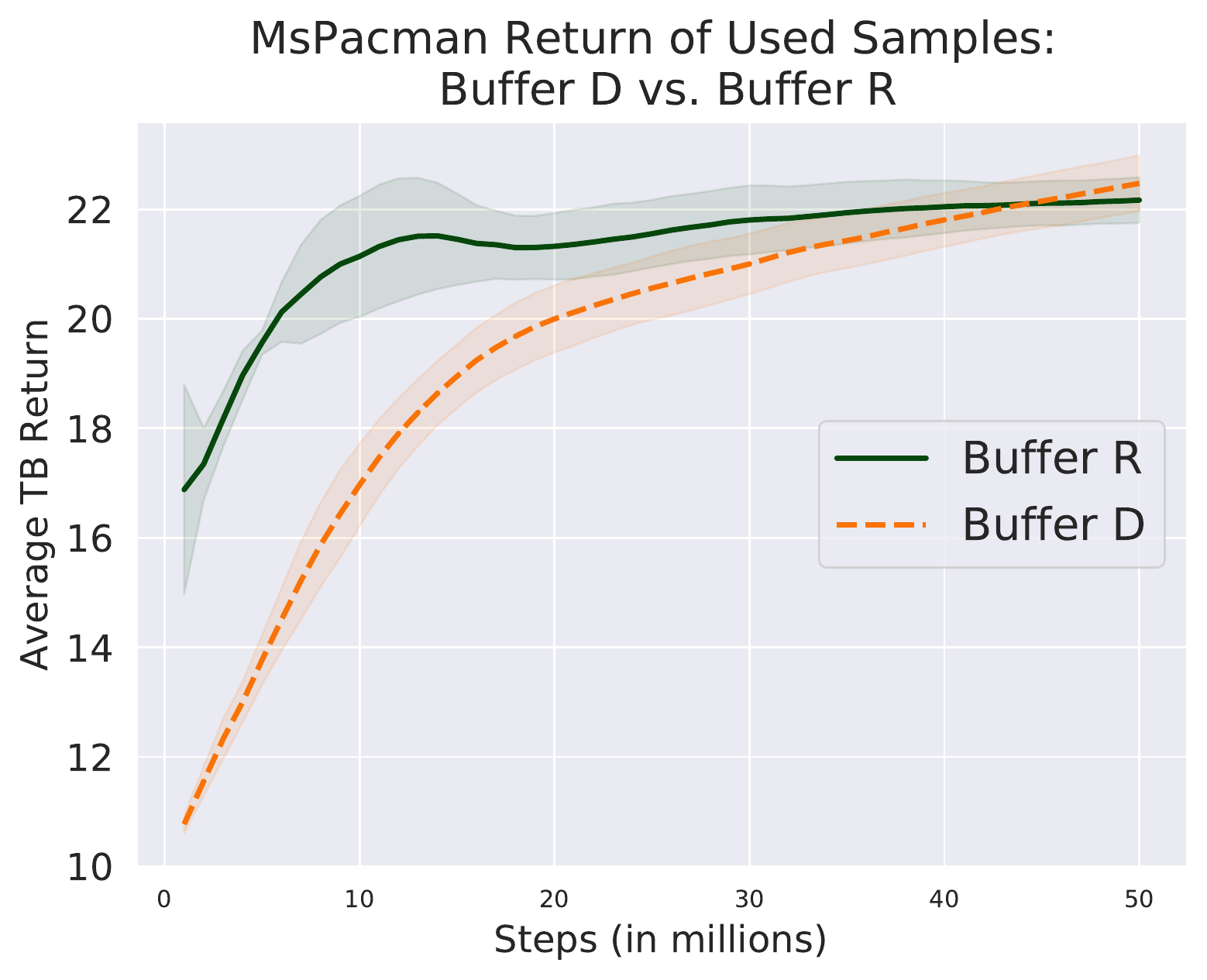}
         \caption{Return of used samples}
         \label{fig:analysis-sil-used-return}
     \end{subfigure}
     \caption{The SIL worker in LiDER leverages more refresher-generated data in buffer $\R$ than A3C-generated data in buffer $\D$ because refreshed data has higher returns. The x-axis is the total number of environmental steps. The y-axis value is averaged over eight trials; shaded regions show the standard deviation.} 
    \label{fig:analysis-sil}
\end{figure}

We have shown the refresher is able to generate higher return trajectories from past states. Next, we analyze the behavior of the SIL worker in LiDER to check whether it can effectively leverage these data. 
We inspect the following quantities for buffer $\D$ and buffer $\R$: 
\begin{itemize}
    \item \emph{Old samples used} (Table \ref{table:old-sample-used}): how many old samples were still used for SIL updates even after the sample has been refreshed to a newer return.
    \item \emph{Batch sample usage ratio} (Figure \ref{fig:analysis-batch-used-ratio}): for one batch of samples, how many samples taken from buffer $\D$ and buffer $\R$ were used for SIL updates (i.e., samples with positive advantages), respectively.
    \item \emph{SIL sample usage ratio} (Figure \ref{fig:analysis-sil-used-ratio}): for samples used for SIL updates, how many of them were taken from buffer $\D$ and buffer $\R$, respectively. 
    \item \emph{Return of used samples} (Figure \ref{fig:analysis-sil-used-return}): the return of samples used for SIL updates.  
\end{itemize}

As mentioned in Section \ref{sec:lider-a3c}, we hypothesize that once a state's return has been refreshed, LiDER tends not to reuse the older return. We investigate whether this hypothesis holds by counting how many older samples were used for SIL updates. 
Specifically, we assign a \emph{False} Boolean value to each state in buffer $\D$. Once a state has been sampled as the input to the refresher worker, we flip the Boolean to \emph{True} for that state. 
For each SIL update, we count the number of samples with Boolean \emph{True} and compute the ratio of \emph{old samples used} over the total number of samples used for that update. 

\begin{table}
\centering
\caption{Old samples used (\%) by the SIL worker in LiDER at 1, 25, and 50 million training steps. LiDER rarely reuses an older state after it has been refreshed. Results were averaged over eight trials.}
\label{table:old-sample-used}
\begin{tabular}{l|l|l|l}
\hline \hline
Steps (in millions)      & 1     & 25    & 50     \\ \hline \hline
Old samples used (\%)    & 0.0103 & 0.0053 & 0.0047 \\ \hline 
Standard deviation (\%) & 0.0029 & 0.0005 & 0.0003 \\ \hline \hline
\end{tabular}
\end{table}

We show the percentage of the old samples used at 1, 25, and 50 million steps of training in Table \ref{table:old-sample-used}. 
It can be seen that less than 0.01\% of the older samples were reused throughout training, and the reuse ratio keeps decreasing as training continues. This evidence validates our observation from Section \ref{sec:lider-a3c} that LiDER replays the higher-return data in buffer $\R$ more frequently than the lower-return data in buffer $\D$. 


Recall that the SIL worker only uses samples with positive advantages (i.e., samples that ``pass'' the max operator) to update the policy. The percentage of such positive samples in buffer $\D$ and $\R$, respectively, can tell us which buffer has higher quality data.  
We call this percentage ``sample usage ratio'' and measure two type of ratios: \emph{batch sample usage ratio} and \emph{SIL sample usage ratio}. 

\emph{Batch sample usage ratio} measures how many positive samples are from buffer $\D$ and $\R$, respectively, over one batch of samples (the batch size is 32 in our experiments). For example, in one batch of 32 samples, suppose that there were 16 samples with positive advantages that were taken from buffer $\R$, and that there were 8 positive samples from buffer $\D$. The batch sample usage ratio for buffer $\R$ is computed as $\frac{16}{32}=50\%$, and for buffer $\D$ is computed as $\frac{8}{32}=25\%$. Figure \ref{fig:analysis-batch-used-ratio} shows that, on average, there are more positive samples from buffer $\R$ than from buffer $\D$ in one batch of data. This trend indicates that buffer $\R$'s samples are more useful for SIL updates throughout training, and are thus of a higher quality than samples in buffer $\D$. 

\emph{SIL sample usage ratio} also measures the ratio of positive samples for each buffer, but it is computed over the total number of positive samples instead of the entire batch. For example, suppose that in one batch of 32 samples, there were 24 with positive advantages. 18 out of the 24 samples come from buffer $\R$ and the other 6 come from buffer $\D$. The SIL sample usage ratio for buffer $\R$ is computed as $\frac{18}{24}=75\%$, and buffer $\D$'s ratio is $\frac{6}{24}=25\%$. Figure \ref{fig:analysis-sil-used-ratio} shows that buffer $\R$ always has a higher proportion of positive samples than buffer $\D$ among all positive samples. 
Both Figure \ref{fig:analysis-batch-used-ratio} and Figure \ref{fig:analysis-sil-used-ratio} indicate that the SIL worker is able to effectively leverage the refreshed samples for policy updates. 

We can also confirm that the \emph{return of used samples} from buffer $\R$ is indeed higher than those from buffer $\D$. Similar to how $\Gnew$ and $\G$ were compared, we compare the average (TB) return of all samples used for SIL updates between buffer $\R$ and buffer $\D$. Figure \ref{fig:analysis-sil-used-return} shows that data in buffer $\R$ 
has a higher return than data in buffer $\D$ 
during earlier stages of training. 
The two values then become similar at the end of training---an expected observation as the agent has learned a stable policy.

\subsubsection{The SIL worker in A3CTBSIL vs. LiDER}

Lastly, we compare the SIL worker between A3CTBSIL and LiDER. We have seen in the previous subsection that the SIL worker in LiDER always prefers to use samples in buffer $\R$, which allows more, and higher-quality, data to be leveraged for policy updates. It is thus interesting to inspect what kind of data has been used in A3CTBSIL, and whether the data is better or worse than the data in LiDER. 
As done in the previous subsection, we examine the \emph{batch sample usage ratio} and \emph{return of used samples} for A3CTBSIL and LiDER. 

For A3CTBSIL, the batch sample usage ratio and return of used samples are measured in buffer $\D$ only. For LiDER, we make a small modification that instead of quantifying the two buffers separately, we treat them as one buffer and measure their values together. For example, to compute the batch sample usage ratio, suppose that 6 buffer $\D$ samples and 18 buffer $\R$ samples were used for a SIL update, the total batch sample usage ratio for LiDER would be $\frac{(18+6)}{32}=75\%$. 

\begin{figure}[t]
 \centering
     \begin{subfigure}[th]{0.9\linewidth}
         \centering
         \includegraphics[width=0.95\linewidth]{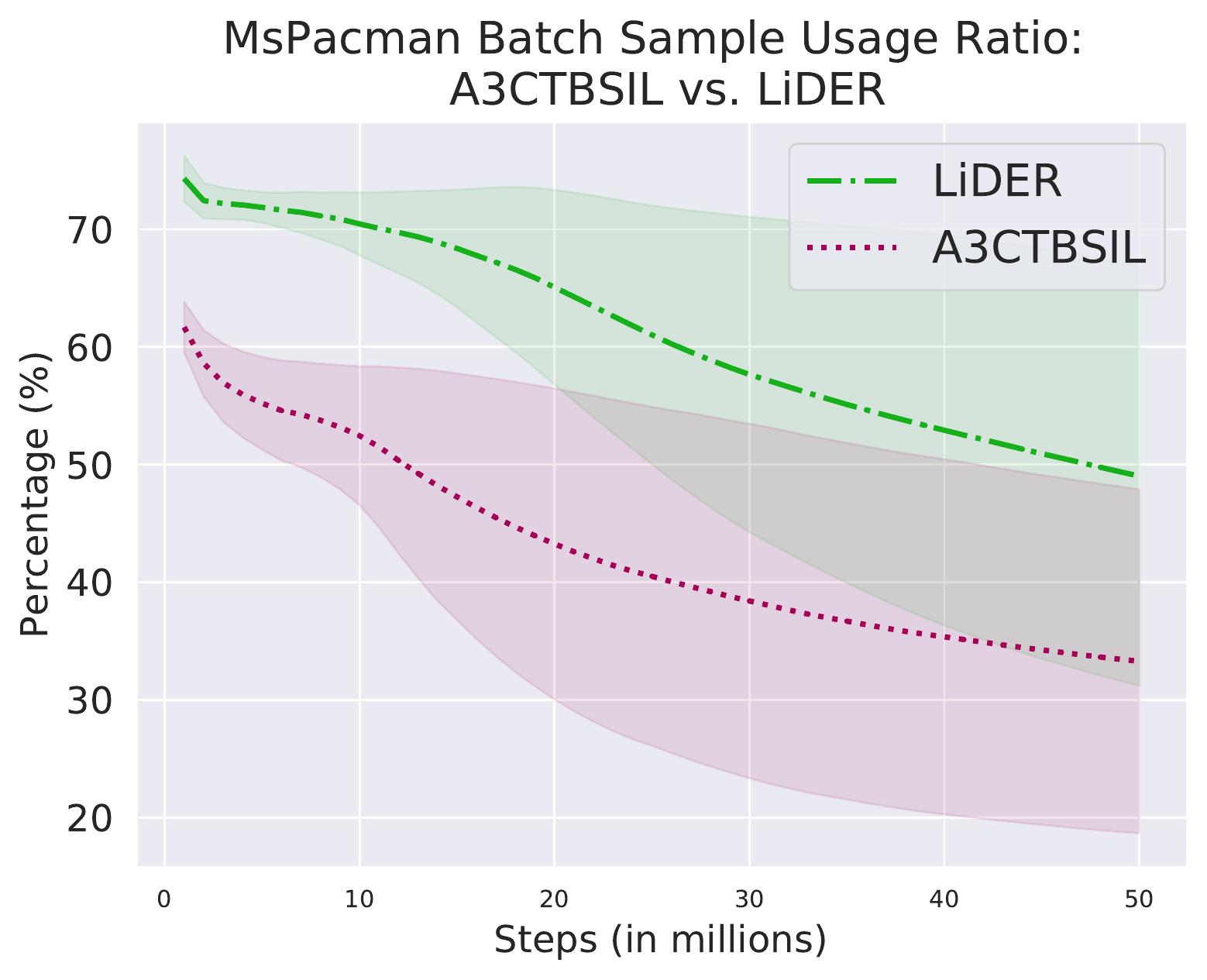}
         \caption{Total sample usage ratio}
         \label{fig:analysis-base-lider-used-ratio}
     \end{subfigure}
     \begin{subfigure}[th]{0.9\linewidth}
         \centering
         \includegraphics[width=0.95\linewidth]{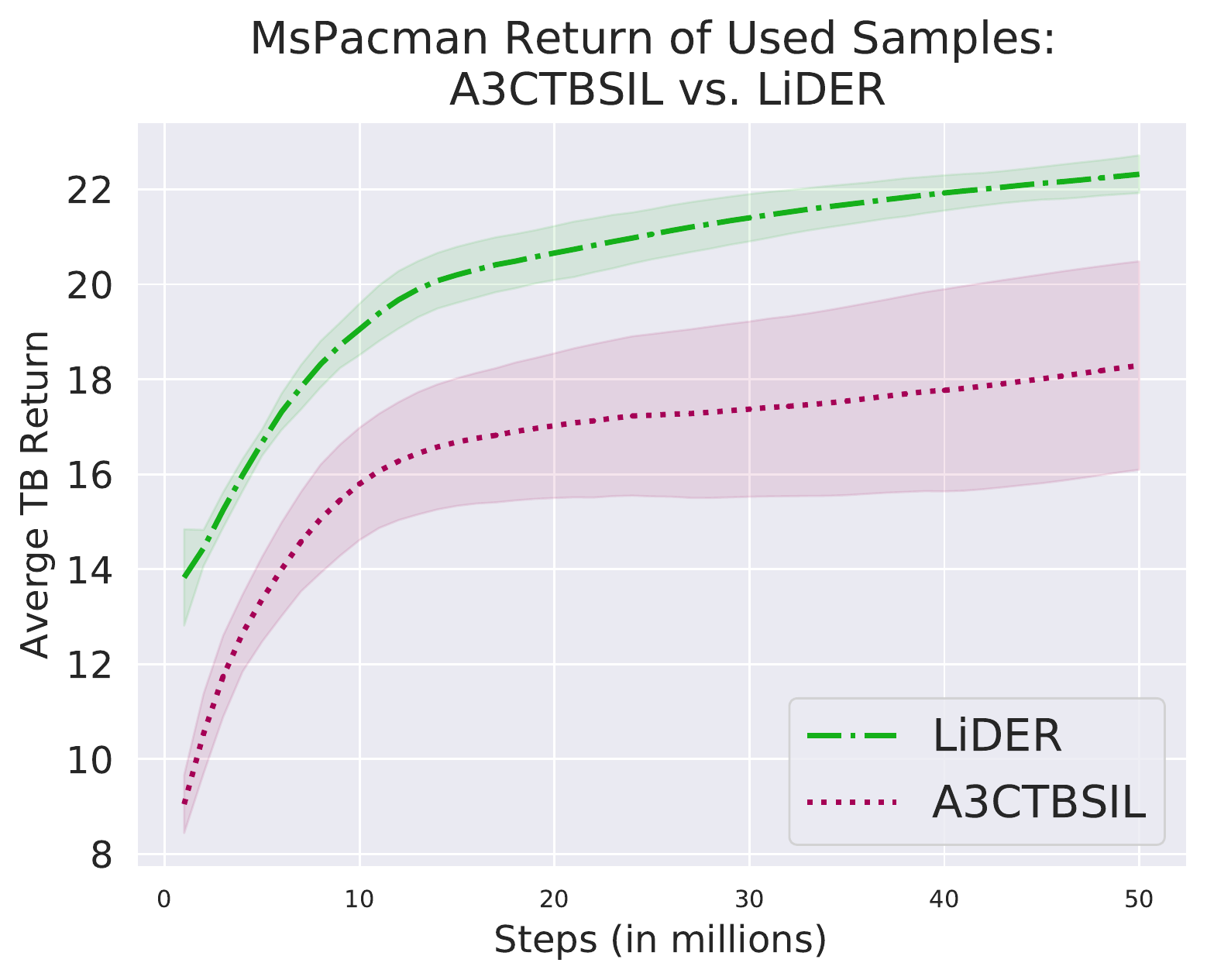}
         \caption{Return of used samples}
         \label{fig:analysis-base-lider-used-return}
     \end{subfigure}
    \caption{Comparing the SIL worker between A3CTBSIL and LiDER. LiDER leverages more and better quality data for policy updates than A3CTBSIL. The x-axis is the total number of environmental steps. The y-axis value is averaged over eight trials; shaded regions show the standard deviation.} 
    \label{fig:analysis-base-lider}
\end{figure}

Figure \ref{fig:analysis-base-lider-used-ratio} shows that, LiDER has a higher overall batch sample usage ratio than A3CTBSIL.  
We can also confirm the return of used samples is always higher in LiDER than in A3CTBSIL (Figure \ref{fig:analysis-base-lider-used-return}). This observation indicates that not only can the refresher generate higher return trajectories, but these trajectories are also effectively leveraged by the SIL worker. Both factors contribute to the performance improvement of LiDER over A3CTBSIL. 

In summary, our analyses show that 1) the refresher can consistently generate good trajectories during training; 2) the SIL worker of LiDER can effectively leverage these good trajectories by sampling from buffer $\R$; and 3) when compared to A3CTBSIL, LiDER performs policy updates with more and higher quality data. All three components contribute to the performance improvement of LiDER over A3CTBSIL. 


\section{Ablation studies}
\label{sec:ablation}

We have shown that LiDER can effectively leverage knowledge from the agent's current policy. In this section, we perform several ablation studies to further validate our design choices.

\subsection{How does the quality of refresher-generated data affect learning?}
As shown in Figure \ref{fig:analysis-refresh-gnew}, LiDER increases the overall data quality because we only use new data when it obtains a higher reward than the old data, i.e., $\Gnew > \G$. We show that it is important to store the refresher-generated experiences and use them to update the global policy only if those experiences are better, i.e., when the new return $\Gnew$ computed from the refresher experience is higher than the return $\G$ that the agent previously obtained. This condition ensures that the data in buffer $\R$ is of a higher quality than that in buffer $\D$. Intuitively, LiDER goes back in time to test if its current self can perform better than before and only provide help \emph{where it can}. To validate 
the importance of this condition, we conduct an experiment in which the refresher adds \emph{all} new experiences to buffer $\R$, i.e., without the $\Gnew > \G$ condition, to check if doing so leads to decreased performance. We denote this experiment as LiDER-AddAll. 

\subsection{How does the buffer architecture affect learning?} 
The other important design choice of LiDER is the two-buffer architecture: buffer $\D$ stores A3C-generated data and buffer $\R$ stores refresher-generated data. 
One hypothesis could be that LiDER performs better simply because the buffer size is doubled and more experiences can be replayed (e.g., \citet{CER} have shown that buffer size can affect learning). We conduct an experiment to show that simply increasing the size of a single buffer does not provide the same performance improvement as LiDER. 
We modify LiDER to have only buffer $\D$ and double its size from $10^5$ to 2$\times10^5$; both A3C-generated and refresher-generated data are stored in buffer $\D$. Prioritized sampling still takes a batch of 32 samples from buffer $\D$ as the input to the SIL worker, but without using the temporary buffer. We denote this experiment as LiDER-OneBuffer. 


\subsection{How does the sampling ratio affect learning?} 
LiDER samples from buffer $\D$ and $\R$ in a flexible manner as described in Section \ref{sec:lider}, 
and in Section \ref{sec:analyses-sil-lider} we have shown the samples from buffer $\R$ are more likely to be used for learning because they have higher returns. The question then arises, ``Should the agent always sample from buffer $\R$ since the refresher-generated data is better?'' We conduct an experiment in which the agent only samples from buffer $\R$; a batch of 32 samples are sampled with priority from buffer $\R$ as the input of the SIL worker, but without using the temporary buffer. Note that although we do not sample from buffer $\D$, we still keep it in the architecture since the refresher worker needs to randomly select a past state from buffer $\D$ to perform the refresh. We denote this experiment as LiDER-SampleR.\footnote{Note that the baseline A3CTBSIL represents the scenario of SampleD, i.e., always sample from buffer $\D$.} 

\begin{figure*}
     \centering
     \begin{subfigure}[th]{0.35\textwidth}
         \centering
         \includegraphics[width=\textwidth]{lider-a3c-legend-final.png}
     \end{subfigure}
     \hspace{0.5\textwidth} 
     \begin{subfigure}[th]{0.8\textwidth}
         \centering
         \includegraphics[width=\textwidth]{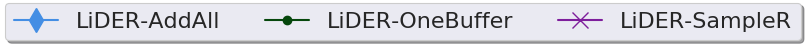}
     \end{subfigure}
     \vspace{10pt}
     \begin{subfigure}[th]{0.48\textwidth}
         \centering
         \includegraphics[width=\textwidth]{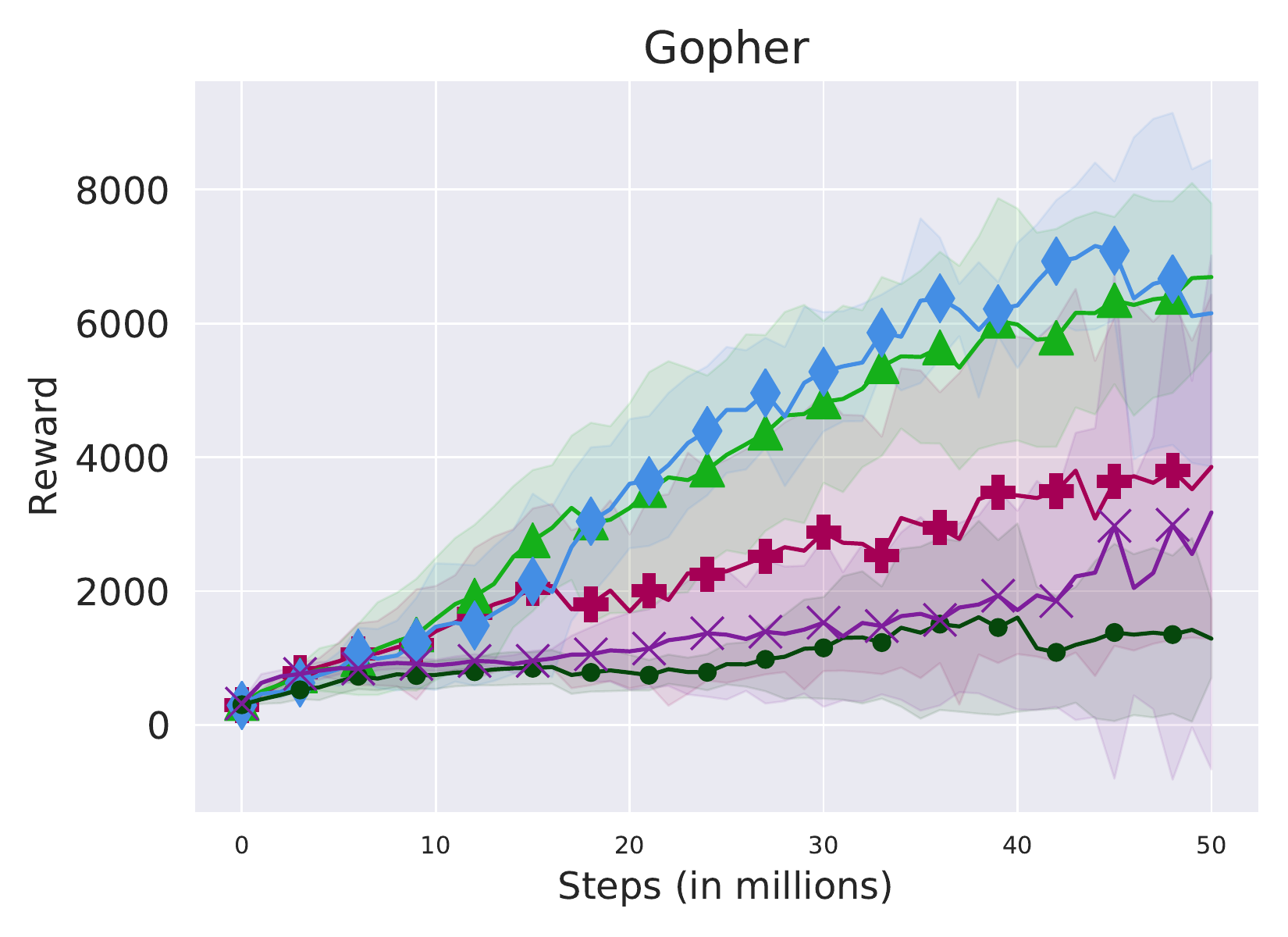}
         \caption{Gopher}
         \label{fig:ablation-gopher}
     \end{subfigure}
     \begin{subfigure}[th]{0.48\textwidth}
         \centering
         \includegraphics[width=\textwidth]{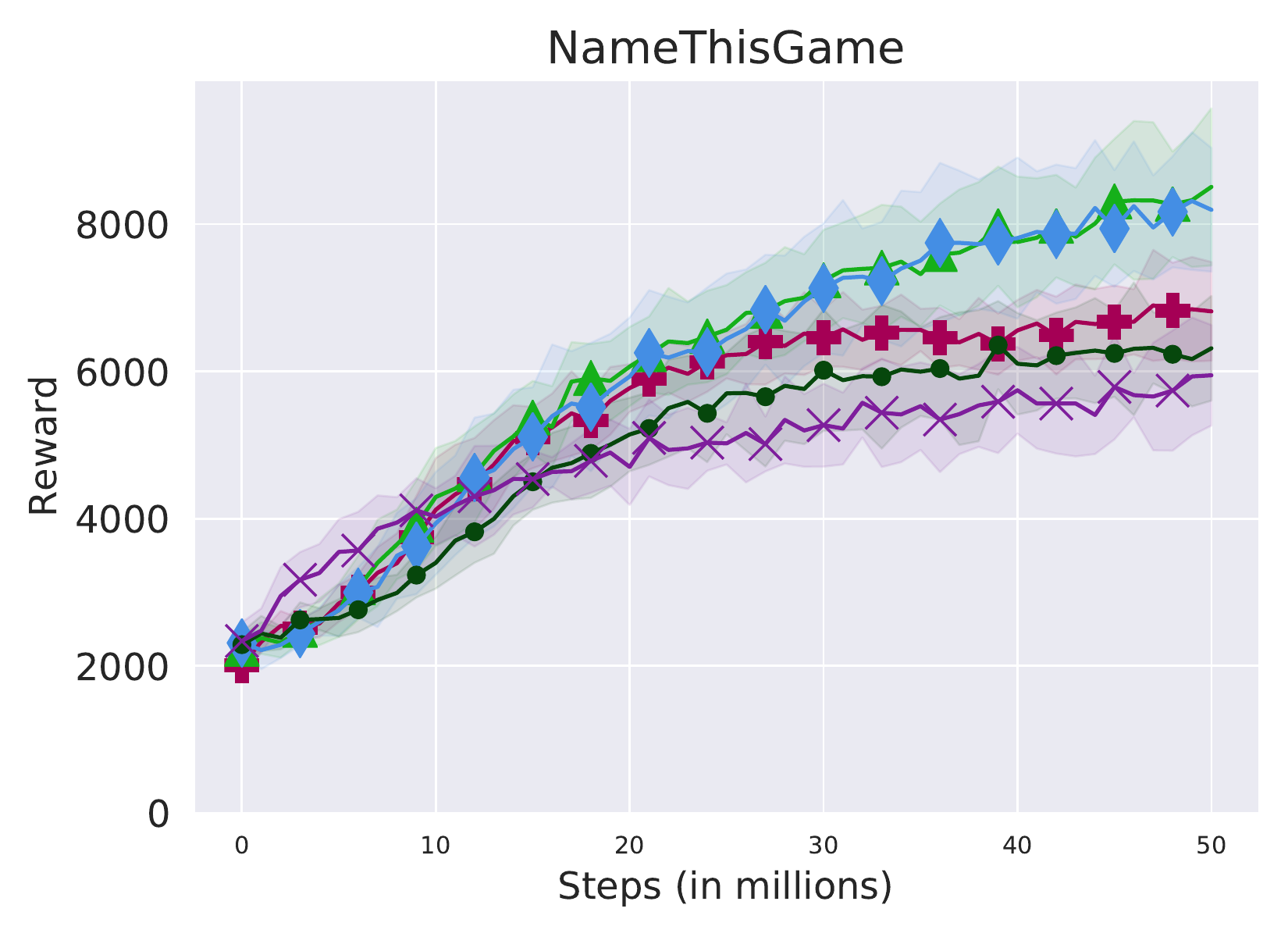}
         \caption{NameThisGame}
         \label{fig:ablation-name}
     \end{subfigure}
     \begin{subfigure}[th]{0.48\textwidth}
         \centering
         \includegraphics[width=\textwidth]{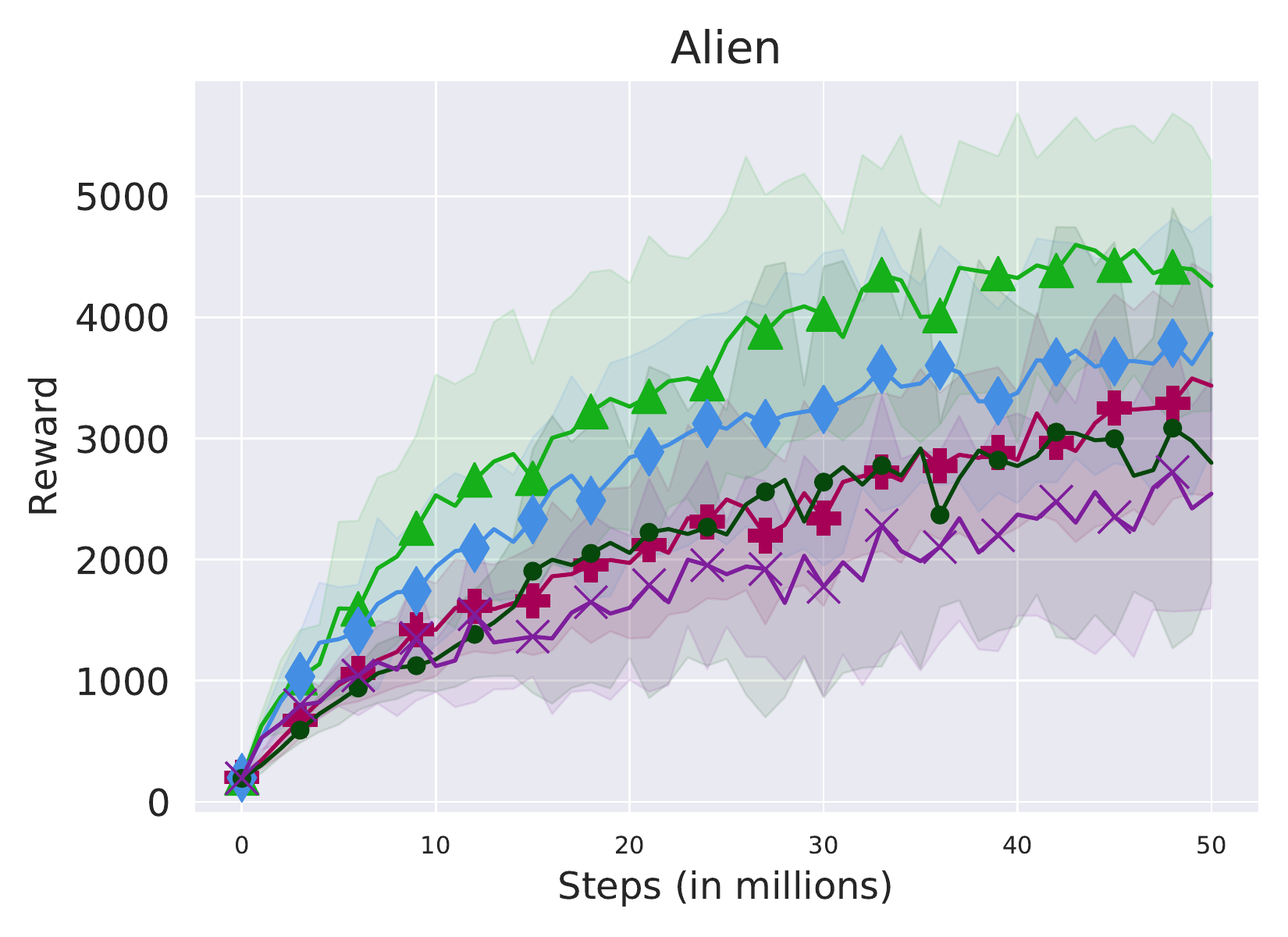}
         \caption{Alien}
         \label{fig:ablation-alien}
     \end{subfigure}
     \begin{subfigure}[th]{0.48\textwidth}
         \centering
         \includegraphics[width=\textwidth]{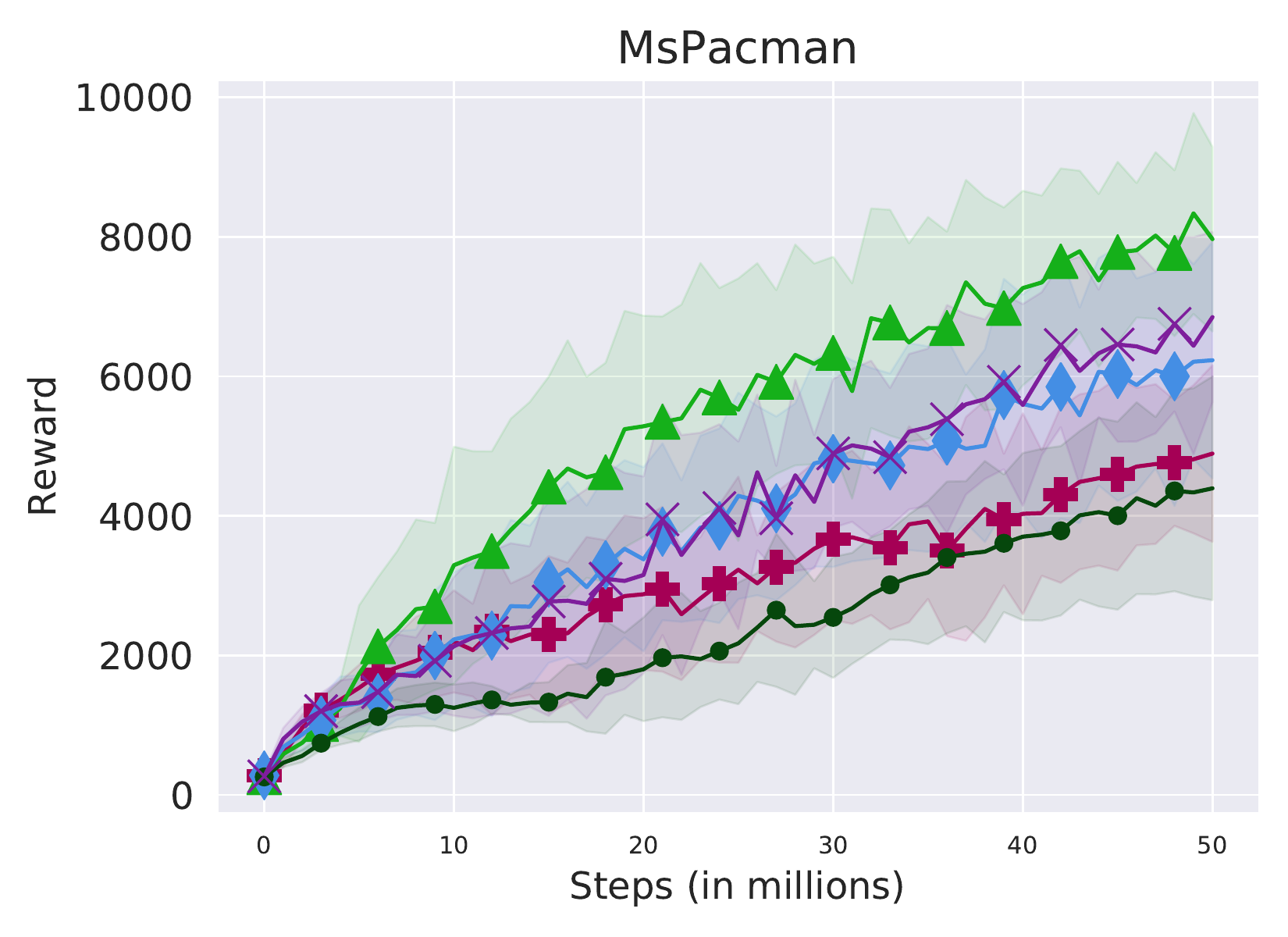}
         \caption{Ms.~Pac-Man}
         \label{fig:ablation-mspacman}
     \end{subfigure}
     \begin{subfigure}[th]{0.48\textwidth}
         \centering
         \includegraphics[width=\textwidth]{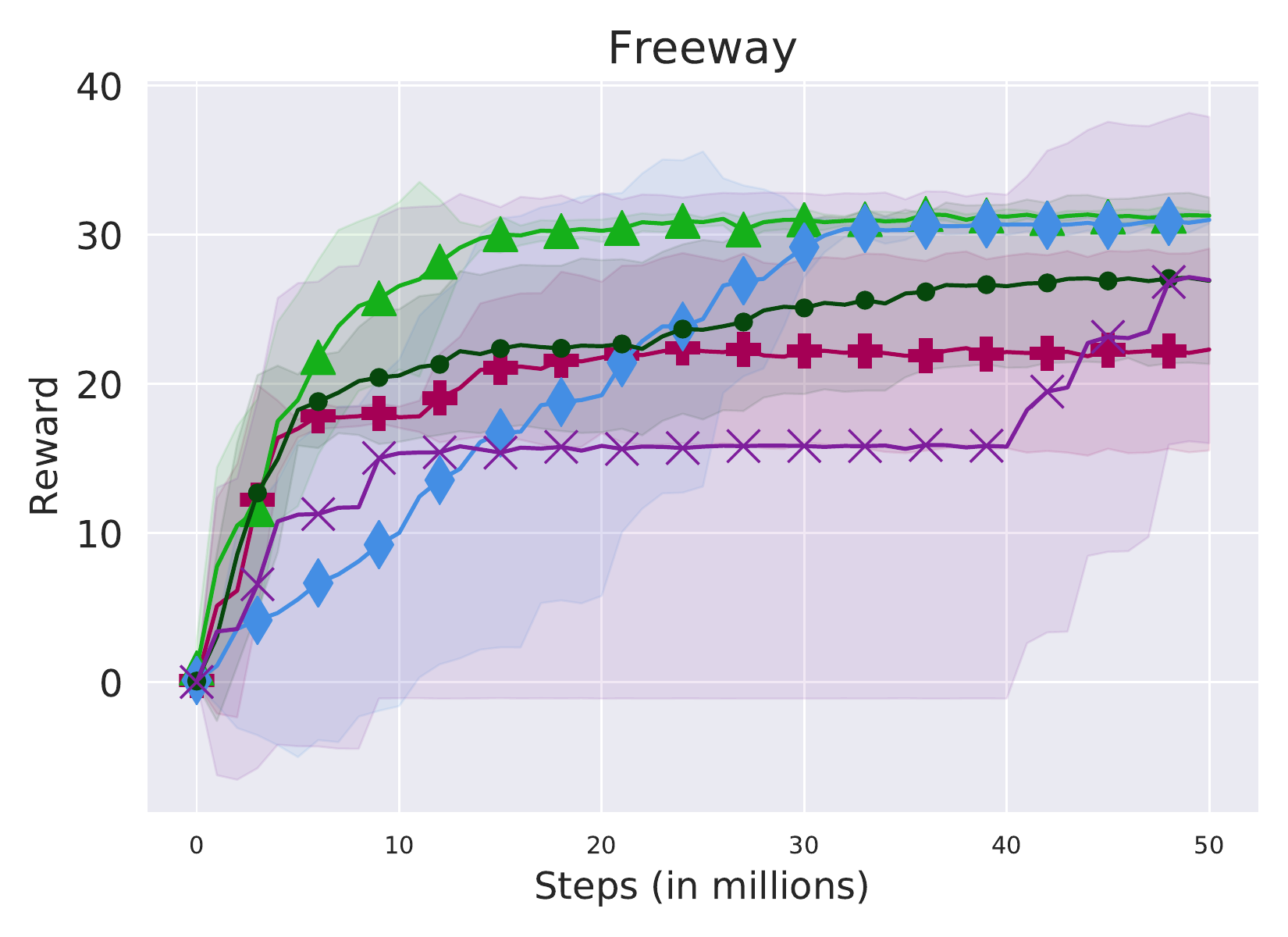}
         \caption{Freeway}
         \label{fig:ablation-freeway}
     \end{subfigure}
     \begin{subfigure}[th]{0.48\textwidth}
         \centering
         \includegraphics[width=\textwidth]{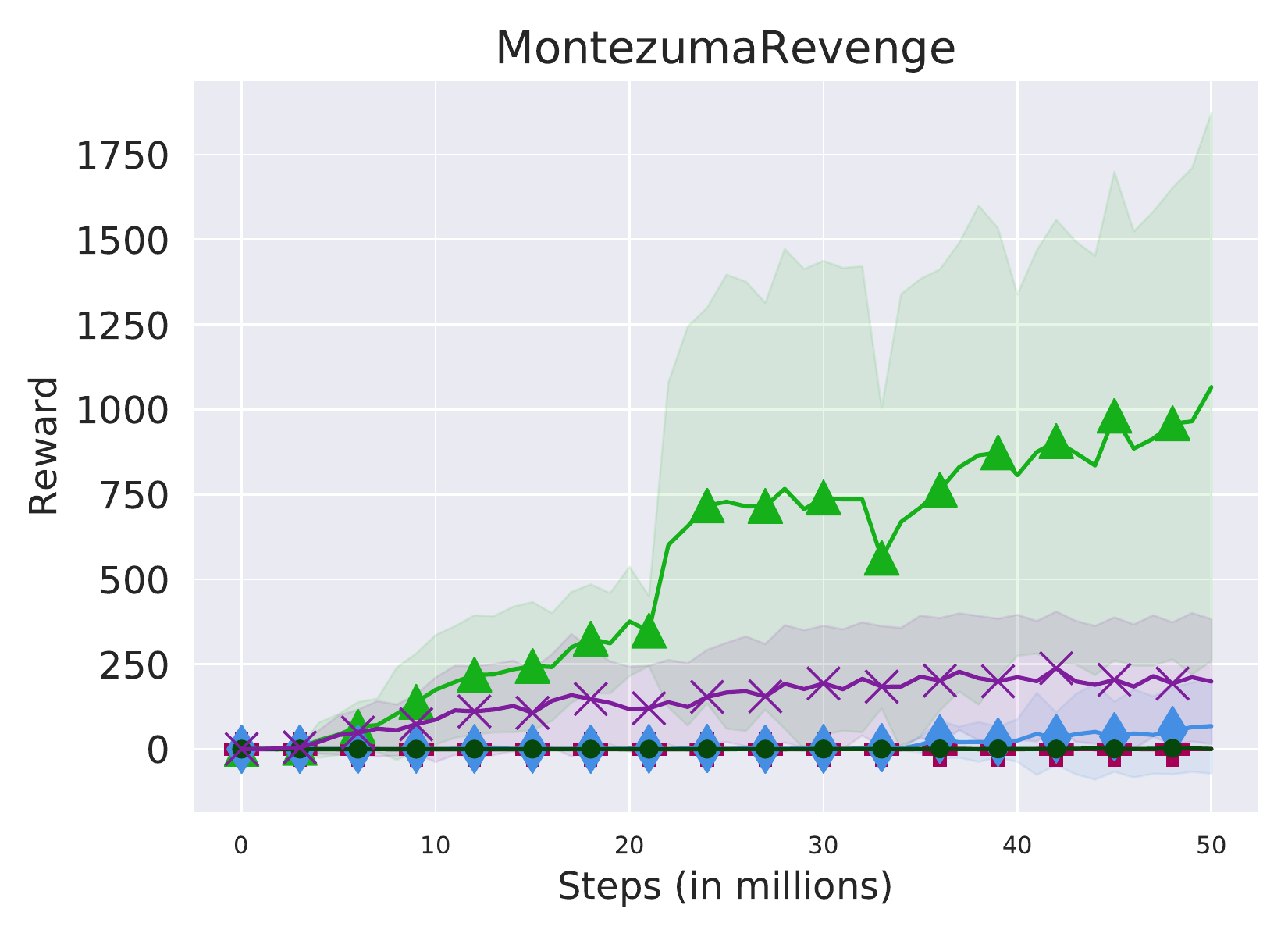}
         \caption{Montezuma's Revenge}
         \label{fig:ablation-montezuma}
     \end{subfigure}
    \caption{Ablation studies on LiDER in six Atari games. Results show that using only experiences where the return is improved, the two-buffer architecture, and the flexible sampling method does indeed improve performance. The x-axis is the total number of environmental steps: A3CTBSIL counts steps from 16 A3C workers, while LiDER counts steps from 15 A3C workers plus one refresher worker. The y-axis is the average testing score over eight trials; shaded regions show the standard deviation. } 
    \label{fig:ablation}
\end{figure*}

\subsection{Results}
Figure \ref{fig:ablation} shows the results of all ablation studies compared to A3CTBSIL and LiDER. The performance of LiDER-AddAll degraded in four out of six games, except for in Gopher and NameThisGame, where LiDER-AddAll performs comparably to LiDER. This could be because they are easy-exploration games with dense reward functions (as categorized by \citet{bellemare2016unifying}) thus the refresher is more likely to generate better trajectories in these games; adding a few ``bad'' samples (i.e., $\Gnew \leq \G$) does not hurt the general performance. In Alien and Montezuma's Revenge, LiDER-AddAll performs at about the same level as the baseline A3CTBSIL method. Ms.~Pac-Man shows the least amount of performance drop for LiDER-AddAll, but it still under-performed LiDER. In Freeway, while LiDER-AddAll eventually reaches the same score as LiDER, it struggled during the early stages of training. 
These results demonstrate the importance of focusing the exploitation only on places where the refresher can do better than what the agent had previously experienced. 

In all games, LiDER-OneBuffer significantly under-performed LiDER ($p \ll 0.001$). Especially in the game of Gopher, NameThisGame, and Ms.~Pac-Man where they also performed worse than the baseline A3CTBSIL. These results confirm our analysis in Section \ref{sec:analyses-sil-lider} that the SIL worker must be able to effectively leverage the high-quality data generated by the refresher to improve learning. When mixing the refresher-generated data and the A3C-generated data into one buffer, it is less likely for the SIL to sample from the good data. Thus, our design of the two-buffer architecture was well-chosen.  

LiDER-SampleR's performance was significantly \\ worse than LiDER in five out of six games ($p \ll 0.001$). Except for in Freeway where LiDER-SampleR eventually reaches the same performance as LiDER---but it acts quite unstable (the variance is high). We hypothesize that sampling only from buffer $\R$ reduces the amount of state the agent can experience, leading to a lack of exploration which impairs the learning. Therefore, despite that the refresher can generate higher quality data, the agent should learn from both A3C-generated data and refresher-generated data. 

In summary, in this section we presented three ablation studies to show the benefits of our design choices of LiDER. Using only experiences where the return is improved, the two-buffer architecture, and the flexible sampling strategy between buffer $\D$ and buffer $\R$ indeed improve performance. 



\section{Extensions: leveraging other policies to refresh past states}
\label{sec:lider-ta-bc}

\begin{figure*}
    \centering
     \begin{subfigure}[th]{0.35\textwidth}
         \centering
         \includegraphics[width=\textwidth]{lider-a3c-legend-final.png}
     \end{subfigure}
     \hspace{0.5\textwidth} 
     \begin{subfigure}[th]{0.95\textwidth}
         \centering
         \includegraphics[width=\textwidth]{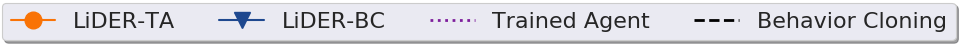}
     \end{subfigure}
     \vspace{10pt}
    \begin{subfigure}[th]{0.48\textwidth}
         \centering
         \includegraphics[width=\textwidth]{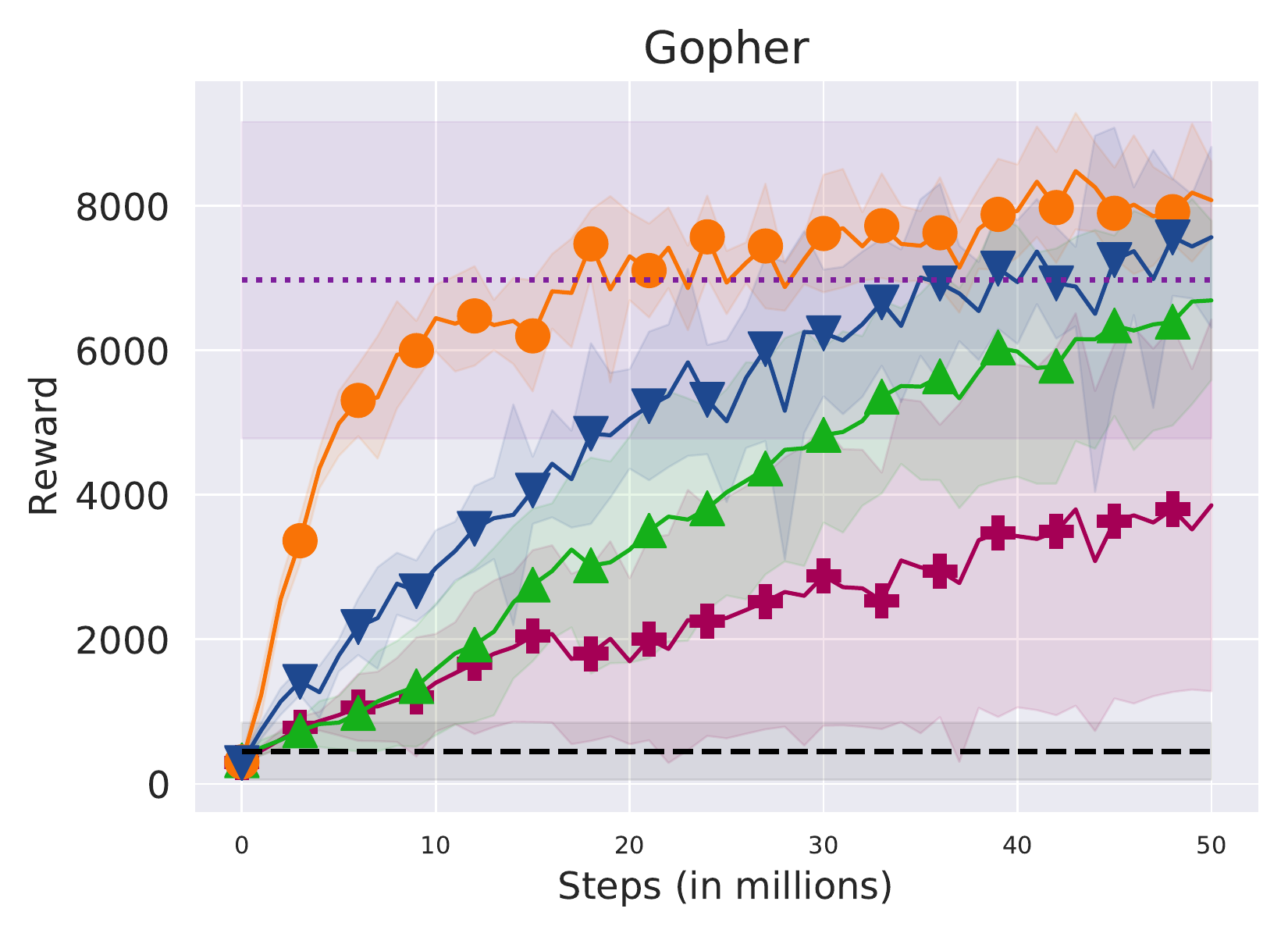}
         \caption{Gopher}
         \label{fig:lider-ta-bc-gopher}
     \end{subfigure}
     \begin{subfigure}[th]{0.48\textwidth}
         \centering
         \includegraphics[width=\textwidth]{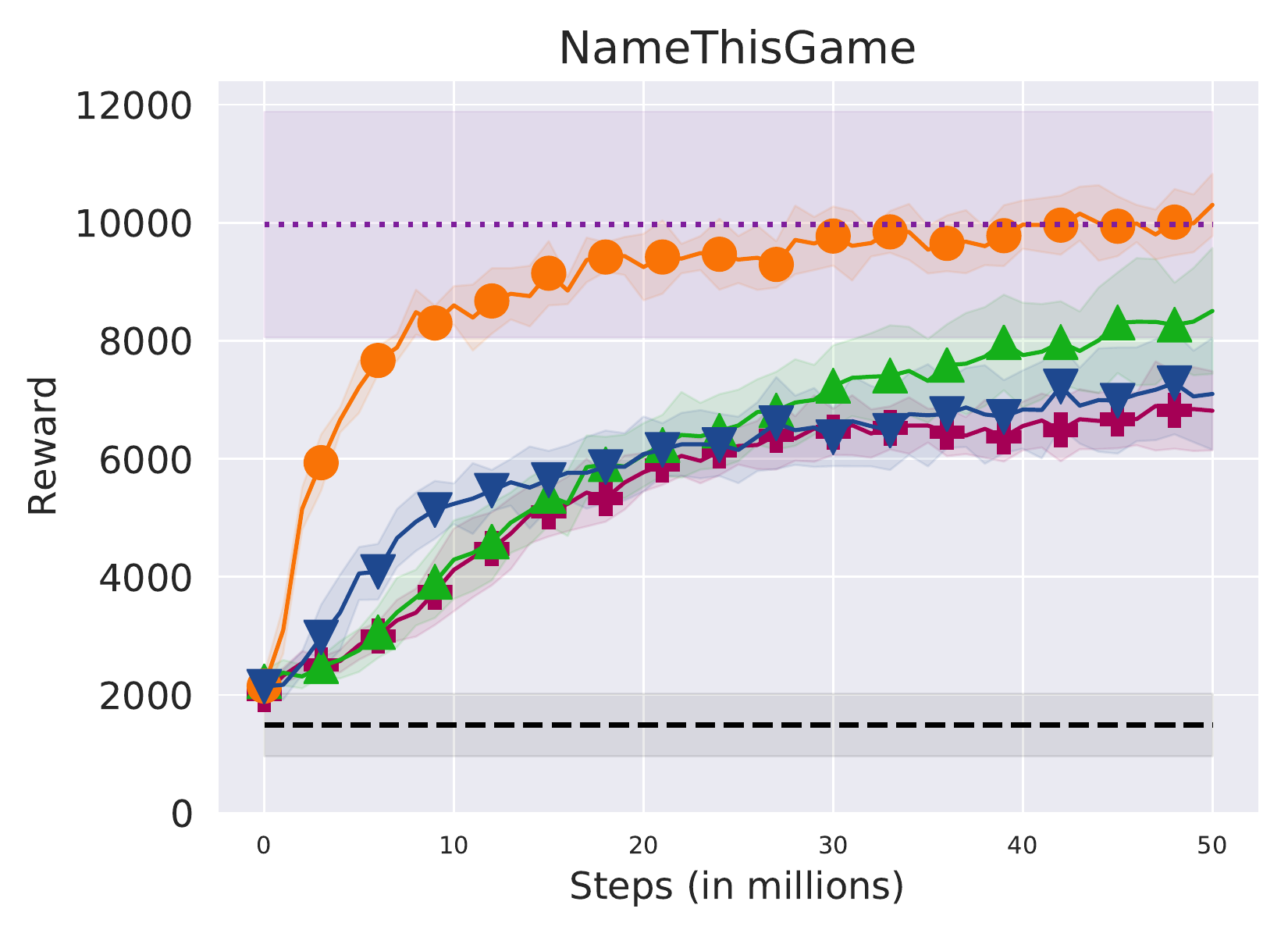}
         \caption{NameThisGame}
         \label{fig:lider-ta-bc-name}
     \end{subfigure}
     \begin{subfigure}[th]{0.48\textwidth}
         \centering
         \includegraphics[width=\textwidth]{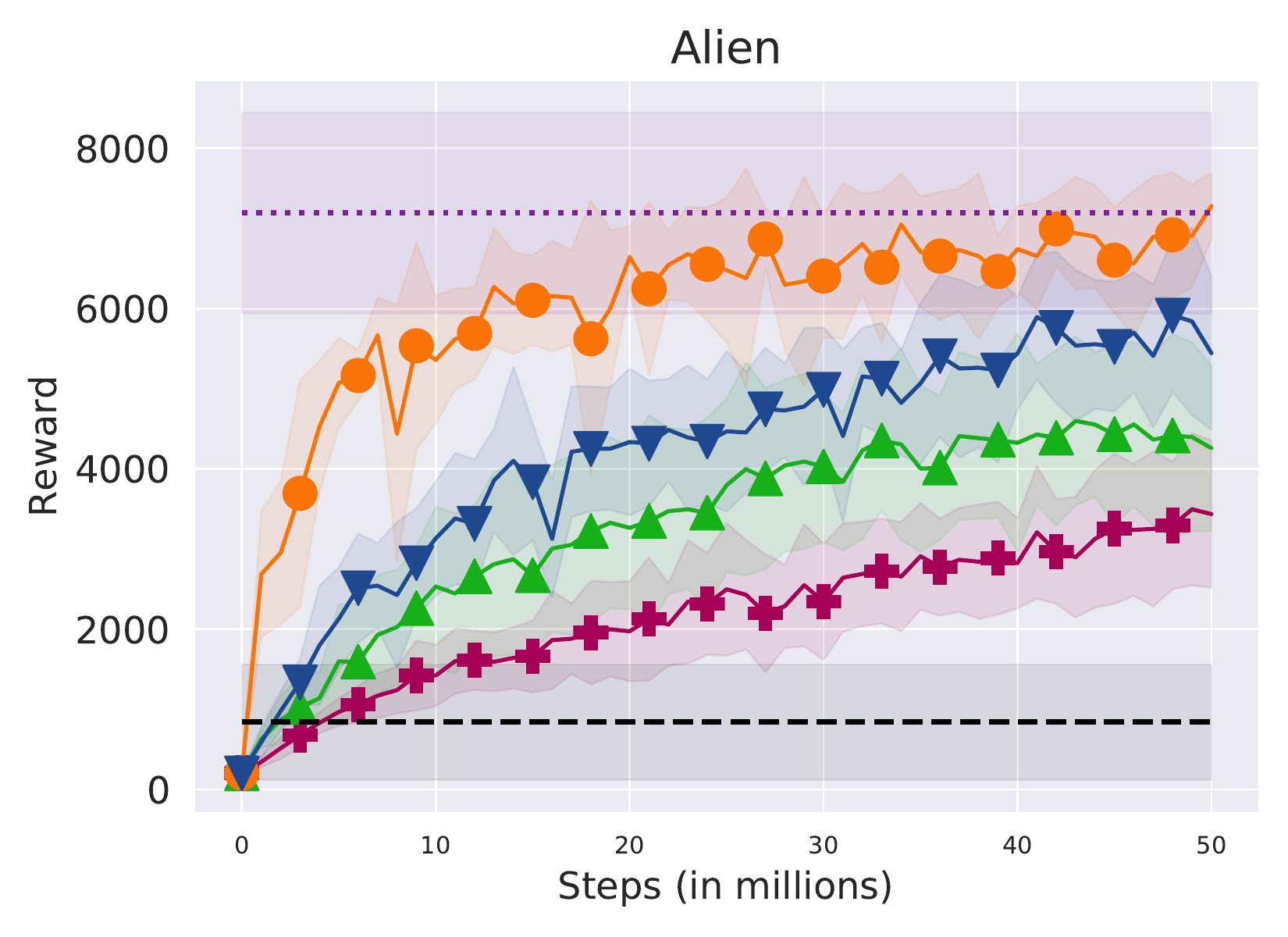}
         \caption{Alien}
         \label{fig:lider-ta-bc-alien}
     \end{subfigure}
     \begin{subfigure}[th]{0.48\textwidth}
         \centering
         \includegraphics[width=\textwidth]{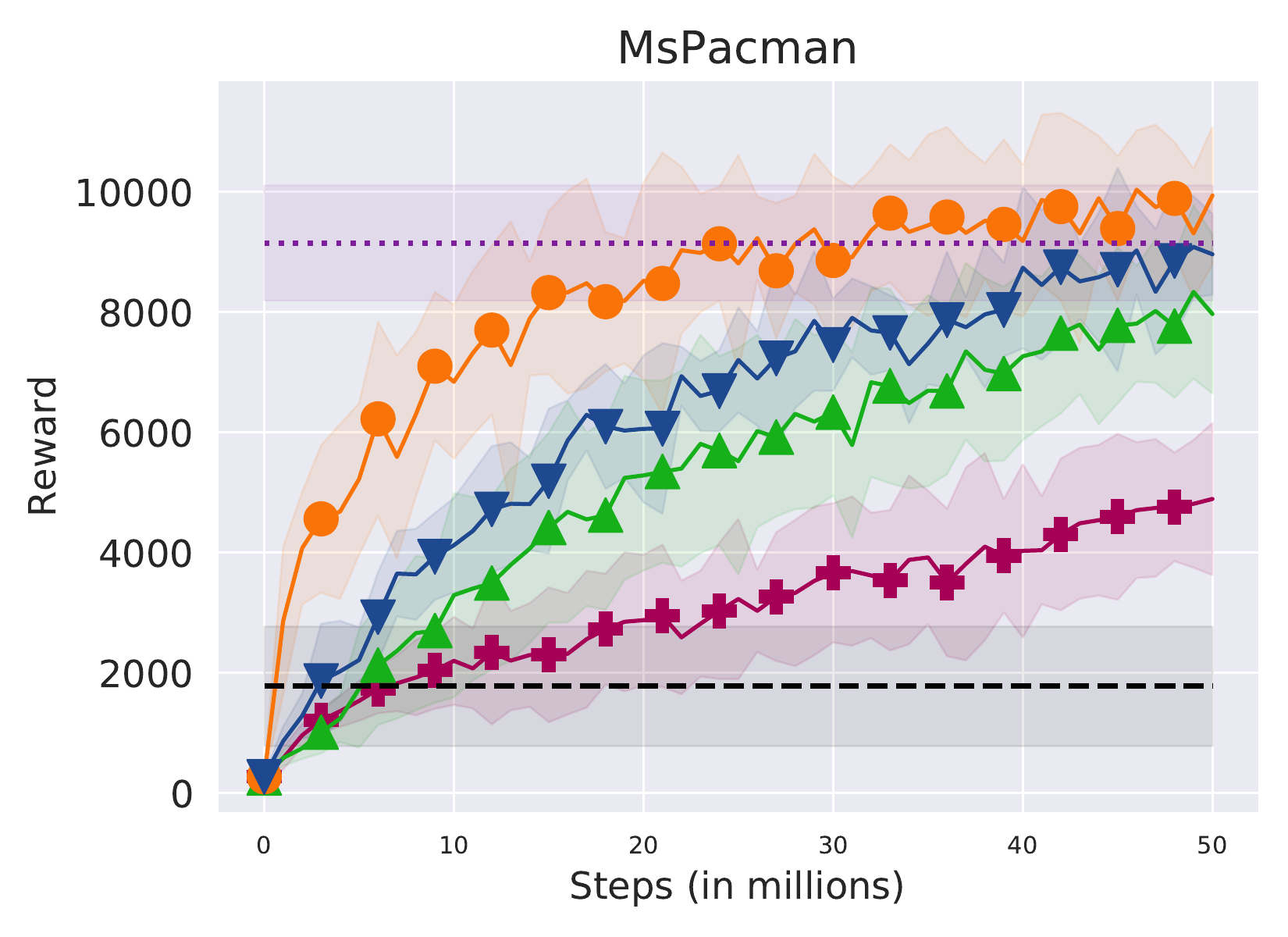}
         \caption{Ms.~Pac-Man}
         \label{fig:lider-ta-bc-mspacman}
     \end{subfigure}
     \begin{subfigure}[th]{0.48\textwidth}
         \centering
         \includegraphics[width=\textwidth]{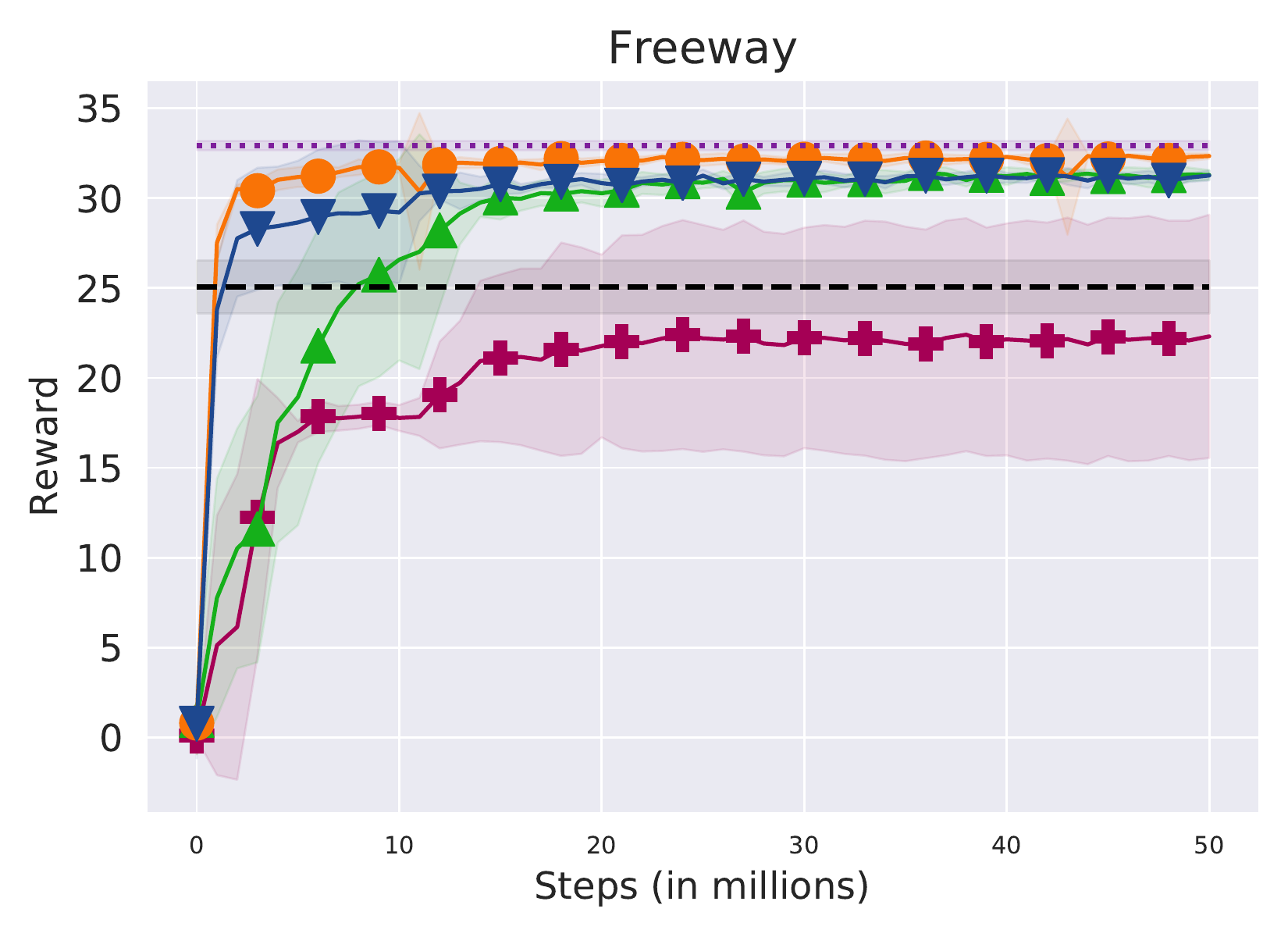}
         \caption{Freeway}
         \label{fig:lider-ta-bc-freeway}
     \end{subfigure}
     \begin{subfigure}[th]{0.48\textwidth}
         \centering
         \includegraphics[width=\textwidth]{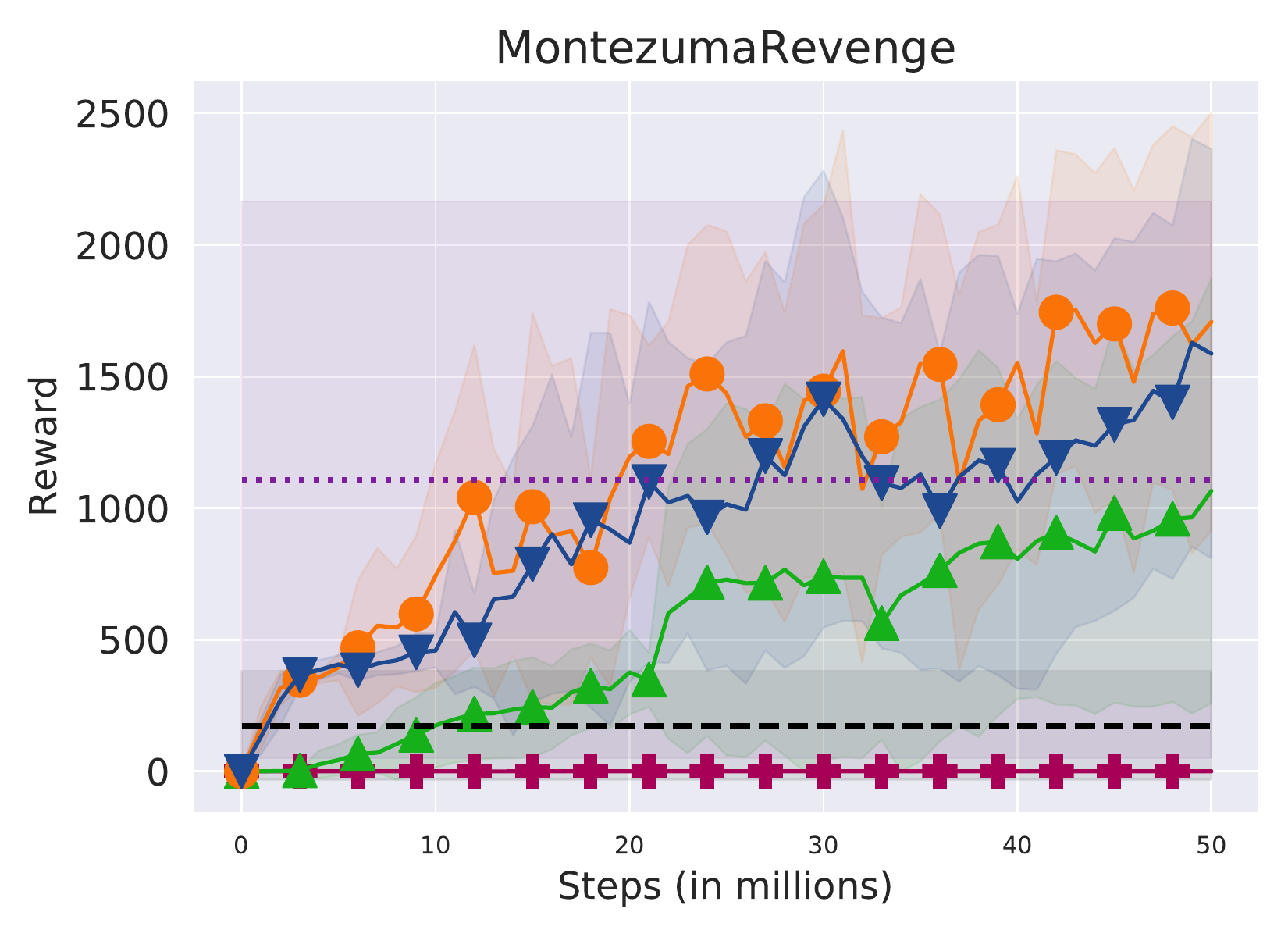}
         \caption{Montezuma's Revenge}
         \label{fig:lider-ta-bc-montezuma}
     \end{subfigure}
    \caption{LiDER-TA and LiDER-BC outperform A3CTBSIL and LiDER. The x-axis is the total number of environmental steps: A3CTBSIL counts steps from 16 A3C workers, while LiDER counts steps from 15 A3C workers plus one refresher worker. The y-axis is the average testing score over eight trials; shaded regions show the standard deviation. } 
    \label{fig:lider-ta-bc}
\end{figure*}

So far, we have shown in Section \ref{sec:lider-a3c} that LiDER outperformed the baseline A3CTBSIL. The analyses in Section \ref{sec:analyses} revealed why LiDER helps learning. Section \ref{sec:ablation} validated the design choices of LiDER through three ablation studies. In this section, we present two extensions to show that LiDER  
can leverage not only the agent's current policy, but also policies from external sources to refresh past states. 

In particular, we consider leveraging a trained agent (TA) and a behavior cloning (BC) model trained from human demonstration data. LiDER-TA uses a trained agent (TA) as the refresher. While the TA could come from any source, we use the best checkpoint from a fully trained LiDER agent from experiments in Section~\ref{sec:lider-a3c} as the TA. This scenario tests whether LiDER can effectively leverage a high-quality policy. 

LiDER-BC uses a behavior cloning (BC) model in the refresher. The BC policy is far from expert and we explore if LiDER can benefit from a sub-optimal policy. The BC model in LiDER-BC is pre-trained with non-expert demonstration data\footnote{The data is publicly available: \url{github.com/gabrieledcjr/atari_human_demo}}, which was collected in our previous work \cite{de2019jointly}. Then, we follow the pre-training method introduced in our previous work \cite{de2019jointly} to jointly pre-train a model with supervised, value, and unsupervised autoencoder losses, which gives us a BC model trained from human demonstration data (see Appendix \ref{sec:appe-pretrain} for pre-training details). The difference between the TA and BC model used in our setting is that the TA is an RL agent trained with LiDER while the BC model is trained only with human demonstrations. 

Figure \ref{fig:lider-ta-bc} shows the results of LiDER-TA and LiDER-BC compared with A3CTBSIL and LiDER (averaged over eight trials). As expected, LiDER-TA performs better than the other three methods, since it uses a trained agent as the refresher---the learning agent can observe and learn from high-quality data generated by an expert. LiDER-TA even exceeds the TA's performance in Gopher and Montezuma's Revenge. The TA's performance is shown in the purple dotted line (shaded regions show the standard deviation), estimated by executing the TA greedily in the game for 50 episodes. See Appendix \ref{sec:appe-TAmodels} for the score of each TA. 

The more interesting result is the performance of LiDER-BC, which demonstrates that LiDER works well even when using a refresher that is far from expert. The black dashed line shows the average performance of the BC model (shaded regions show the standard deviation), estimated by executing the model greedily in the game for 50 episodes (see Appendix \ref{sec:appe-pretrain} for the score of each BC). LiDER-BC can learn to quickly outperform BC and achieve better results than the baseline. LiDER-BC also slightly outperforms LiDER in five out of the six games, except for NameThisGame in which LiDER-BC outperforms LiDER initially, but later plateaued lower than LiDER. These results suggest that the sub-optimal BC model was able to provide better-than-random data during the early stages of training, which in turn helps the learning in the later stages. LiDER-BC could thus be one method of leveraging imperfect demonstrations to improve RL.



\section{Related work}
LiDER is related to several research directions in the RL literature; we briefly review four of them in this section. 

\subsection{Experience replay and extensions} 
ER was first introduced to improve the data efficiency of off-policy RL algorithms \cite{ERlin1992} and has since become an essential component for off-policy deep RL \cite{dqn}. Many techniques have been proposed to enhance ER for better data efficiency and generally fall into two categories. One category focuses on biasing the sampling strategy such that important experiences are reused more frequently for policy updates \cite{RER,ReFER,PER,sinha2020likelihood-exp,ER-RND,ERO}. The other category focuses on tuning the replay buffer architecture, such as changing the buffer size \cite{de2015importance,aER,CER}, combining experiences from multiple workers to generate more data to replay \cite{espeholt2018impala,dqn-apex,rnn-apex}, or augmenting the structure or content of replay experiences (e.g., generating additional ``goal states'' \cite{HER} or modifying experiences based on a teacher's advice \cite{chan2019ACTRCE}). 

LiDER does not fall into the first category but is complementary to existing sampling methods. We leverage prioritized experience replay \cite{PER} in our experiments: experiences are prioritized by advantages in buffer $\D$ and buffer $\R$, the SIL worker samples from both buffers with priority (although the refresher worker samples randomly from buffer $\D$). LiDER is related to the second category but differs in three ways. First, LiDER uses two replay buffers which double the buffer size, but we have shown that simply extending the size of a single buffer does not achieve the same performance as LiDER. Second, the refresher worker generates additional data, which is similar to using multiple workers to generate more data, but we kept the total number of workers the same between LiDER and the baseline and accounted for all environmental steps. Third, the refresher-generated data is stored in a separate buffer only when it has a higher return than the old data, which can be viewed as augmenting the quality of the data, but we do not change the data structure when storing them. 

Recently, \citet{Fedus2020revisitingER} revisited the fundamentals of ER to study how the architecture and the content of a replay buffer can affect learning. One of their key findings was that performance can be improved by increasing the replay buffer size and decreasing the age of the oldest data stored in the buffer. LiDER’s refreshing mechanism achieves exactly this purpose: the two-buffer architecture doubles the replay size and the refresher worker refreshes older experiences with a newer policy, which reduces the age of the overall policy. Therefore, LiDER can be viewed as a validation of the above finding by \citet{Fedus2020revisitingER}. 

\subsection{Experience replay for actor-critic algorithms} 
The difficulty of combining ER into actor-critic algorithms is caused by the discrepancy between the current policy and the past policy that generated the experience. This problem is usually solved by leveraging various importance sampling techniques, such that the bias from past experiences can be corrected when used for updating the current policy \cite{espeholt2018impala,reactor,munos2016retrace,ACER,wawrzynski2009real}. In this work, we chose to use the SIL algorithm over the other actor-critic with ER algorithms because SIL provides a straightforward way of integrating ER into A3C without importance sampling \cite{sil}. 

As proven theoretically by \citet{sil}, the SIL objective (Equation \ref{eq:sil_update}) updates the policy and the value function directly towards optimal by leveraging the Mo-nte-Carlo return $\G$, which can be viewed as a form of lower-bound-soft-Q learning. Thus, off-policy correction techniques like importance sampling are not needed even though the SIL worker learns from off-policy data (i.e., from a replay buffer), while the A3C worker learns on-policy. In addition, \citet{sil} have shown that the SIL objective is compatible, not conflicting, with off-policy correction algorithms like ACER \cite{ACER}. LiDER builds upon the SIL objective and thus shares similar properties. Incorporating LiDER into other off-policy RL algorithms is important for future work (as described in Section \ref{sec:conclusion}). 

\subsection{Learning from past good experiences of oneself} 
The main idea of LiDER is to allow the agent to learn from past states that have been improved by its current policy. Several existing methods have shown that it is beneficial for the agent to learn from its past good experiences. For example, the optimality tightening proposed by \citet{opt-tightening} constrains the Q function with lower and upper bounds, with the intuition that the Q function should be updated using trajectories that perform better than the current policy. The self-imitation learning (SIL) algorithm was inspired by the lower-bound Q learning from optimality tightening that only trajectories with positive advantages should be used to update the policy \cite{sil}. 

\citet{gangwani2018learning} and \citet{guo2020memory} extended the SIL algorithm and found that the performance can be further improved if the past good experiences are also diverse---diversity helps drive exploration. While we did not design LiDER to explicitly leverage exploration techniques, LiDER revisits a past state, then generates new trajectories using a different policy, which could potentially lead to unseen states and increase the data diversity. This implicit exploration could be one of the reasons that LiDER improves the performance of two hard exploration Atari games.

A generalized form of SIL algorithm was proposed recently by \citet{tang2020self}. This new algorithm leverages n-step lower bound Q-learning which improves the original SIL algorithm in two aspects: 1) the agent can now self-imitate partial trajectories while the original SIL algorithm requires learning from a full trajectory, and 2) bootstrapping from learned Q-functions is enabled while the original SIL algorithm does not bootstrap from learned Q-functions. The generalized SIL algorithm can be applied to both deterministic and stochastic RL algorithms and outperforms SIL in a wide range of continuous control tasks. Leveraging the generalized SIL algorithm could be an interesting future work to improve LiDER. 

Interestingly, the idea of learning from refreshed past states was also used in the MuZero algorithm \cite{muzero}, a tree-based searching algorithm that combines a learned model. Specifically, MuZero introduced a second variant called MuZero Reanalyze, in which the agent revisits a past time step and performs Monte-Carlo tree search again using the current model parameters. According to \citet{muzero}, MuZero Reanalyze largely improves the performance of MuZero because the reanalyze process potentially results in better policy than the original search. LiDER's results align with the findings of MuZero Reanalyze: an agent's current policy can be used to generate better quality data from a past state; leveraging these data leads to improved performance. 

\subsection{Relocating the agent to a past state} 
LiDER assumes there is a simulator for the task where resetting to a previously seen state is possible. The idea of relocating the agent to past states has been explored in the literature (e.g., \citet{mihalkova2006relocation}). Particularly, in the research area of curriculum learning, 
it is common to assume a simulator is available and the agent can be reset to any arbitrary state at the beginning of training (e.g., \citet{florensa17a}). A similar line of work has found that resetting the agent to a past state, instead of the simulator's default initial state, can benefit the learning. Such a state can be drawn from different distributions over the replay buffer \cite{tavakoli2018exploring} or from human demonstrations \cite{human-checkpoint-replay,nair2018overcoming,resnick2018backplay,mont-single-demo,zhang2020teachingdim}. Many simulators are already equipped with the ability to relocate the agent so that they can reset the agent to an initial state when an episode ends. LiDER makes full use of this common feature. 

While we can exploit simulators' relocation features if one is available, there are also situations when such a feature does not exist.  
The recently developed policy-based Go-Explore algorithm 
learned a goal-conditioned policy to guide the agent to return to a past state, which enables relocating without using the simulator reset feature \cite{newgoexplore}.\footnote{The policy-based Go-Explore algorithm is an extension of the Go-Explore without a policy framework, which was presented in an earlier pre-print \cite{goexplore}. Go-Explore without a policy framework also leverages the simulator reset feature.} 
Concurrently with policy-based Go-Explore, \citet{guo2020memory} proposed the Diverse Trajectory-conditi-oned Self-Imitation Learning (DTSIL) algorithm. It 
uses similar mechanisms as \citet{newgoexplore} to train a goal-condition based policy for the relocating process and no simulator reset is needed for DTSIL.\footnote{\citet{newgoexplore} made a detailed comparison between the policy-based Go-Explore and DTSIL. We refer the interested readers to \citet{newgoexplore} for further reading. } 

Both the policy-based Go-Explore and DTSIL algorithms share similarities with LiDER in that they first teleport the agent to a past state then explore from there. However, LiDER is distinct from these two algorithms in three perspectives. First, the functionality of the learned policy is different. LiDER learns an actor-critic policy that maximizes the cumulative return; its policy takes a state as input and produces actions that can achieve maximum return. While the policy-based Go-Explore and DTSIL's policies are goal-conditioned and only learn how to return the agent to a past state, not how to maximize return. Their policies take both the agent's current state and the selected state (called a goal state) as input and produce actions that will lead the agent back to the goal state. Second, the state selection strategy is different. LiDER teleports the agent back to a randomly selected state, while the policy-based Go-Explore and DTSIL algorithms select "novel'" states to return to (i.e., states that are rarely visited). Lastly, from the relocated state, LiDER then performs a refresh with its current policy. While the policy-based Go-Explore explores with either random actions or actions sampled from the goal-conditioned policy (with equal probability), the DTSIL algorithm only explores randomly from that state.  

Besides the three key differences, there are many minor distinctions between LiDER and these two algorithms, such as the main goal and structure of the algorithm, the replay buffer architecture, the state representations, and the hyperparameters. Because of these differences, policy-based Go-Explore and DTSIL are not directly comparable to LiDER. On the other hand, LiDER can be considered as more evidence that supports the core benefits of ``agent relocation,'' rather than a competing method. Nevertheless, leveraging the relocation mechanism of these two algorithms in LiDER can be an important step towards allowing LiDER to work outside of simulations, as mentioned in the future work discussion in Section \ref{sec:conclusion}.  

\section{Discussion and future work}
\label{sec:conclusion}
In this paper, we proposed \emph{\textbf{L}uc\textbf{i}d \textbf{D}reaming for \textbf{E}xperi-\\ence \textbf{R}eplay (LiDER)}, a conceptually new framework that allows experiences in the replay buffer to be refreshed by leveraging the agent's current policy, leading to improved performance compared to the baseline method without refreshing past experiences. We investigated the underlying behavior of the refresher to better understand why LiDER helps learning. We also conducted several ablation studies to validate our design choices of LiDER. Two extensions demonstrated that LiDER is also capable of leveraging knowledge from external policies, such as a trained agent and a behavior cloning model. One potential limitation of LiDER is that it must have access to a simulator that can return to previously visited states before resuming.

This paper opens up several new interesting directions for future work. First, based on the initial positive results reported in this paper, additional computational resources ought to be devoted to evaluating LiDER in a broad variety of domains. 

Second, while we have presented in this paper a case study of applying LiDER to a multi-worker, actor-critic based algorithm, future work could investigate extending LiDER to other types of off-policy RL algorithms that leverage ER. We expect LiDER to be most applicable to the PPO+SIL algorithm \citet{sil}. 
PPO+SIL's multi-worker, actor-critic architecture allows the refresher worker to be easily added as was done in A3CTBSIL. Similarly, the SIL objective (Equation \eqref{eq:sil_update}) of PPO+SIL enables integrating ER, and the agent can learn from both PPO-generated and refresher-generated experiences. 

On the other hand, applying LiDER to single-worker, value-based algorithms, such as the deep Q-network (DQN) algorithm \cite{dqn}, presents more challenges. The first is that it is non-trivial to decide how often one should ``pause'' the training and perform a refresh. In multi-worker architectures, we do not need to explicitly control this frequency as all workers are running in parallel. With a single worker, the training and the refreshing cannot happen simultaneously. Another challenge is that, in value-based algorithms like DQN, a value function is learned instead of directly learning a policy. The Q value is updated using a one-step TD error instead of the Monte-Carlo return. Since the key concept of LiDER is to leverage higher returns for policy updates, how to integrate returns in value updates should be considered carefully before applying LiDER to DQN.
Nevertheless, LiDER has the potential to benefit other off-policy algorithms that use ER, which is a good direction to explore in future work. 

Third, the refresher in LiDER-BC uses a fixed policy from behavior cloning. Future work could investigate whether it helps to use different policies during training. For example, one could use the BC policy during the early stages of training, and then once A3C's current policy outperforms BC, replace it with the A3C policy. Additionally, it is thus natural to consider adding multiple SIL and/or refresher works to enable leveraging multiple policies. Investigating how the proportion among the number of A3C, SIL, and refresher workers affects performance would make for an interesting future study. 

Fourth, it would be interesting to allow LiDER to work outside of simulations by returning to a similar, but not identical state, and from there generate new trajectories. For example, in robotics, a robot may be able to return to a position that is close to, but not identical to, a previously experienced state.

\begin{acknowledgements}
We thank Gabriel V.~{de la Cruz Jr.} for helpful discussions; his open-source code at \url{github.com/gabrieledcjr/DeepRL} is used for training the behavior  cloning models in this work. This research used resources of Kamiak, Washington State University's high-performance computing cluster. Assefaw Gebremedhin is supported by the NSF award IIS-1553528. Part of this work has taken place in the Intelligent Robot Learning (IRL) Lab at the University of Alberta, which is supported in part by research grants from the Alberta Machine Intelligence Institute (Amii), CIFAR, and NSERC. Part of this work has taken place in the Learning Agents Research Group (LARG) at UT Austin. LARG research is supported in part by NSF (CPS-1739964, IIS-1724157, NRI-1925082), ONR (N00014-18-2243), FLI (RFP2-000), ARL, DARPA, Lockheed Martin, GM, and Bosch. Peter Stone serves as the Executive Director of Sony AI America and receives financial compensation for this work. The terms of this arrangement have been reviewed and approved by the University of Texas at Austin in accordance with its policy on objectivity in research.
\end{acknowledgements}

%
\section*{Conflict of interest}
The authors declare that they have no conflict of interest.

\bibliographystyle{spbasic}      
\bibliography{reference}   

\begin{thebibliography}{49}
\providecommand{\natexlab}[1]{#1}
\providecommand{\url}[1]{{#1}}
\providecommand{\urlprefix}{URL }
\expandafter\ifx\csname urlstyle\endcsname\relax
  \providecommand{\doi}[1]{DOI~\discretionary{}{}{}#1}\else
  \providecommand{\doi}{DOI~\discretionary{}{}{}\begingroup
  \urlstyle{rm}\Url}\fi
\providecommand{\eprint}[2][]{\url{#2}}

\bibitem[{Andrychowicz et~al.(2017)Andrychowicz, Wolski, Ray, Schneider, Fong,
  Welinder, McGrew, Tobin, Pieter~Abbeel, and Zaremba}]{HER}
Andrychowicz M, Wolski F, Ray A, Schneider J, Fong R, Welinder P, McGrew B,
  Tobin J, Pieter~Abbeel O, Zaremba W (2017) {Hindsight Experience Replay}. In:
  Guyon I, Luxburg UV, Bengio S, Wallach H, Fergus R, Vishwanathan S, Garnett R
  (eds) Advances in Neural Information Processing Systems, Curran Associates,
  Inc., vol~30, pp 5048--5058,
  \urlprefix\url{https://proceedings.neurips.cc/paper/2017/file/453fadbd8a1a3af50a9df4df899537b5-Paper.pdf}

\bibitem[{Bellemare et~al.(2016)Bellemare, Srinivasan, Ostrovski, Schaul,
  Saxton, and Munos}]{bellemare2016unifying}
Bellemare M, Srinivasan S, Ostrovski G, Schaul T, Saxton D, Munos R (2016)
  {Unifying Count-Based Exploration and Intrinsic Motivation}. In: Lee D,
  Sugiyama M, Luxburg U, Guyon I, Garnett R (eds) Advances in Neural
  Information Processing Systems, Curran Associates, Inc., vol~29, pp
  1471--1479,
  \urlprefix\url{https://proceedings.neurips.cc/paper/2016/file/afda332245e2af431fb7b672a68b659d-Paper.pdf}

\bibitem[{Bellemare et~al.(2013)Bellemare, Naddaf, Veness, and
  Bowling}]{bellemare2013arcade}
Bellemare MG, Naddaf Y, Veness J, Bowling M (2013) {The Arcade Learning
  Environment: An Evaluation Platform for General Agents}. Journal of
  Artificial Intelligence Research 47(1):253–279

\bibitem[{Chan et~al.(2019)Chan, Wu, Kiros, Fidler, and Ba}]{chan2019ACTRCE}
Chan H, Wu Y, Kiros J, Fidler S, Ba J (2019) {ACTRCE: Augmenting Experience via
  Teacher's Advice For Multi-Goal Reinforcement Learning}. arXiv preprint
  arXiv:190204546 abs/1902.04546

\bibitem[{de~la Cruz et~al.(2019)de~la Cruz, Du, and Taylor}]{de2019pre}
de~la Cruz GV, Du Y, Taylor ME (2019) {Pre-training with Non-expert Human
  Demonstration for Deep Reinforcement Learning}. The Knowledge Engineering
  Review 34:e10, \doi{10.1017/S0269888919000055}

\bibitem[{de~la Cruz~Jr et~al.(2019)de~la Cruz~Jr, Du, and
  Taylor}]{de2019jointly}
de~la Cruz~Jr GV, Du Y, Taylor ME (2019) {Jointly Pre-training with Supervised,
  Autoencoder, and Value Losses for Deep Reinforcement Learning}. In: Adaptive
  and Learning Agents Workshop, AAMAS

\bibitem[{{Dao} and {Lee}(2019)}]{RER}
{Dao} G, {Lee} M (2019) {Relevant Experiences in Replay Buffer}. In: 2019 IEEE
  Symposium Series on Computational Intelligence (SSCI), pp 94--101,
  \doi{10.1109/SSCI44817.2019.9002745}

\bibitem[{De~Bruin et~al.(2015)De~Bruin, Kober, Tuyls, and
  Babu{\v{s}}ka}]{de2015importance}
De~Bruin T, Kober J, Tuyls K, Babu{\v{s}}ka R (2015) {The Importance of
  Experience Replay Database Composition in Deep Reinforcement Learning}. In:
  Deep Reinforcement Learning Workshop, NIPS

\bibitem[{Ecoffet et~al.(2019)Ecoffet, Huizinga, Lehman, Stanley, and
  Clune}]{goexplore}
Ecoffet A, Huizinga J, Lehman J, Stanley KO, Clune J (2019) {Go-explore: a New
  Approach for Hard-exploration Problems}. arXiv preprint arXiv:190110995

\bibitem[{Ecoffet et~al.(2020)Ecoffet, Huizinga, Lehman, Stanley, and
  Clune}]{newgoexplore}
Ecoffet A, Huizinga J, Lehman J, Stanley KO, Clune J (2020) {First Return Then
  Explore}. arXiv preprint arXiv:200412919

\bibitem[{Espeholt et~al.(2018)Espeholt, Soyer, Munos, Simonyan, Mnih, Ward,
  Doron, Firoiu, Harley, Dunning, Legg, and Kavukcuoglu}]{espeholt2018impala}
Espeholt L, Soyer H, Munos R, Simonyan K, Mnih V, Ward T, Doron Y, Firoiu V,
  Harley T, Dunning I, Legg S, Kavukcuoglu K (2018) {IMPALA: Scalable
  Distributed Deep-{RL} with Importance Weighted Actor-Learner Architectures}.
  Proceedings of Machine Learning Research 80:1407--1416,
  \urlprefix\url{http://proceedings.mlr.press/v80/espeholt18a.html}

\bibitem[{Fedus et~al.(2020)Fedus, Ramachandran, Agarwal, Bengio, Larochelle,
  Rowland, and Dabney}]{Fedus2020revisitingER}
Fedus W, Ramachandran P, Agarwal R, Bengio Y, Larochelle H, Rowland M, Dabney W
  (2020) {Revisiting Fundamentals of Experience Replay}. In: Proceedings of the
  37th International Conference on Machine Learning, PMLR,
  \urlprefix\url{https://proceedings.icml.cc/paper/2020/hash/5460b9ea1986ec386cb64df22dff37be-Abstract.html}

\bibitem[{Florensa et~al.(2017)Florensa, Held, Wulfmeier, Zhang, and
  Abbeel}]{florensa17a}
Florensa C, Held D, Wulfmeier M, Zhang M, Abbeel P (2017) {Reverse Curriculum
  Generation for Reinforcement Learning}. In: Levine S, Vanhoucke V, Goldberg K
  (eds) Proceedings of Machine Learning Research, PMLR, vol~78, pp 482--495,
  \urlprefix\url{http://proceedings.mlr.press/v78/florensa17a.html}

\bibitem[{Gangwani et~al.(2019)Gangwani, Liu, and Peng}]{gangwani2018learning}
Gangwani T, Liu Q, Peng J (2019) {Learning Self-Imitating Diverse Policies}.
  In: International Conference on Learning Representations,
  \urlprefix\url{https://openreview.net/forum?id=HyxzRsR9Y7}

\bibitem[{Gruslys et~al.(2018)Gruslys, Dabney, Azar, Piot, Bellemare, and
  Munos}]{reactor}
Gruslys A, Dabney W, Azar MG, Piot B, Bellemare M, Munos R (2018) {The Reactor:
  a Fast and Sample-efficient Actor-Critic Agent for Reinforcement Learning}.
  In: International Conference on Learning Representations,
  \urlprefix\url{https://openreview.net/forum?id=rkHVZWZAZ}

\bibitem[{Guo et~al.(2020)Guo, Choi, Moczulski, Feng, Bengio, Norouzi, and
  Lee}]{guo2020memory}
Guo Y, Choi J, Moczulski M, Feng S, Bengio S, Norouzi M, Lee H (2020) {Memory
  Based Trajectory-conditioned Policies for Learning from Sparse Rewards}. In:
  Advances in Neural Information Processing Systems,
  \urlprefix\url{https://papers.nips.cc/paper/2020/hash/2df45244f09369e16ea3f9117ca45157-Abstract.html}

\bibitem[{He et~al.(2017)He, Liu, Schwing, and Peng}]{opt-tightening}
He FS, Liu Y, Schwing AG, Peng J (2017) {Learning to Play in a Day: Faster Deep
  Reinforcement Learning by Optimality Tightening}. In: International
  Conference on Learning Representations,
  \urlprefix\url{https://openreview.net/forum?id=rJ8Je4clg}

\bibitem[{Hester et~al.(2018)Hester, Vecerik, Pietquin, Lanctot, Schaul, Piot,
  Horgan, Quan, Sendonaris, Osband, Dulac-Arnold, Agapiou, Leibo, and
  Gruslys}]{dqfd}
Hester T, Vecerik M, Pietquin O, Lanctot M, Schaul T, Piot B, Horgan D, Quan J,
  Sendonaris A, Osband I, Dulac-Arnold G, Agapiou J, Leibo JZ, Gruslys A (2018)
  {Deep Q-learning from Demonstrations}. In: Annual Meeting of the Association
  for the Advancement of Artificial Intelligence (AAAI), New Orleans (USA)

\bibitem[{Horgan et~al.(2018)Horgan, Quan, Budden, Barth-Maron, Hessel, van
  Hasselt, and Silver}]{dqn-apex}
Horgan D, Quan J, Budden D, Barth-Maron G, Hessel M, van Hasselt H, Silver D
  (2018) {Distributed Prioritized Experience Replay}. In: International
  Conference on Learning Representations,
  \urlprefix\url{https://openreview.net/forum?id=H1Dy---0Z}

\bibitem[{Hosu and Rebedea(2016)}]{human-checkpoint-replay}
Hosu IA, Rebedea T (2016) {Playing Atari Games with Deep Reinforcement Learning
  and Human Checkpoint Replay}. arXiv preprint arXiv:160705077

\bibitem[{Kapturowski et~al.(2019)Kapturowski, Ostrovski, Dabney, Quan, and
  Munos}]{rnn-apex}
Kapturowski S, Ostrovski G, Dabney W, Quan J, Munos R (2019) {Recurrent
  Experience Replay in Distributed Reinforcement Learning}. In: International
  Conference on Learning Representations,
  \urlprefix\url{https://openreview.net/forum?id=r1lyTjAqYX}

\bibitem[{Le et~al.(2018)Le, Patterson, and White}]{sae}
Le L, Patterson A, White M (2018) {Supervised Autoencoders: Improving
  Generalization Performance with Unsupervised Regularizers}. In: Bengio S,
  Wallach H, Larochelle H, Grauman K, Cesa-Bianchi N, Garnett R (eds) Advances
  in Neural Information Processing Systems, Curran Associates, Inc., vol~31, pp
  107--117,
  \urlprefix\url{https://proceedings.neurips.cc/paper/2018/file/2a38a4a9316c49e5a833517c45d31070-Paper.pdf}

\bibitem[{Lillicrap et~al.(2016)Lillicrap, Hunt, Pritzel, Heess, Erez, Tassa,
  Silver, and Wierstra}]{ddpg}
Lillicrap TP, Hunt JJ, Pritzel A, Heess N, Erez T, Tassa Y, Silver D, Wierstra
  D (2016) {Continuous Control with Deep Reinforcement Learning}. In:
  International Conference on Learning Representations,
  \urlprefix\url{https://openreview.net/forum?id=tX_O8O-8Zl}

\bibitem[{Lin(1992)}]{ERlin1992}
Lin LJ (1992) {Self-improving Reactive Agents Based on Reinforcement Learning,
  Planning and Teaching}. Machine learning 8(3-4):293--321

\bibitem[{{Liu} and {Zou}(2018)}]{aER}
{Liu} R, {Zou} J (2018) {The Effects of Memory Replay in Reinforcement
  Learning}. In: The 56th Annual Allerton Conference on Communication, Control,
  and Computing, pp 478--485

\bibitem[{Mihalkova and Mooney(2006)}]{mihalkova2006relocation}
Mihalkova L, Mooney R (2006) {Using Active Relocation to Aid Reinforcement
  Learning}. In: Prodeedings of the 19th International FLAIRS Conference
  (FLAIRS-2006), Melbourne Beach, FL, pp 580--585,
  \urlprefix\url{http://www.cs.utexas.edu/users/ai-lab?mihalkova:flairs06}

\bibitem[{Mnih et~al.(2015)Mnih, Kavukcuoglu, Silver, Rusu, Veness, Bellemare,
  Graves, Riedmiller, Fidjeland, Ostrovski et~al.}]{dqn}
Mnih V, Kavukcuoglu K, Silver D, Rusu AA, Veness J, Bellemare MG, Graves A,
  Riedmiller M, Fidjeland AK, Ostrovski G, et~al. (2015) {Human-level Control
  Through Deep Reinforcement Learning}. Nature 518(7540):529

\bibitem[{Mnih et~al.(2016)Mnih, Badia, Mirza, Graves, Lillicrap, Harley,
  Silver, and Kavukcuoglu}]{a3c}
Mnih V, Badia AP, Mirza M, Graves A, Lillicrap T, Harley T, Silver D,
  Kavukcuoglu K (2016) {Asynchronous Methods for Deep Reinforcement Learning}.
  In: Balcan MF, Weinberger KQ (eds) Proceedings of Machine Learning Research,
  PMLR, New York, New York, USA, vol~48, pp 1928--1937,
  \urlprefix\url{http://proceedings.mlr.press/v48/mniha16.html}

\bibitem[{Munos et~al.(2016)Munos, Stepleton, Harutyunyan, and
  Bellemare}]{munos2016retrace}
Munos R, Stepleton T, Harutyunyan A, Bellemare M (2016) {Safe and Efficient
  Off-Policy Reinforcement Learning}. In: Lee D, Sugiyama M, Luxburg U, Guyon
  I, Garnett R (eds) Advances in Neural Information Processing Systems, Curran
  Associates, Inc., vol~29, pp 1054--1062,
  \urlprefix\url{https://proceedings.neurips.cc/paper/2016/file/c3992e9a68c5ae12bd18488bc579b30d-Paper.pdf}

\bibitem[{{Nair} et~al.(2018){Nair}, {McGrew}, {Andrychowicz}, {Zaremba}, and
  {Abbeel}}]{nair2018overcoming}
{Nair} A, {McGrew} B, {Andrychowicz} M, {Zaremba} W, {Abbeel} P (2018)
  {Overcoming Exploration in Reinforcement Learning with Demonstrations}. In:
  2018 IEEE International Conference on Robotics and Automation (ICRA), pp
  6292--6299, \doi{10.1109/ICRA.2018.8463162}

\bibitem[{Novati and Koumoutsakos(2019)}]{ReFER}
Novati G, Koumoutsakos P (2019) {Remember and Forget for Experience Replay}.
  In: Chaudhuri K, Salakhutdinov R (eds) Proceedings of Machine Learning
  Research, PMLR, Long Beach, California, USA, vol~97, pp 4851--4860,
  \urlprefix\url{http://proceedings.mlr.press/v97/novati19a.html}

\bibitem[{Oh et~al.(2018)Oh, Guo, Singh, and Lee}]{sil}
Oh J, Guo Y, Singh S, Lee H (2018) {Self-Imitation Learning}. In: Dy J, Krause
  A (eds) Proceedings of Machine Learning Research, PMLR, Stockholmsmässan,
  Stockholm Sweden, vol~80, pp 3878--3887,
  \urlprefix\url{http://proceedings.mlr.press/v80/oh18b.html}

\bibitem[{Pohlen et~al.(2018)Pohlen, Piot, Hester, Azar, Horgan, Budden,
  Barth-Maron, van Hasselt, Quan, Ve{\v{c}}er{\'\i}k et~al.}]{apexdqn}
Pohlen T, Piot B, Hester T, Azar MG, Horgan D, Budden D, Barth-Maron G, van
  Hasselt H, Quan J, Ve{\v{c}}er{\'\i}k M, et~al. (2018) {Observe and Look
  Further: Achieving Consistent Performance on {A}tari}. arXiv preprint
  arXiv:180511593

\bibitem[{Resnick et~al.(2018)Resnick, Raileanu, Kapoor, Peysakhovich, Cho, and
  Bruna}]{resnick2018backplay}
Resnick C, Raileanu R, Kapoor S, Peysakhovich A, Cho K, Bruna J (2018)
  {Backplay:" Man muss immer umkehren"}. In: Workshop on Reinforcement Learning
  in Games, AAAI

\bibitem[{Ross and Bagnell(2010)}]{ross2010efficient}
Ross S, Bagnell D (2010) {Efficient Reductions for Imitation Learning}. In: Teh
  YW, Titterington M (eds) Proceedings of Machine Learning Research, JMLR
  Workshop and Conference Proceedings, Chia Laguna Resort, Sardinia, Italy,
  vol~9, pp 661--668,
  \urlprefix\url{http://proceedings.mlr.press/v9/ross10a.html}

\bibitem[{Salimans and Chen(2018)}]{mont-single-demo}
Salimans T, Chen R (2018) {Learning Montezuma's Revenge from a Single
  Demonstration}. arXiv preprint arXiv:181203381

\bibitem[{Schaul et~al.(2016)Schaul, Quan, Antonoglou, and Silver}]{PER}
Schaul T, Quan J, Antonoglou I, Silver D (2016) {Prioritized Experience
  Replay}. In: International Conference on Learning Representations,
  \urlprefix\url{http://arxiv.org/abs/1511.05952}

\bibitem[{Schrittwieser et~al.(2019)Schrittwieser, Antonoglou, Hubert,
  Simonyan, Sifre, Schmitt, Guez, Lockhart, Hassabis, Graepel et~al.}]{muzero}
Schrittwieser J, Antonoglou I, Hubert T, Simonyan K, Sifre L, Schmitt S, Guez
  A, Lockhart E, Hassabis D, Graepel T, et~al. (2019) {Mastering Atari, Go,
  Chess and Shogi by Planning with a Learned Model}. arXiv preprint
  arXiv:191108265

\bibitem[{Sinha et~al.(2020)Sinha, Song, Garg, and
  Ermon}]{sinha2020likelihood-exp}
Sinha S, Song J, Garg A, Ermon S (2020) {Experience Replay with Likelihood-free
  Importance Weights}. arXiv preprint arXiv:200613169

\bibitem[{{Sovrano}(2019)}]{ER-RND}
{Sovrano} F (2019) {Combining Experience Replay with Exploration by Random
  Network Distillation}. In: 2019 IEEE Conference on Games (CoG), pp 1--8,
  \doi{10.1109/CIG.2019.8848046}

\bibitem[{Stumbrys et~al.(2016)Stumbrys, Erlacher, and
  Schredl}]{Stumbrys2016EffectivenessOM}
Stumbrys T, Erlacher D, Schredl M (2016) {Effectiveness of Motor Practice in
  Lucid Dreams: a Comparison with Physical and Mental Practice}. Journal of
  Sports Sciences 34:27 -- 34

\bibitem[{Sutton and Barto(2018)}]{sutton2018}
Sutton RS, Barto AG (2018) {Reinforcement Learning: An Introduction}. MIT press

\bibitem[{Tang(2020)}]{tang2020self}
Tang Y (2020) {Self-Imitation Learning via Generalized Lower Bound Q-learning}.
  In: Advances in Neural Information Processing Systems, vol~33,
  \urlprefix\url{https://papers.nips.cc/paper/2020/file/a0443c8c8c3372d662e9173c18faaa2c-Paper.pdf}

\bibitem[{Tavakoli et~al.(2018)Tavakoli, Levdik, Islam, Smith, and
  Kormushev}]{tavakoli2018exploring}
Tavakoli A, Levdik V, Islam R, Smith CM, Kormushev P (2018) {Exploring Restart
  Distributions}. arXiv:181111298

\bibitem[{Wang et~al.(2017)Wang, Bapst, Heess, Mnih, Munos, Kavukcuoglu, and
  de~Freitas}]{ACER}
Wang Z, Bapst V, Heess NMO, Mnih V, Munos R, Kavukcuoglu K, de~Freitas N (2017)
  {Sample Efficient Actor-Critic with Experience Replay}. In: International
  Conference on Learning Representations,
  \urlprefix\url{https://openreview.net/pdf?id=HyM25Mqel}

\bibitem[{Wawrzy{\'n}ski(2009)}]{wawrzynski2009real}
Wawrzy{\'n}ski P (2009) {Real-time Reinforcement Learning by Sequential
  Actor-Critics and Experience Replay}. Neural Networks 22(10):1484--1497

\bibitem[{Zha et~al.(2019)Zha, Lai, Zhou, and Hu}]{ERO}
Zha D, Lai KH, Zhou K, Hu X (2019) {Experience Replay Optimization}. In:
  Proceedings of the Twenty-Eighth International Joint Conference on Artificial
  Intelligence, {IJCAI-19}, International Joint Conferences on Artificial
  Intelligence Organization, pp 4243--4249, \doi{10.24963/ijcai.2019/589},
  \urlprefix\url{https://doi.org/10.24963/ijcai.2019/589}

\bibitem[{Zhang and Sutton(2017)}]{CER}
Zhang S, Sutton RS (2017) {A Deeper Look at Experience Replay}. arXiv preprint
  arXiv:171201275

\bibitem[{Zhang et~al.(2020)Zhang, Bharti, Ma, Singla, and
  Zhu}]{zhang2020teachingdim}
Zhang X, Bharti SK, Ma Y, Singla A, Zhu X (2020) {The Teaching Dimension of
  {Q}-learning}. arXiv preprint arXiv:200609324

\end{thebibliography}


\clearpage
\newpage

\onecolumn
\appendix

\noindent \textbf{\LARGE{Appendices for ``Lucid Dreaming for Experience Replay: Refreshing Past States with the Current Policy''}}
\newline

We provide further details of our work in the following six appendices:
\begin{itemize}

\item Appendix \ref{sec:appe-parameters} contains the implementation details of LiDER, including neural network architecture, hyperparameters, and computation resources used for all experiments. 

\item Appendix \ref{sec:appe-a3csil} presents the pseudo-code for the A3C and SIL workers. Both follow the original work of \citet{a3c} and \citet{sil} respectively, we add them here for completeness. 

\item Appendix \ref{sec:appe-ttest} provides detailed statistics of the one-tailed independent-samples t-tests: 1) A3CTBSIL compared to LiDER, 2) A3CTBSIL compared to the three ablation studies of LiDER, 3) A3CTBSIL compared to the two extensions of LiDER, 4) LiDER compared to the three ablation studies of LiDER, and 5) LiDER compared to the two extensions of LiDER. 

\item Appendix \ref{sec:appe-reducesil} discusses the differences between the A3CTBSIL algorithm in \citet{de2019jointly} and the original SIL algorithm in \citet{sil} (as mentioned in Section \ref{sec:lider-a3c}). 

\item Appendix \ref{sec:appe-TAmodels} presents the performance of the trained agents (TA) used in LiDER-TA. 

\item Appendix \ref{sec:appe-pretrain} details the pre-training process for obtaining the BC models used in LiDER-BC, including the statistics of the demonstration collected by \citet{de2019pre}, the network architecture, the hyperparamters used for pre-training, and the performance of the trained BC models.  

\end{itemize}

\section{Implementation details} 
\label{sec:appe-parameters}
We use the same neural network architecture as in the original A3C algorithm \cite{a3c} for all A3C, SIL, and refresher workers (the blue, orange, and green components in Figure \ref{fig:arch} respectively). The network consists of three convolutional layers, one fully connected layer, followed by two branches of a fully connected layer: a policy function output layer and a value function output layer. Atari images are converted to grayscale and resized to 88$\times$88 with 4 images stacked as the input. 

We run each experiment for eight trials due to computation limitations. Each experiment uses one GPU (Tesla K80 or TITAN V), five CPU cores, and 40 GB of memory (each LiDER-OneBuffer experiment uses 64 GB of memory since the buffer size was doubled). The refresher worker runs on GPU to generate data as quickly as possible; the A3C and SIL workers run distributively on CPU cores. In all games, the wall-clock time is roughly 0.8 to 1 million steps per hour and around 50 to 60 hours to complete one trial of 50 million steps. 

The baseline A3CTBSIL is trained with 17 parallel workers; 16 A3C workers and 1 SIL worker. The RMSProp optimizer is used with a learning rate = 0.0007. We use $t_{max}=20$ for $n$-step bootstrap $Q^{(n)}$ ($n \leq t_{max}$). The SIL worker performs $M=4$ SIL policy updates (Equation \eqref{eq:sil_update}) per step $t$ with minibatch size 32 (i.e., 32$\times$4=128 total samples per step). Buffer $\D$ is of size $10^5$. The SIL loss weight $\beta^{sil}=0.5$.

LiDER is also trained with 17 parallel workers: 15 A3C workers, 1 SIL worker, and 1 refresher worker---we keep the total number of workers in A3CTBSIL and LiDER the same to ensure a fair performance comparison. The SIL worker in LiDER also uses a minibatch size of 32, samples are taken from buffer $\D$ and $\R$ as described in Section \ref{sec:lider}. All other parameters are identical to that of A3CTBSIL. We summarize the details of the network architecture and experiment parameters in Table \ref{table:hyperparameters}.

\begin{table}[th]
    \caption{Hyperparameters for all experiments. We train each game for 50 million steps with a frame skip of 4, i.e., 200 million game frames were consumed for training.}
    \label{table:hyperparameters}
    \centering
    \adjustbox{max width=\textwidth}{
        \begin{tabular}{r|r}
        \textbf{Network Architecture}        & \textbf{Value} \\ \hline \hline
        Input size             & 88$\times$88$\times$4 \\ \hline
        Tensorflow Padding method         & SAME     \\ \hline 
        Convolutional layer 1 & 32 filters of size 8$\times$8 with stride 4 \\ \hline 
        Convolutional layer 2 & 64 filters of size 4$\times$4 with stride 2 \\ \hline
        Convolutional layer 3 & 64 filters of size 3$\times$3 with stride 1 \\ \hline
        Fully connected layer & 512 \\ \hline
        Policy output layer & number of actions \\ \hline
        Value output layer & 1 \\ \hline \hline
        \multicolumn{2}{c}{\textbf{Common Parameters}} \\ \hline \hline
        RMSProp initial learning rate  & $7 \times 10^{-4}$  \\ \hline
        RMSProp epsilon        & $1 \times 10^{-5}$  \\ \hline
        RMSProp decay          & 0.99      \\ \hline 
        RMSProp momentum       & 0      \\ \hline 
        Maximum gradient norm  & 0.5     \\ \hline
        Discount factor $\gamma$ & 0.99 \\ \hline
        \multicolumn{2}{c}{\textbf{Parameters for A3CTB}} \\ \hline \hline
        A3C entropy regularizer weight $\beta^{a3c}$  & 0.01          \\ \hline
        A3C maximum bootstrap step $t_{max}$  & 20          \\ \hline
        A3C value loss weight $\alpha$  & 0.5          \\ \hline
        $k$ parallel actors  & 16      \\ \hline
        Transformed Bellman operator $\varepsilon$ & $10^{-2}$ \\ \hline 
        \multicolumn{2}{c}{\textbf{Parameters for SIL}} \\ \hline \hline
        SIL value loss weight $\beta^{sil}$ & $0.1$ \\ \hline
        SIL update per step M & 4 \\ \hline
        Replay buffer $\D$ size & $10^5$\\ \hline
        Replay buffer priority $\alpha$ & $0.6$ \\ \hline
        Minibatch size & 32 \\ \hline
        \multicolumn{2}{c}{\textbf{Parameters for LiDER (refresher worker)}} \\ \hline \hline
        Replay buffer $\R$ size & $10^5$\\ \hline
        Minibatch size & 32 \\ \hline
        \end{tabular}
    }
\end{table}

\clearpage
\newpage
\section{Pseudo-code for the A3C and SIL workers}
\label{sec:appe-a3csil}
\begin{algorithm*}
    \caption{LiDER: A3C Worker (as in \citet{a3c})}
    \label{alg:a3cworker}
    \begin{algorithmic}[1]
    \State \textit{// Assume global network parameters $\theta$ and $\theta_{v}$ and global step $T=0$}
    \State \textit{// Assume replay buffer $\D \leftarrow \emptyset$, replay buffer $\R \leftarrow \emptyset$}
    \State Initialize worker-specified local network parameters, $\theta^{'}, \theta_{v}^{'}$
    \State Initialize worker-specified local time step $t=0$ and local episode buffer $\mathcal{E} \leftarrow \emptyset$
    \While{$T<T_{max}$} \Comment{$T_{max}$ = 50 million}
        \State Reset gradients: $d\theta \leftarrow 0$, $d\theta_{v} \leftarrow 0$
        \State Synchronize local parameters with global parameters $\theta^{'} \leftarrow \theta$ and $\theta_{v}^{'} \leftarrow \theta_{v}$
        \State $t_{start} \leftarrow t$
        \While{$s_{t+1}$ is not terminal or $t < t_{max}$} \Comment{$t_{max}=20$}
            \State Execute an action $s_t$, $a_t$, $r_t$, $s_{t+1}$ $\sim$ $\pi(a_t|s_t, \theta^{'})$
            \State Store transition to local buffer: $\mathcal{E} \leftarrow \mathcal{E} \cup$ \{$s_t$, $a_t$, $r_t$, \}
            \State $T \leftarrow T+1$, $t \leftarrow t+1$
        \EndWhile
        \State
            G $\leftarrow$ 
            $\begin{cases}
                0 & \text{if $s_{t+1}$ is terminal} \\
                V(S_{t+1};\theta_{v}^{'}) & \text{otherwise}
            \end{cases}$ \Comment{Perform A3C update \cite{a3c}}
        \For{$i \in \{t,..., t_{start}\}$}
            \State $G \leftarrow r_{i} + \gamma G$
            \State Accumulate gradients w.r.t. $\theta^{'}$:
            $d\theta \leftarrow d\theta + \nabla_{\theta^{'}}log\pi(a_i|s_i, \theta^{'})(G-V(s_i; \theta_{v}^{'}))$
            \State Accumulate gradients w.r.t. $\theta_{v}^{'}$:
            $d\theta_v \leftarrow d\theta_v + \partial(G-V(s_i; \theta_{v}^{'}))^{2}/\partial \theta_{v}^{'}$
        \EndFor
       
        \If{$s_{t+1}$ is terminal}:  \Comment{Prepare for SIL worker \cite{sil}}
            \State compute $G_{t} = \sum_{k}^{\infty} \gamma^{k-t}r_{k}$ for all $t$ in $\mathcal{E}$
            \State Store transition to global replay buffer $\D \leftarrow \D\cup \{s_t, a_t, G_t\}$ for all $t$ in $\mathcal{E}$
            \State Reset local buffer $\mathcal{E} \leftarrow \emptyset$
        \EndIf
    \State Asynchronously update global parameters using local parameters
    \EndWhile
    \end{algorithmic}
\end{algorithm*}

\begin{algorithm*}
    \caption{LiDER: SIL Worker (as in \citet{sil})}
    \label{alg:silworker}
    \begin{algorithmic}[1]
        \State \textit{// Assume global network parameters, $\theta$, $\theta_v$}
        \State \textit{// Assume (Non-empty) replay buffer $\D$, replay buffer $\R$} 
        \State Initialize worker-specific local network parameters, $\theta^{'}$, $\theta_{v}^{'}$
        \State Initialize local buffer $\mathcal{B} \leftarrow \emptyset$
        \While{$T<T_{max}$} \Comment{$T_{max}$ = 50 million}
            \State Synchronize global step $T$ from the most recent A3C worker
            \State Synchronize parameters $\theta^{'} \leftarrow \theta$ and $\theta_{v}^{'} \leftarrow \theta_v$
            \For{$m=1$ to $M$} \Comment{$M=4$}
                \State Sample a minibatch of size 32 $\{s_D, a_D, G_D\}$ from $\D$
                \State Sample a minibatch of size 32 $\{s_R, a_R, G_R\}$ from $\R$
                \State Store both batches into $\mathcal{B}$: $\mathcal{B} \leftarrow$ \{$s_D$, $a_D$, $r_D$\} $\cup$ \{$s_R$, $a_R$, $r_R$\} \Comment{Length of $\mathcal{B}$=64}
                \State Sample a minibatch of 32 $\{s_B, a_B, G_B\}$ from $\mathcal{B}$ \Comment{Perform SIL update \cite{sil}}
                \State Compute gradients w.r.t. $\theta^{'}: d\theta \leftarrow \nabla_{\theta^{'}} \text{log}\pi(a_B|s_B;\theta^{'})(G_B-V(s_B;\theta_{v}^{'}))_{+}$ 
                \State Compute gradients w.r.t. $\theta_{v}^{'}: d\theta_v \leftarrow \partial((G_B-V(s_B;\theta_{v}^{'}))_{+})^2 / \partial\theta_{v}^{'}$ 
                \State Perform asynchronous update of $\theta$ using $d\theta$ and $\theta_v$ using $d\theta_v$ 
                \State Reset local buffer $\mathcal{B} \leftarrow \emptyset$
            \EndFor
        \EndWhile
    \end{algorithmic}
\end{algorithm*}

\clearpage
\newpage
\section{One-tailed independent-samples t-tests}
\label{sec:appe-ttest}
We conducted one-tailed independent-samples t-tests (equal variances not assumed) in all games to compare the differences in the mean episodic reward among all methods in this paper. For each game, we restored the best model checkpoint from each trial (eight trials per method) and executed the model in the game following a deterministic policy for 100 episodes (an episode ends when the agent loses all its lives) and recorded the reward per episode. This gives us 800 data points for each method in each game. We use a significance level $\alpha=0.001$ for all tests.

First, we check the statistical significance of the baseline A3CTBSIL compared to LiDER (Section \ref{sec:lider-a3c}), the main framework proposed in this paper. We report the detailed statistics in Table \ref{table:ttest-base-lider}. Results show that the mean episodic reward of LiDER is significantly higher than A3CTBSIL ($p \ll 0.001$) in all games. 

\begin{table}[th]
    \caption{One-tailed independent-samples t-test for the differences of the mean episodic reward between A3CTBSIL and LiDER. Equal variances are not assumed.}
    \label{table:ttest-base-lider}
    \centering
    \adjustbox{max width=\textwidth}{
        \begin{tabular}{l|c|c|c}
        \hline \hline
        \multirow{2}{*}{Methods} & Mean episodic reward & \multirow{2}{*}{Standard deviation} & \multirow{2}{*}{One-tailed p-value}  \\ 
         & (800 episodes) & & \\ \hline \hline
         
        \multicolumn{4}{c}{\textbf{Gopher}} \\ \hline \hline
        A3CTBSIL    & 4291.20   & 2913.52   & - \\ \hline
        LiDER       & 6618.88   & 3300.10   &  1.24$\times 10^{-47}$ \\\hline \hline
        
        \multicolumn{4}{c}{\textbf{NameThisGame}} \\ \hline \hline
        A3CTBSIL    & 6786.75   & 1275.87   & - \\ \hline
        LiDER       & 8332.50   & 1754.30   & 4.09$\times 10^{-80}$ \\\hline \hline
        
        \multicolumn{4}{c}{\textbf{Alien}} \\ \hline \hline
        A3CTBSIL    & 3558.58   & 1596.18   & - \\ \hline
        LiDER       & 5065.04   & 2012.93   & 3.77$\times 10^{-57}$ \\ \hline \hline
        
        \multicolumn{4}{c}{\textbf{Ms.~Pac-Man}} \\ \hline \hline
        A3CTBSIL    & 4975.03   & 1527.05   & - \\ \hline
        LiDER       & 8532.34   & 2477.02   & 1.49$\times 10^{-187}$ \\\hline \hline
        
        \multicolumn{4}{c}{\textbf{Freeway}} \\ \hline \hline
        A3CTBSIL    & 23.10     & 5.84      &  - \\ \hline
        LiDER       & 31.62     & 0.98      &  1.19$\times 10^{-201}$ \\ \hline \hline
        
        \multicolumn{4}{c}{\textbf{Montezuma's Revenge}} \\ \hline \hline
        A3CTBSIL    & 0.25      & 4.99      & - \\ \hline
        LiDER       & 987.63    & 951.69    &  3.36$\times 10^{-129}$ \\ \hline \hline
        \end{tabular}
    }
\end{table}

\newpage
Second, we compare A3CTBSIL to the three ablation studies, LiDER-AddAll, LiDER-OneBuffer, and LiDER-SampleR (Section \ref{sec:ablation}). Table \ref{table:ttest-base-ablation} shows that all ablations were helpful in Freeway and Montezuma's Revenge, in which the mean episodic rewards of the ablations are significantly higher than the baseline ($p \ll 0.001$). 
LiDER-AddAll also performed significantly better than A3CTBSIL in all games ($p \ll 0.001$). 
LiDER-OneBuffer outperformed A3CTBSIL in Freeway and Montezuma's Revenge ($p \ll 0.001$), 
but it performed worse than the other four games 
($p \ll 0.001$). 
LiDER-SampleR outperformed A3CTBSIL in Ms.~Pac-Man, Freeway, and Montezuma's Revenge ($p \ll 0.001$), but under-performed A3CTBSIL in Gopher, NameThisGame, and Alien ($p \ll 0.001$). 

\begin{table}[th]
    \caption{One-tailed independent-samples t-test for the differences of the mean episodic reward between A3CTBSIL and LiDER-AddALL, between A3CTBSIL and LiDER-OneBuffer, and between A3CTBSIL and LiDER-SampleR. Equal variances are not assumed.}
    \label{table:ttest-base-ablation}
    \centering
    \adjustbox{max width=\textwidth}{
        \begin{tabular}{l|c|c|c}
        \hline \hline
        \multirow{2}{*}{Methods} & Mean episodic reward & \multirow{2}{*}{Standard deviation} & \multirow{2}{*}{One-tailed p-value}  \\ 
         & (800 episodes) & & \\ \hline \hline
         
        \multicolumn{4}{c}{\textbf{Gopher}} \\ \hline \hline
        A3CTBSIL                            & 4291.20   & 2913.52   & - \\ \hline
        (Ablation) LiDER-AddAll             & 7086.53   & 3188.04   & 2.72$\times 10^{-68}$ \\ \hline 
        (Ablation) LiDER-OneBuffer          & 1962.05   & 1872.94   & 5.97$\times 10^{-72}$ \\ \hline
        (Ablation) LiDER-SampleR            & 3072.40   & 5146.23   & 3.61$\times 10^{-9}$ \\ \hline \hline
        
        \multicolumn{4}{c}{\textbf{NameThisGame}} \\ \hline \hline
        A3CTBSIL                            & 6786.75   & 1275.87   & - \\ \hline
        (Ablation) LiDER-AddAll             & 8200.04   & 1580.23   & 2.86$\times 10^{-77}$ \\ \hline 
        (Ablation) LiDER-OneBuffer          & 6422.48   & 1374.87   & 2.34$\times 10^{-8}$ \\ \hline
        (Ablation) LiDER-SampleR            & 5819.81   & 1743.05   & 3.35$\times 10^{-35}$ \\ \hline \hline
        
        \multicolumn{4}{c}{\textbf{Alien}} \\ \hline \hline
        A3CTBSIL                            & 3558.58   & 1596.18   & - \\ \hline
        (Ablation) LiDER-AddAll             & 4054.28   & 1837.20   & 5.13$\times 10^{-9}$ \\ \hline 
        (Ablation) LiDER-OneBuffer          & 3204.41   & 1998.95   & 4.74$\times 10^{-5}$ \\ \hline
        (Ablation) LiDER-SampleR            & 3104.99   & 1548.04   & 4.86$\times 10^{-9}$ \\ \hline \hline
        
        \multicolumn{4}{c}{\textbf{Ms.~Pac-Man}} \\ \hline \hline
        A3CTBSIL                            & 4975.03   & 1527.05   & - \\ \hline
        (Ablation) LiDER-AddAll             & 6828.82   & 2562.33   & 1.81$\times 10^{-62}$ \\ \hline 
        (Ablation) LiDER-OneBuffer          & 4625.37   & 1920.67   & 2.95$\times 10^{-5}$ \\ \hline 
        (Ablation) LiDER-SampleR            & 7303.22   & 1869.98   & 4.32$\times 10^{-134}$ \\ \hline \hline
        
        \multicolumn{4}{c}{\textbf{Freeway}} \\ \hline \hline
        A3CTBSIL                            & 23.10     & 5.84      &  - \\ \hline
        (Ablation) LiDER-AddAll             & 31.20     & 0.99      & 1.35$\times 10^{-189}$ \\ \hline
        (Ablation) LiDER-OneBuffer          & 27.55     & 5.15      & 6.34$\times 10^{-55}$ \\ \hline
        (Ablation) LiDER-SampleR            & 27.45     & 10.46     & 4.17$\times 10^{-24}$ \\ \hline \hline
        
        \multicolumn{4}{c}{\textbf{Montezuma's Revenge}} \\ \hline \hline
        A3CTBSIL                            & 0.25      & 4.99      & - \\ \hline
        (Ablation) LiDER-AddAll             & 77.63     & 144.18    & 3.93$\times 10^{-46}$ \\ \hline
        (Ablation) LiDER-OneBuffer          & 3.00      & 24.31     & 8.97$\times 10^{-4}$ \\ \hline 
        (Ablation) LiDER-SampleR            & 265.86    & 178.74    & 8.46$\times 10^{-205}$ \\ \hline \hline
        \end{tabular}
    }
\end{table}

\newpage 
Third, we compare A3CTBSIL to the two extensions, LiDER-BC and LiDER-TA (Section \ref{sec:lider-ta-bc}). Table \ref{table:ttest-base-extension} shows that the two extensions outperformed the baseline significantly in all games ($p \ll 0.001$).

\begin{table}[th]
    \caption{One-tailed independent-samples t-test for the differences of the mean episodic reward between A3CTBSIL and LiDER-BC, and between A3CTBSIL and LiDER-TA. Equal variances are not assumed.}
    \label{table:ttest-base-extension}
    \centering
    \adjustbox{max width=\textwidth}{
        \begin{tabular}{l|c|c|c}
        \hline \hline
        \multirow{2}{*}{Methods} & Mean episodic reward & \multirow{2}{*}{Standard deviation} & \multirow{2}{*}{One-tailed p-value}  \\ 
         & (800 episodes) & & \\ \hline \hline
        
        \multicolumn{4}{c}{\textbf{Gopher}} \\ \hline \hline
        A3CTBSIL                & 4291.20   & 2913.52   & - \\ \hline
        (Extension) LiDER-TA    & 8133.50   & 3800.38   & 1.97$\times 10^{-98}$ \\ \hline 
        (Extension) LiDER-BC    & 7775.75   & 3480.92   & 1.11$\times 10^{-91}$ \\ \hline \hline 
        
        \multicolumn{4}{c}{\textbf{NameThisGame}} \\ \hline \hline
        A3CTBSIL                & 6786.75   & 1275.87   & - \\ \hline
        (Extension) LiDER-TA    & 10227.69  & 2222.20   & 7.63$\times 10^{-212}$ \\ \hline 
        (Extension) LiDER-BC    & 7303.74   & 1649.01   & 1.81$\times 10^{-12}$ \\ \hline \hline 
        
        \multicolumn{4}{c}{\textbf{Alien}} \\ \hline \hline
        A3CTBSIL                & 3558.58   & 1596.18   & - \\ \hline
        (Extension) LiDER-TA    & 7753.54   & 1681.06   & 0.000 \\ \hline 
        (Extension) LiDER-BC    & 6261.79   & 1865.67   & 4.01$\times 10^{-166}$ \\ \hline \hline 
        
        \multicolumn{4}{c}{\textbf{Ms.~Pac-Man}} \\ \hline \hline
        A3CTBSIL                & 4975.03   & 1527.05   & - \\ \hline
        (Extension) LiDER-TA    & 10272.18  & 2035.98   & 0.000 \\ \hline
        (Extension) LiDER-BC    & 9613.89   & 2875.71   & 2.40$\times 10^{-226}$ \\ \hline \hline 
        
        \multicolumn{4}{c}{\textbf{Freeway}} \\ \hline \hline
        A3CTBSIL                & 23.10     & 5.84      &  - \\ \hline
        (Extension) LiDER-TA    & 32.42     & 0.73      & 2.81$\times 10^{-223}$ \\ \hline 
        (Extension) LiDER-BC    & 31.68     & 0.85      & 4.63$\times 10^{-203}$ \\ \hline \hline 
        
        \multicolumn{4}{c}{\textbf{Montezuma's Revenge}} \\ \hline \hline
        A3CTBSIL                & 0.25      & 4.99      & - \\ \hline
        (Extension) LiDER-TA    & 1677.50   & 1050.33   & 2.53$\times 10^{-222}$ \\ \hline 
        (Extension) LiDER-BC    & 1811.86   & 994.38    & 2.30$\times 10^{-256}$ \\ \hline \hline 
        \end{tabular}
    }
\end{table}

\newpage 
Fourth, we check the statistical significance of LiDER compared to the three ablation studies, LiDER-AddAll, LiDER-OneBuffer, and LiDER-SampleR (Section \ref{sec:ablation}). Results in Table \ref{table:ttest-lider-ablation} show that most of the ablations significantly under-performed LiDER ($p \ll 0.001$) in terms of the mean episodic reward. Except for Gopher and NameThisGame, in which LiDER-AddAll performs at the same level as LiDER ($p > 0.001$). 

\begin{table}[th]
    \caption{One-tailed independent-samples t-test for the differences of the mean episodic reward between LiDER and LiDER-AddAll, between LiDER and LiDER-OneBuffer, and between LiDER and LiDER-SampleR. Equal variances are not assumed. Methods in bold are \textbf{not} significant at level $\alpha=0.001$.}
    \label{table:ttest-lider-ablation}
    \centering
    \adjustbox{max width=\textwidth}{
        \begin{tabular}{l|c|c|c}
        \hline \hline
       \multirow{2}{*}{Methods} & Mean episodic reward & \multirow{2}{*}{Standard deviation} & \multirow{2}{*}{One-tailed p-value}  \\ 
         & (800 episodes) & & \\ \hline \hline
        
        \multicolumn{4}{c}{\textbf{Gopher}} \\ \hline \hline
        LiDER                               & 6618.88   & 3300.10   & - \\ \hline
        \textbf{(Ablation) LiDER-AddAll}    & 7086.53   & 3188.04   & \textbf{0.002} \\ \hline 
        (Ablation) LiDER-OneBuffer          & 1962.05   & 1872.94   & 3.65$\times 10^{-186}$ \\ \hline
        (Ablation) LiDER-SampleR            & 3072.40   & 5146.23   & 1.38$\times 10^{-55}$ \\ \hline \hline
        
        \multicolumn{4}{c}{\textbf{NameThisGame}} \\ \hline \hline
        LiDER                               & 8332.50   & 1754.30   & - \\ \hline
        \textbf{(Ablation) LiDER-AddAll}    & 8200.04   & 1580.23   & \textbf{0.056} \\ \hline 
        (Ablation) LiDER-OneBuffer          & 6422.48   & 1374.87   & 4.25$\times 10^{-110}$ \\ \hline
        (Ablation) LiDER-SampleR            & 5819.81   & 1743.05   & 6.54$\times 10^{-147}$ \\ \hline \hline
        
        \multicolumn{4}{c}{\textbf{Alien}} \\ \hline \hline
        LiDER                       & 5065.04   & 2012.93   & - \\ \hline 
        (Ablation) LiDER-AddAll     & 4054.28   & 1837.20   & 3.28$\times 10^{-25}$ \\ \hline 
        (Ablation) LiDER-OneBuffer  & 3204.41   & 1998.95   & 5.92$\times 10^{-70}$ \\ \hline 
        (Ablation) LiDER-SampleR    & 3104.99   & 1548.04   & 3.55$\times 10^{-92}$ \\ \hline \hline 
        
        \multicolumn{4}{c}{\textbf{Ms.~Pac-Man}} \\ \hline \hline
        LiDER                       & 8532.34   & 2477.02   &  - \\ \hline 
        (Ablation) LiDER-AddAll     & 6828.82   & 2562.33   & 9.06$\times 10^{-40}$  \\ \hline 
        (Ablation) LiDER-OneBuffer  & 4625.37   & 1920.67   & 4.18$\times 10^{-199}$ \\ \hline 
        (Ablation) LiDER-sampleR    & 7303.22   & 1869.98   & 2.76$\times 10^{-28}$  \\ \hline \hline 
        
        \multicolumn{4}{c}{\textbf{Freeway}} \\ \hline \hline
        LiDER                       & 31.62     & 0.98      &  - \\ \hline 
        (Ablation) LiDER-AddAll     & 31.20     & 0.99      & 1.32$\times 10^{-17}$ \\ \hline
        (Ablation) LiDER-OneBuffer  & 27.55     & 5.15      & 1.62$\times 10^{-85}$ \\ \hline 
        (Ablation) LiDER-SampleR    & 27.45     & 10.46     & 1.22$\times 10^{-27}$\\ \hline \hline 
        
        \multicolumn{4}{c}{\textbf{Montezuma's Revenge}} \\ \hline \hline
        LiDER                       & 987.63    & 951.69    &  - \\ \hline
        (Ablation) LiDER-AddAll     & 77.63     & 144.18    & 1.68$\times 10^{-114}$ \\ \hline 
        (Ablation) LiDER-OneBuffer  & 3.00      & 24.31     & 1.09$\times 10^{-128}$ \\ \hline 
        (Ablation) LiDER-SampleR    & 265.86    & 178.74    & 5.31$\times 10^{-80}$ \\ \hline \hline
        \end{tabular}
    }
\end{table}

\newpage
Lastly, we compare LiDER to the two extensions, LiDER-TA and LiDER-BC (Section \ref{sec:lider-ta-bc}). Results in Table \ref{table:ttest-lider-tabc} show that LiDER-TA always outperforms LiDER ($p \ll 0.001$). LiDER-BC outperformed LiDER in Gopher, Alien, Ms.~Pac-Man, and Montezuma's Revenge. In Freeway, LiDER-BC performs the same as LiDER ($p > 0.001$), while in NameThisGame LiDER-BC performed worse than LiDER ($p \ll 0.001$).  

\begin{table}[th]
    \caption{One-tailed independent-samples t-test for the differences of the mean episodic reward between LiDER and LiDER-TA, and between LiDER and LiDER-BC. Equal variances are not assumed. Methods in bold are \textbf{not} significant at level $\alpha=0.001$.}
    \label{table:ttest-lider-tabc}
    \centering
    \adjustbox{max width=\textwidth}{
        \begin{tabular}{l|c|c|c}
        \hline \hline
       \multirow{2}{*}{Methods} & Mean episodic reward & \multirow{2}{*}{Standard deviation} & \multirow{2}{*}{One-tailed p-value}  \\ 
         & (800 episodes) & & \\ \hline \hline
         
        \multicolumn{4}{c}{\textbf{Gopher}} \\ \hline \hline
        LiDER                   & 6618.86   & 3300.10   & - \\ \hline
        (Extension) LiDER-TA    & 8133.50   & 3800.38   & 2.07$\times 10^{-17}$ \\ \hline 
        (Extension) LiDER-BC    & 7775.75   & 3480.92   & 6.55$\times 10^{-12}$ \\ \hline \hline 
        
        \multicolumn{4}{c}{\textbf{NameThisGame}} \\ \hline \hline
        LiDER                   & 8332.50   & 1754.30   & - \\ \hline
        (Extension) LiDER-TA    & 10227.69  & 2222.20   & 3.75$\times 10^{-72}$ \\ \hline 
        (Extension) LiDER-BC    & 7303.74   & 1649.01   & 1.68$\times 10^{-32}$ \\ \hline \hline 
         
        \multicolumn{4}{c}{\textbf{Alien}} \\ \hline \hline
        LiDER                   & 5065.04   & 2012.93   & - \\ \hline 
        (Extension) LiDER-TA    & 7753.54   & 1681.06   & 3.53$\times 10^{-148}$ \\ \hline 
        (Extension) LiDER-BC    & 6261.79   & 1865.67   & 1.05$\times 10^{-33}$ \\ \hline \hline 
        
        \multicolumn{4}{c}{\textbf{Ms.~Pac-Man}} \\ \hline \hline
        LiDER                   & 8532.34   & 2477.02   &  - \\ \hline 
        (Extension) LiDER-TA    & 10272.18  & 2035.98   & 8.01$\times 10^{-50}$ \\ \hline 
        (Extension) LiDER-BC    & 9613.89   & 2875.71   & 7.81$\times 10^{-16}$ \\ \hline \hline
        
        \multicolumn{4}{c}{\textbf{Freeway}} \\ \hline \hline
        LiDER                           & 31.62     & 0.98  &  - \\ \hline 
        (Extension) LiDER-TA            & 32.42     & 0.73  & 8.54$\times 10^{-69}$ \\ \hline
        \textbf{(Extension) LiDER-BC}   & 31.68     & 0.85  & \textbf{0.104} \\ \hline \hline 
       
        \multicolumn{4}{c}{\textbf{Montezuma's Revenge}} \\ \hline \hline
        LiDER                   & 987.63    & 951.69    &  - \\ \hline
        (Extension) LiDER-TA    & 1677.50   & 1050.33   & 4.55$\times 10^{-41}$ \\ \hline 
        (Extension) LiDER-BC    & 1811.88   & 994.38    & 1.53$\times 10^{-59}$ \\ \hline \hline
        \end{tabular}
    }
\end{table}

\newpage
\section{Differences between A3CTBSIL and SIL}
\label{sec:appe-reducesil}
There is a performance difference in Montezuma's Revenge between the A3CTBSIL algorithm (our previous work in \citet{de2019jointly}, which is used as the baseline method in this article) and the original SIL algorithm (by \citet{sil}). The A3CTBSIL agent fails to achieve any reward while the SIL agent can achieve a score of 1100 (Table 5 in \cite{sil}). 

We hypothesize that the difference is due to the different number of SIL updates (Equation \eqref{eq:sil_update}) that can be performed in A3CTBSIL and SIL; lower numbers of SIL updates would decrease the performance. In particular, \citet{sil} proposed to add the ``Perform self-imitation learning'' step in \emph{each} A3C worker (Algorithm 1 of \citet{sil}). That is, when running with 16 A3C workers, the SIL agent is actually using 16 SIL workers to update the policy. However, A3CTBSIL only has one SIL worker, which means A3CTBSIL performs strictly fewer SIL updates compared to that of the original SIL algorithm, and thus resulting in lower performance. 

We empirically validate the above hypothesis by conducting an experiment in the game of Ms.~Pac-Man by modifying the A3CTBSIL algorithm from our previous work \cite{de2019jointly}. Instead of performing a SIL update whenever the SIL worker can, we force the SIL worker to only perform an update at even global steps; this setting reduces the total number of SIL updates by half. We denote this experiment as A3CTBSIL-ReduceSIL. 

\begin{figure}[th]
\centering
  \includegraphics[width=0.48\textwidth]{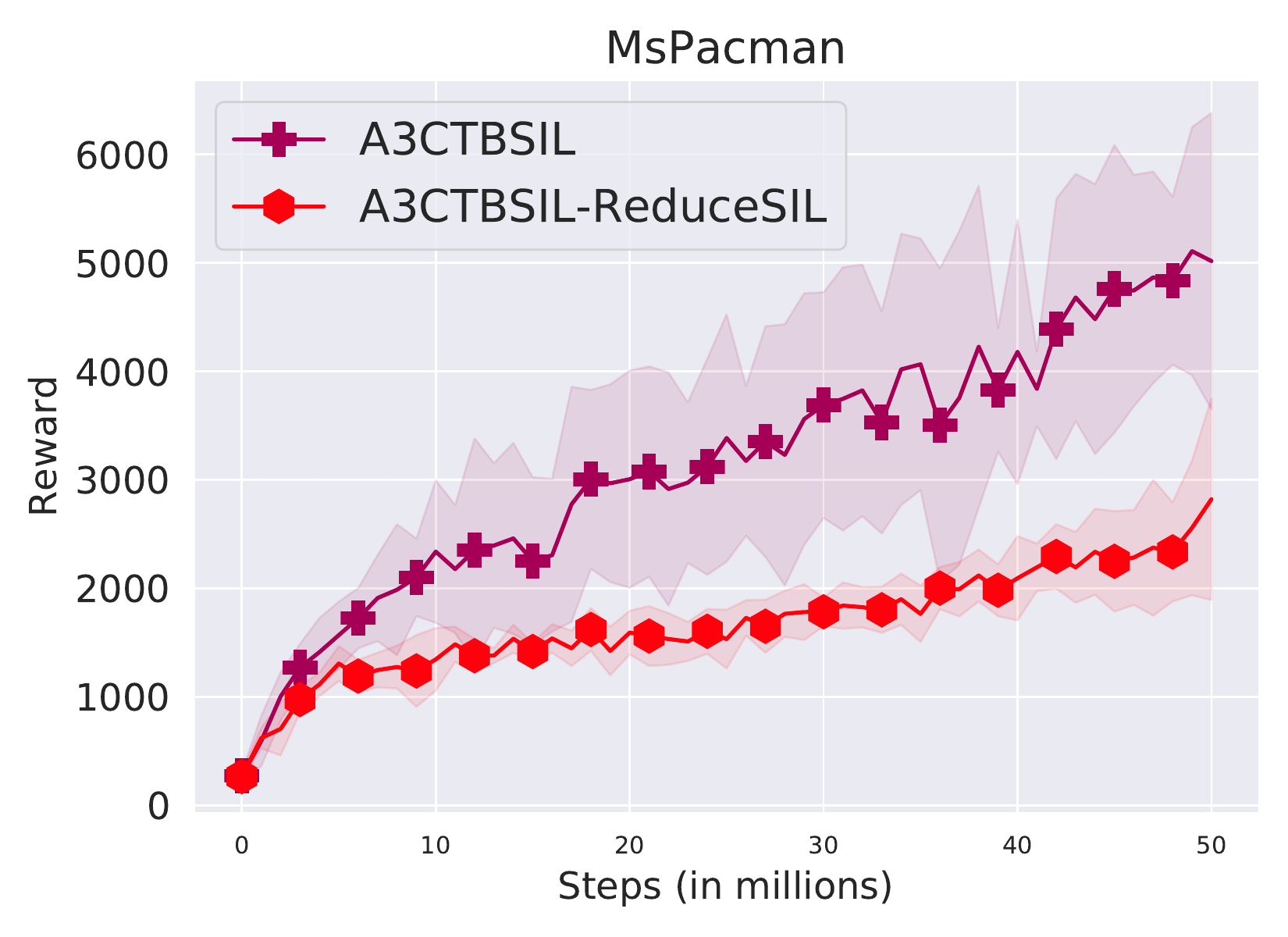}
  \caption{A3CTBSIL-ReduceSIL compared to A3CTBSIL in the game of Ms.~Pac-Man. The x-axis is the total number of environmental steps. The y-axis is the average testing score over five trials. We ran A3CTBSIL-ReduceSIL for five trials due to limited computing resources; we plot the first five trials out of eight for A3CTBSIL for a fair comparison to the number of trials in A3CTBSIL-ReduceSIL. Shaded regions show the standard deviation.} 
  \label{fig:reducesil}
 \end{figure}

Figure \ref{fig:reducesil} shows that A3CTBSIL-ReduceSIL under-performed A3CTBSIL, which provides preliminary evidence that the number of SIL updates is positively correlated to performance. More experiments will be performed in future work to further validate this correlation. 

\section{The performance of trained agents used in LiDER-TA}
\label{sec:appe-TAmodels}
Section \ref{sec:lider-ta-bc} shows that LiDER can leverage knowledge from a trained agent (TA). While the TA could come from any source, we use the best checkpoint of a fully trained LiDER agent. Table \ref{table:ta} shows the average performance of the TA used in each game. The score is estimated by executing the TA greedily in the game for 50 episodes. An episode ends when the agent loses all its lives.

\begin{table}[h]
    \caption{The performance of trained agents used in LiDER-TA, shown as the purple dotted line in Figure \ref{fig:lider-ta-bc}. The score is estimated by executing the TA greedily in the game for 50 episodes.}
    \label{table:ta}
    \centering
    \adjustbox{max width=.8\textwidth}{
        \begin{tabular}{l|c|c}
        \hline \hline
        Game                 & Trained TA score & standard deviation \\\hline\hline
        Gopher               & 6972.4    & 2190.26             \\ \hline
        NameThisGame         & 9969.0   & 1910.91             \\ \hline
        Alien                & 7190.4    & 1251.27             \\ \hline
        Ms.~Pac-Man          & 9145.42   & 955.94             \\ \hline
        Freeway              & 32.92    & 0.27               \\ \hline
        Montezuma's Revenge  & 1108.0    &  1057.14           \\ \hline
        \end{tabular}
    }
\end{table}

\newpage
\section{Pre-training the behavior cloning model for LiDER-BC}
\label{sec:appe-pretrain}
In Section \ref{sec:lider-ta-bc}, we demonstrated that a BC model can be incorporated into LiDER to improve learning. The BC model is pre-trained using a publicly available human demonstration dataset. Dataset statistics are shown in Table \ref{table:demo}.  

\begin{table}[h]
    \caption{Demonstration size and quality, collected in \citet{de2019pre}. All games are limited to 20 minutes of demonstration time per episode.}
    \label{table:demo}
    \centering
    \adjustbox{max width=.8\textwidth}{
        \begin{tabular}{l|c|c|c|c}
        \hline \hline
        Game                 & Worst score  & Best score    & \# of states & \# of episodes \\\hline\hline
        Gopher               & 1420         & 5800          & 16847        &  8             \\ \hline
        NameThisGame         & 2510         & 4840          & 17113        &  4             \\ \hline
        Alien                & 3000         & 8240          & 12885        &  5             \\ \hline
        Ms.~Pac-Man          & 4020         & 18241         & 14504        &  8             \\ \hline
        Freeway              & 26           & 31            & 24396        &  12            \\ \hline
        Montezuma's Revenge  & 500          & 10100         & 18751        &  9             \\ \hline
        \end{tabular}
    }
\end{table}

The BC model uses the same network architecture as the A3C algorithm \cite{a3c} and pre-training a BC model for A3C requires a few more steps than just using supervised learning as to how it is normally done in standard imitation learning (e.g., \citet{ross2010efficient}). A3C has two output layers: a policy output layer and a value output layer. The policy output is what we usually train a supervised classifier for. However, the value output layer is usually initialized randomly without being pre-trained. Our previous work \cite{de2019jointly} observed this inconsistency and leveraged demonstration data to also pre-train the value output layer. In particular, since the demonstration data contains the true return $\G$, we can obtain a value loss that is almost identical to A3C's value loss $L^{a3c}_{value}$: instead of using the n-step bootstrap value $Q^{(n)}$ to compute the advantage, the true return $\G$ is used. 

Inspired by the supervised autoencoder (SAE) framework \cite{sae}, Our previous work \cite{de2019jointly} also blended in an unsupervised loss for pre-training. In SAE, an image reconstruction loss is incorporated with the supervised loss to help extract better feature representations and achieve better performance. A BC model pre-trained jointly with supervised, value, and unsupervised losses can lead to better performance after fine-tuning with RL, compared to pre-training with the supervised loss only.

We copy this approach by jointly pre-training the BC model for 50,000 steps with a minibatch of size 32. Adam optimizer is used with a learning rate = 0.0005. After training, we perform testing for 50 episodes 
by executing the model greedily in the game and record the average episodic reward (an episode ends when the agent loses all its lives). For each set of demonstration data, we train five models and use the one with the highest average episodic reward as the BC model 
in LiDER-BC. The performance of the trained BC models is present in Table \ref{table:bc}. All parameters are based on those from our previous work \cite{de2019jointly} and we summarize them in Table \ref{table:bc-parameters}.

\vspace{-10pt}

\begin{table}[h]
    \caption{The performance of behavior cloning models used in LiDER-BC, shown as the black dashed line in Figure \ref{fig:lider-ta-bc}. The score is estimated by executing the BC greedily in the game for 50 episodes.}
    \label{table:bc}
    \centering
    \adjustbox{max width=.8\textwidth}{
        \begin{tabular}{l|c|c}
        \hline \hline
        Game                 & Trained BC model score & standard deviation \\\hline\hline
        Gopher               & 450.8    & 393.57             \\ \hline
        NameThisGame         & 1491.2   & 530.55             \\ \hline
        Alien                & 839.2    & 718.72             \\ \hline
        Ms.~Pac-Man          & 1776.6   & 993.94             \\ \hline
        Freeway              & 25.06    & 1.48               \\ \hline
        Montezuma's Revenge  & 174.0    & 205.72             \\ \hline
        \end{tabular}
    }
\end{table}

\vspace{-30pt}

\begin{table}[h]
    \caption{Hyperparameters for pre-training the behavior cloning (BC) model used in LiDER-BC.}
    \label{table:bc-parameters}
    \centering
    \adjustbox{max width=\textwidth}{
        \begin{tabular}{r|r}
        \hline \hline
        \textbf{Network Architecture}        & \textbf{Value} \\ \hline \hline
        Input size             & 88$\times$88$\times$4 \\ \hline
        Tensorflow Padding method         & SAME     \\ \hline 
        Convolutional layer 1 & 32 filters of size 8$\times$8 with stride 4 \\ \hline 
        Convolutional layer 2 & 64 filters of size 4$\times$4 with stride 2 \\ \hline
        Convolutional layer 3 & 64 filters of size 3$\times$3 with stride 1 \\ \hline
        Fully connected layer & 512 \\ \hline
        Classification output layer & number of actions \\ \hline \hline
        Value output layer & 1 \\ \hline \hline
        \multicolumn{2}{c}{\textbf{Parameters for pre-training}} \\ \hline \hline
        Adam learning rate &  $5\times 10^{-4}$ \\ \hline
        Adam epsilon &  $1\times 10^{-5}$ \\ \hline
        Adam $\beta_1$ &  $0.9$ \\ \hline
        Adam $\beta_2$ &  $0.999$ \\ \hline
        L2 regularization weight & $1\times 10^{-5}$     \\ \hline
        Number of minibatch updates & 50,000     \\ \hline
        Batch size             & 32     \\ \hline \hline
        \end{tabular}
    }
\end{table}

\end{document}